\documentclass[twoside,11pt]{article}

\usepackage{jmlr2e}
\usepackage{microtype}
\usepackage{graphicx}
\usepackage{subcaption}
\usepackage{booktabs} 
\usepackage{amsmath}
\usepackage{amssymb}
\usepackage{mathtools}
\usepackage[capitalize,noabbrev]{cleveref}
\usepackage{fancybox} 
\usepackage{tabu}
\usepackage{arydshln}
\usepackage{tikz}
\usepackage{collcell}
\usepackage{xstring}
\usepackage{floatrow}
\usepackage{algorithmic}
\usepackage{algorithm}
\usepackage{float}
\floatstyle{plaintop}
\restylefloat{table}

\def \w {\mathbf{w}}
\def \v {\mathbf{v}}

\def \x {\mathbf{x}}

\def \E {\mathrm{E}}

\def \x {\mathbf{x}}

\def \1 {\mathbf{1}}

\def \u {\mathbf{u}}

\def \B {\mathcalB}

\def \E {\mathrm{E}}
\def \x {\mathbf{x}}

\def \D {\mathcal{D}}

\def \u {\mathbf{u}}

\def \w {\mathbf{w}}

\def \v {\mathbf{v}}

\def \B {\mathcal{B}}

\ShortHeadings{Benchmarking Deep AUROC Optimization}{}
\firstpageno{1}

\begin{document}

\title{Benchmarking Deep AUROC Optimization: \\Loss Functions and Algorithmic Choices}

\author{\name Dixian Zhu \email dixian-zhu@uiowa.edu \\
       \addr Department of Computer Science\\
       University of Iowa\\
       Iowa City, IA 52246, USA
       \AND
       \name Xiaodong Wu \email xiaodong-wu@uiowa.edu \\
       \addr  Department of Electrical and Computer Engineering\\
       University of Iowa\\
       Iowa City, IA 52246, USA
       \AND
       \name Tianbao Yang \email tianbao-yang@uiowa.edu \\
       \addr Department of Computer Science\\
       University of Iowa\\
       Iowa City, IA 52246, USA}

\editor{}

\maketitle

\begin{abstract}
The area under the ROC curve (AUROC) has been vigorously applied for imbalanced classification and moreover combined with deep learning techniques. However, there is no existing work that provides sound information for peers to choose appropriate deep AUROC maximization techniques. In this work, we fill this gap from three aspects. (i) We benchmark a variety of loss functions with different algorithmic choices for deep AUROC optimization problem. We study the loss functions in two categories: pairwise loss and composite loss, which includes a total of 10 loss functions. Interestingly, we find composite loss, as an innovative loss function class, shows more competitive performance than pairwise loss from both training convergence and testing generalization perspectives. Nevertheless, data with more corrupted labels favors a pairwise symmetric loss. (ii) Moreover, we benchmark and highlight the essential algorithmic choices such as positive sampling rate, regularization, normalization/activation, and optimizers. Key findings include: higher positive sampling rate is likely to be beneficial for deep AUROC maximization; different datasets favors different weights of regularizations; appropriate normalization techniques, such as sigmoid and $\ell_2$ score normalization, could improve model performance. (iii) For optimization aspect, we benchmark SGD-type, Momentum-type, and Adam-type optimizers for both pairwise and composite loss. Our findings show that although Adam-type method is more competitive from training perspective, but it does not outperform others from testing perspective.
\end{abstract}


\section{Introduction}
It is widely accepted that the area under the ROC curve (AUROC)~\citep{hanley1982meaning} is a better metric than accuracy for classification problems on imbalanced datasets because the majority class could be dominating in the calculation of the accuracy metric and the surrogate loss function as well for model training process. On the other hand, AUROC evaluates how well a model classifies positive samples given threshold fixed by every negative sample. Alternatively, AUROC could also be interpreted as the probability that a random positive sample enjoys a larger predicted value than a random negative sample. AUROC metric has been applied in a variety of areas, such as, medical imaging diagnosis, molecular property prediction, natural hazards, meteorology~\citep{irvin2019chexpert,wu2018moleculenet,peres2014derivation,murphy1996finley}; and potentially any imbalanced datasets. Moreover, deep learning has been achieving many successes since last decade in many areas, such as, computer vision, natural language processing, molecular property prediction~\citep{krizhevsky2012imagenet,vaswani2017attention,wu2018moleculenet}. Given that deep learning could handle more complicated tasks than traditional machine learning, it is unavoidable to consider optimizing AUROC under the deep learning setting. 

AUROC optimization has been vigorously studied since last two decades. Some of the early studies utilize boosting algorithms, support vector machine (SVM), decision tree as the learning framework to optimize AUROC~\citep{cortes2003auc,joachims2005support,ferri2002learning}. However, unfortunately those approaches cannot be applied to deep learning setting. Inspired by Wilcoxon-Man-Whitney statistic, a variety of pairwise losses have been studied for AUROC optimization, e.g., pairwise square loss, pairwise squared hinge loss, pairwise hinge loss, pairwise logistic loss, pairwise sigmoid loss and pairwise barrier hinge loss~\citep{gao2013one,zhao2011online,kotlowski2011bipartite,gao2015consistency,calders2007efficient,charoenphakdee2019symmetric}; all of them are convex surrogate functions of 0-1 loss for AUROC and could be potentially applied to deep learning setting by mini-batch approximation. 
In this paper, we investigate the performance of deep AUROC maximization with a broader family of objectives beyond the pairwise losses, which includes a family of composite losses.  The bridge between the family of pairwise losses and the family of composite losses is a min-max objective corresponding to the pairwise square loss function~\citep{ying2016stochastic}. The min-max objective has been extended to a min-max margin objective (known as AUC-M) in~\citep{yuan2020robust}. In order to study a broad family of objectives that includes min-max objective and min-max margin objective, we formulate a family of composite loss, which is composed of three components, where the first two components correspond to the within-class variance of prediction scores for both positive and negative class, the last component penalizes the large difference between the average score of positive data and the average score of negative data.  
Similar to the pairwise loss, the composite loss could have different variants by utilizing different surrogate loss functions for formulating the third component, such as hinge loss, logistic loss, sigmoid loss, etc. Overall, there are more than ten different choices of loss functions for deep AUROC maximization. 

Moreover, over-sampling and algorithmic regularization for model parameters are two techniques that are useful for learning on the imbalanced dataset~\citep{cao2019learning,yuan2020robust}. Besides, other algorithmic choices  are essential for deep learning, such as weight decay, activation  normalization, and optimizers~\citep{goodfellow2016deep}. To the best of our knowledge, there is no benchmark research for AUROC optimization under deep learning setting. This work investigates different methods for AUROC optimization under deep learning setting with the consideration of multiple key algorithmic choices. Our contribution could be summarized as three points:
\vspace{-0.2 cm}
\begin{itemize}
    \item We benchmark all existing ten main-stream loss functions (either pairwise or composite) for deep AUROC maximization  with the considerations of different algorithmic choices, including different positive sampling ratios, algorithmic regularization, weight decay, activation/normalization functions at the output layer, and optimizers. We conduct experiments on six image datasets and six molecule graph datasets, including large scale dataset CheXpert that contains 191,027 samples and five targets for recognition. To the best of our knowledge, this is the most comprehensive study for deep AUROC learning to date.
    \item We analyze the our findings based on our 14,400 unique runs and summarize the insights from three aspects: (i) how different algorithmic choices impact the learning performance. Specifically, we find over-sampling is useful for deep AUROC maximization; the sigmoid and $\ell_2$  score normalization are two effective choices for the output layer of deep learning model. (ii) Optimizer matters: Adam-type optimizers have better training performance for most of the experiments; however, for validation and testing performance, it is worse than Momentum and SGD type of methods. The finding verifies the drawback for adaptive optimizers from previous work~\citep{wilson2017marginal}. (iii) For loss functions: the composite loss function is more competitive from both training and testing perspective. On the other hand, using the pairwise barrier hinge loss could achieve better testing performance for the data (CheXpert Edema and Consolidation) with larger label noise~\citep{charoenphakdee2019symmetric}.
    \item We package and publish our research code in a modularized manner that would be easy for peers to conduct future experiments and make extensions. All the ten loss functions, multiple networks and optimizers in this work could be further utilized for future studies on deep AUROC maximization. 
\end{itemize}

\subsection{Related Work}
The benchmark for deep learning is more challenging because there are many factors (network structure, activation function, normalization, optimizer, etc.) that matter. Moreover, deep learning is an emerging area that has been studied vigorously during the last decade. Therefore, the benchmark research for deep learning is limited. Some prior work benchmarks reinforcement deep learning~\citep{duan2016benchmarking}, interpretability for time series prediction~\citep{ismail2020benchmarking}, optimizers for deep learning~\citep{schmidt2021descending}. There is no existing benchmark research with different deep AUROC maximization methods. Most of the prior knowledge comes from the previous deep AUROC maximization research work that proposes a novel method and then presents comparison with several other baselines~\citep{yuan2020robust}. Given that we have a large space of possible combinations of algorithmic choices for deep AUROC maximization, there is no clear answer or insights for how to choose appropriate techniques for a task. 

\section{Methods}
In this section, we first illustrate the loss functions, optimizers and other algorithmic choices that are utilized for benchmarking. Then we introduce the benchmarking process for this work.
\subsection{Loss Functions}\label{sec:loss}
According to the definition for AUROC~\citep{hanley1982meaning}:
$$\text{AUC}(\w)=\E_{\x,\x'}[\mathbb{I}(h_\w(\x)\ge h_\w(\x'))|y=+1,y'=-1]$$
where $h_\w(\cdot)$ denotes model prediction with model parameters $\w$; $\x$ and $\x'$ represent a positive data and negative data; $y$ and $y'$ represent their class labels, respectively. We consider binary classification by AUROC maximization in this work. Based on this definition, a surrogate loss is usually used. Denote by $\ell(\cdot)$ a surrogate
loss, one usually minimizes the following empirical AUC loss:
$$\min_\w\frac{1}{N_+}\frac{1}{N_-}\sum_{x_i\in\D_+}\sum_{\x_j\in\D_-}\ell(h_\w(\x_i)-h_\w(\x_j))$$
where $\D_+$ represents positive data pool with size $N_+$ and $\D_-$ represents negative data pool with size $N_-$. Given that the form of the loss function involves pairs of positive and negative data samples, it is known as \textbf{pairwise loss}. We study six different pairwise loss functions in this work, which are summarized at Table~\ref{tab:pairwise-def}. It is notable that pairwise barrier hinge (PBH) loss is a symmetric loss, which was proposed for handling noisy labels. 
\begin{table}[t]
    \centering
    \resizebox{\linewidth}{!}{
    \begin{tabular}{lll}
       Loss name & reference &  Formulation for Each Pair, ~$\ell(\cdot)$\\
        \hline
        Pairwise Square (PSQ)&\citep{gao2013one} &  $\ell_{\text{PSQ}}(t,c)=(c-t)^2$ \\
        Pairwise Squared Hinge (PSH)&\citep{zhao2011online} &  $\ell_{\text{PSH}}(t,c)=(c-t)_+^2$ \\
        Pairwise Hinge (PH)&\citep{kotlowski2011bipartite} &  $\ell_{\text{PH}}(t,c)=(c-t)_+$ \\
        Pairwise Logistic (PL) &\citep{gao2015consistency} &  $\ell_{\text{PL}}(t,s)=\log(1+\exp(-st))$ \\
        Pairwise Sigmoid (PSM)&\citep{calders2007efficient} &  $\ell_{\text{PSM}}(t,s)=(1+\exp(st))^{-1}$ \\
        Pairwise Barrier Hinge (PBH)&\citep{charoenphakdee2019symmetric} &  $\ell_{\text{PBH}}(t,s,c)=\max(-s(c+t)+c, \max(s(t-c), c-t))$ \\
    \end{tabular}}
    \caption{Pairwise loss functions. For the sake of simplicity, denote $\max(0,t)$ by $t_+$, denote the scaling hyper-parameter by $s$ and margin hyper-parameter by $c$. For AUROC learning: $t=h_\w(\x_i)-h_\w(\x_j), x_i\in\D_+, x_j\in\D_-$. Pairwise loss is defined as the empirical mean of their individual formulation for all pairs, i.e. $\frac{1}{N_+}\frac{1}{N_-}\sum_{x_i\in\D_+}\sum_{\x_j\in\D_-}\ell(h_\w(\x_i)-h_\w(\x_j))$.}
    \label{tab:pairwise-def}
\end{table}

Some researchers have studied a min-max objective corresponding to a pairwise square loss for online/stochastic setting~\citep{ying2016stochastic}:
\begin{align*}
    \min_{\w,a,b}\E_{\x|y=1}[(h_\w(\x)&-a)^2]+\E_{\x'|y'=-1}[(h_\w(\x')-b)^2]\\
    +&\max_\alpha\alpha(c+b(\w)-a(\w))-\frac{1}{2}\alpha^2,
\end{align*}
where $a(\w)=\E[h_\w(\x)|y=1]$ and $b(\w)=\E[h_\w(\x)|y=-1]$.  The min-max margin objective (i.e., AUC-M loss)~\citep{yuan2020robust} adds a constraint for $\alpha$ as $\alpha\geq 0$. A stochastic primal-dual stochastic algorithm (PESG) can be used for solving the above min-max objectives~\citep{yuan2020robust}.

We study a family of composite losses given by  the following formulation:
\begin{align*}
    \min_{\w, a, b}\E_{\x|y=1}[(h_\w(\x)&-a)^2]+\E_{\x'|y'=-1}[(h_\w(\x')-b)^2]\\
    &+\ell(a(\w)-b(\w)),
\end{align*}
where $\ell(\cdot)$ is a proper surrogate function.  When $\ell(\cdot)$ is a square function, the above composite loss is equivalent to min-max objective of the square loss, and when it is squared hinge function it reduces to the min-max margin objective.  
We study four different forms for the composite loss based on different $\ell$. We summarize them in Table~\ref{tab:composite-def}. It is worth noting that CSH loss is equivalent with AUC-M loss~\citep{yuan2020robust}.

\begin{table}[t]
    \centering
        \caption{Composite loss functions. For the sake of simplicity, we denote $\max(0,t)$ by $t_+$, and denote the scaling hyper-parameter by $s$ and margin hyper-parameter by $c$.}
    \resizebox{1.0\textwidth}{!}{
    \begin{tabular}{ll}
       Loss name  & Formulation\\
        \hline
        Composite Square (CSQ)& $\E_{\x|y=1}[(h_\w(\x)-a)^2]+\E_{\x'|y'=-1}[(h_\w(\x')-b)^2]+\frac{1}{2}(c+b(\w)-a(\w))^2$ \\
        Composite Squared Hinge (CSH)& $\E_{\x|y=1}[(h_\w(\x)-a)^2]+\E_{\x'|y'=-1}[(h_\w(\x')-b)^2]+\frac{1}{2}(c+b(\w)-a(\w))_+^2$ \\
        Composite Hinge (CH)& $\E_{\x|y=1}[(h_\w(\x)-a)^2]+\E_{\x'|y'=-1}[(h_\w(\x')-b)^2]+(c+b(\w)-a(\w))_+$ \\
        Composite Logistic (CL)& $\E_{\x|y=1}[(h_\w(\x)-a)^2]+\E_{\x'|y'=-1}[(h_\w(\x')-b)^2]+\log(1+\exp(s(b(\w)-a(\w))))$ \\
    \end{tabular}}
    \label{tab:composite-def}
    \vspace*{-0.1in}
\end{table}

\subsection{Optimizers}
We provide a brief description for optimizers at this section. Due to limit of space, the detailed steps of algorithms are included in the supplement.   We benchmark three different types of optimizers for both pairwise and composite losses, namely SGD-style, Momentum-style and Adam-style.

For optimizing the pairwise loss, we utilize the standard SGD, Momentum and Adam optimizers~\citep{sutskever2013importance,kingma2014adam}. A unified description of these optimizers is presented in Alg.~\ref{alg:pairwise}.

For optimizing the composite loss, we develop SGD, Momentum and Adam style optimizers. The development is inspired by a vast literature of stochastic compositional optimization~\citep{wang2017scgd,ghadimi2020nasa,wang2021momentum} for handling the compositional term $\ell(c+b(\w) - a(\w))$.  The unified description of all optimizers are presented in Alg.~\ref{alg:composite}. It is notable that the SGD-style optimizer is applying the SCGD method in~\citep{wang2017scgd}, the Momentum-style optimizer is applying the NASA method in~\citep{ghadimi2020nasa}, and the Adam-style optimizer is a simplified version of that proposed in~\citep{wang2021momentum}. 
\begin{algorithm}[h]
   \caption{Optimizer for pairwise loss}
   \label{alg:pairwise}
\begin{algorithmic}
   \STATE {\bfseries Input:} dataset, iteration number $T$. Individual pairwise loss function $\ell(\cdot)$. Learning rate schedule $\eta_t$. Optimizer types at \{`SGD', `Momentum', `Adam'\}. $\beta_t$ as extra momentum parameter, $\beta_t'$ and $G_0$ as extra Adam parameters.
   \FOR{$t=0$ {\bfseries to} $T$}
   \STATE Sample mini-batch of positive data $\B_+$ and negative data $\B_-$ and compute stochastic gradient: $$\nabla_{F}(\w_t)=\sum_{x_i\in\B_+,\x_j\in\B_-}\frac{\nabla\ell(h_\w(\x_i)-h_\w(\x_j))}{(|\B_+||\B_-|)}$$
   \IF{optimizer = `SGD'}
   \STATE $g_{t+1}=\nabla_{F}(\w_t)$
   \ELSE
   \STATE $\v_{t+1}=(1-\beta_t)\v_{t}+\beta_t\nabla_{F}(\w_t)$
   \IF{optimizer = `Momentum'}
   \STATE $g_{t+1}=\v_{t+1}$
   \ELSIF{optimizer = `Adam'}
   \STATE $\u_{t+1}=(1-\beta')\u_t+\beta'\nabla_{F}^2(\w_t)$
   \STATE $g_{t+1}=\frac{\v_{t+1}}{\sqrt{\u_{t+1}+G_0}}$
   \ENDIF
   \ENDIF
   \STATE $\w_{t+1} = \w_t-\eta_t g_t$
   \ENDFOR
\end{algorithmic}
\end{algorithm}

\begin{algorithm}[h]
   \caption{Optimizer for composite loss}
   \label{alg:composite}
\begin{algorithmic}
   \STATE {\bfseries Input:} dataset, iteration number $T$. Composite loss function with $\ell(\cdot)$. Learning rate schedule $\eta_t$. Optimizer types at \{`SGD-style', `momentum-style', `Adam-style'\}. $\beta_0$ for composite function moving average parameter. $\beta_t$ as extra momentum parameter, $\beta_t'$ and $G_0$ as extra Adam parameters.
   \STATE Encode model parameters as $\bar\w=(\w,a,b)$
   \FOR{$t=0$ {\bfseries to} $T$}
   \STATE Sample mini-batch of positive data $\B_+$ and negative data $\B_-$ and compute the following components:
   \begin{align*}
       &H_+(\w_t,a_t)=\sum_{\x_i\in\B_+}(h_{w_t}(\x_i)-a_t)^2/|\B_+|\\
       &H_-(\w_t,b_t)=\sum_{\x_j\in\B_-}(h_{w_t}(\x_j)-b_t)^2/|\B_-|\\
       &A(\w_t)=\sum_{\x_i\in\B_+}h_{w_t}(\x_i)/|\B_+|\\
       &B(\w_t)=\sum_{\x_j\in\B_-}h_{w_t}(\x_j)/|\B_-|\\
       &d_{t+1}=(1-\beta_0)d_t + \beta_0(A(\w_t)-B(\w_t))
   \end{align*}
   \begin{align*}
       \nabla_F(\bar\w_t)&=\partial H_+(\w_t,a_t)+\partial H_-(\w_t,b_t)+\nabla\ell(d_{t+1})(\partial A(\w_t)-\partial B(\w_t))
   \end{align*}
   \IF{optimizer = `SGD-style'}
   \STATE $g_{t+1}=\nabla_{F}(\bar\w_t)$
   \ELSE
   \STATE $\v_{t+1}=(1-\beta_t)\v_{t}+\beta_t\nabla_{F}(\bar\w_t)$
   \IF{optimizer = `momentum-style'}
   \STATE $g_{t+1}=\v_{t+1}$
   \ELSIF{optimizer = `Adam-style'}
   \STATE $\u_{t+1}=(1-\beta')\u_t+\beta'\nabla_{F}^2(\bar\w_t)$
   \STATE $g_{t+1}=\frac{\v_{t+1}}{\sqrt{\u_{t+1}+G_0}}$
   \ENDIF
   \ENDIF
   \STATE $\bar\w_{t+1} = \bar\w_t-\eta_t g_{t+1}$
   \ENDFOR
\end{algorithmic}
\end{algorithm}

\subsection{Algorithmic Choices}
We introduce the four deep learning algorithmic aspects for our benchmark research at this subsection. We investigate the best choices for these aspects.

{\bf Sampling Positive Rate (SPR):} previous research has  shown that over-sampling could be beneficial for learning on the imbalanced dataset~\citep{cao2019learning}. However, it has not been investigated for deep AUROC maximization.  The SPR is defined as: $\text{SPR}=|\B_+|/|\B|$, where $\B_+$ is the sampled positive set and $\B$ includes all sampled examples per-iteration. 

{\bf Consecutive Epoch Regularization (CER)}~\citep{yuan2020robust}: this is an algorithmic regularization for model parameters that is added to each stage of training (one stage is defined as a number of iterations using a fixed learning rate). It is defined as: $\text{CER}=\gamma\|\w - \bar\w^{k-1}\|_2^2,$
    where $\bar\w^{k-1}$ is the averaged model parameter from the  previous stage.

{\bf Weight Decay (WD)}: this is a standard $\ell_2$-norm regularization of the model parameter:
    $\text{WD}=\frac{\lambda}{2}\|\w\|_2^2.$

{\bf Normalization/Activation for output layer}: Normalization is generally effective for deep learning training process. Specifically for deep AUROC maximization, it further provides the benefit that the outputs could be regularized into a  certain region, which might prevent extreme outlier prediction for AUROC. In the following experiments,  we consider an $\ell_2$-norm normalization that is applied to the  non-activated scores of the mini-batch examples, which is adopted in the previous work~\citep{yuan2020robust}, and a sigmoid activation function that directly transforms the output into $[0,1]$. We also study  $\ell_1$-norm normalization applied to the non-activated scores,  and the batch-normalization as well.

\subsection{Benchmarking Process}
\textbf{Exhaustive tuning is impossible.}
There are ten loss functions to benchmark in this work. For each loss functions, there could be scale or/and margin hyper-parameters need to be tuned. There are also many other choices to be tuned, including  three types of optimizers and their learning rates, and four other algorithmic aspects mentioned above.  Once we fix all the settings (more details are included in Section~\ref{sec:experiment}), we have to run multiple times to reduce the random effects to make results more robust. Last but not the least, we have twelve datasets to conduct experiments. Based on our tuning settings and a simplified calculation: \{10: loss functions\}$\times$\{3: hyper-parameter range for loss function\}$\times$\{3: optimizer type\}$\times$\{5: learning rate\}$\times$\{4: sampling positive rate\}$\times$\{3: consecutive epoch regularization\}$\times$\{5: weight decay\}$\times$\{5: normalization\}$\times$\{12: datasets\}$\times$\{5: repeat by different data splits or random seeds\} = 8,100,000. It is the estimated number of unique runs. Notice that some loss functions such as PBH, need more tuning efforts because it has two hyper-parameters and CheXpert has multiple targets. Hence, the big number is even optimistic.

\textbf{Simplified Tuning Process.} To simplify the tuning efforts, we will focus on two loss functions namely PSQ from the pairwise loss family and CSQ from the compsite loss family in order to study the effect of SPR, CER, WD, normalization/activation and optimizers. For each of these losses, we first tune their hyper-parameters,  optimizer types, learning rate of the optimizer. Once the optimal choices of these coordinates are fixed, we study the effect of SPR, CER, WD, normalization/activation, and optimizer.  This gives a total of \{2:loss functions\}$\times$\{3: hyper-parameter range for loss function\}$\times$\{12: datasets\}$\times$\{5: repeat by different data or random seed\}$\times$ (\{3: optimizer type\}$\times$\{5: learning rate\}+ \{4: sampling positive rate\} + \{3: consecutive epoch regularization\} + \{5: weight decay\} + \{5:normalization\}) = 11,520 unique runs.  
Lastly, we compare different loss functions on different datasets, which gives \{11+5: tasks\}$\times$\{9+3: loss functions, PBH contributes for 3 times more tuning\}$\times$\{3: hyper-parameter range for each loss function\}$\times$\{5: repeats by different data splits or random seeds\} = 2880. The number of all unique runs is 14,400.

The high level overview for the process is that we experiment with different settings sequentially (SPR$\rightarrow$CER$\rightarrow$WD$\rightarrow$normalization/activation$\rightarrow$optimizer). Once we finish the tuning of one option, we fix it as a reasonable value based on the results at the time. Until we finish the tuning for all settings, we choose the best combination of the algorithmic settings and optimizer settings for the pairwise loss and the composite loss  and run the final comparison experiments for all loss functions on the datasets.

\section{Experiments}\label{sec:experiment}

The experiments are conducted on 6 image datasets and 6 molecule datasets, namely: STL10, CIFAR10, CIFAR100, Cat vs Dog, Melanoma, CheXpert, HIV, MUV, Tox21(t0), Tox21(t2), ToxCast(t8), ToxCast(t12)~\citep{coates2011analysis,krizhevsky2009learning,elson2007asirra,rotemberg2021patient,irvin2019chexpert,wu2018moleculenet}. We arrange this section as following: 1) provide the information for the datasets and pre-processing. 2) illustrate the experiment setups and steps we tune the hyper-parameters. 3) deliver the final comprehensive comparison for all loss functions. 4) present numerical analysis for training by different optimizers.
\subsection{Datasets, Pre-processing and Settings}
\subsubsection{Datasets and Pre-processing}
All of the molecule datasets and the two medical image datasets (Melanoma and CheXpert) are naturally imbalanced, which are suitable for AUROC maximization. For the four datasets: STL10, CIFAR10, CIFAR100, Cat vs Dog~\footnote{Data available at:~\url{https://libauc.org}}, we manually make it imbalanced in order to better fit the AUROC optimization task. The process is as following: first we treat the first half classes with smaller class index as positive samples, and the remaining as negative samples; then for training data we manually delete partial positive samples to certain amount so that the positive ratio (PR), i.e. 
$|\D_+|/|\D|$ is 10\% and keep the origin 50\% positive ratio for testing data. We keep the original train/test split for these four datasets.

For Melanoma dataset~\footnote{\url{https://challenge2020.isic-archive.com/}}, we split the original data as training and testing according to  90\%/10\%; and we resize each image to 96$\times$96 resolution. For CheXpert dataset~\footnote{\url{https://stanfordmlgroup.github.io/competitions/chexpert/}}, we use the original training as training data and validation as testing data for this work. The image resolution is scaled to 224$\times$224. We follow the same way to handle uncertain and missing labels as the previous research~\citep{yuan2020robust}. For the sake of simplicity, we directly refer to the targeted diseases of CheXpert: cardiomegaly, edema, consolidation, atelectasis, pleural effusion, as datasets below.

The MUV dataset has 93,127 molecules from the PubChem library, and molecules
are labelled by whether a bioassay property exists or not. Note that the MUV dataset provides labels of 17 properties in total and we only conduct experiments to predict the third property as this property is more imbalanced. The Tox21 and ToxCast contain 8014 and 8589 molecules, respectively. There are 12 property
prediction tasks in Tox21, and we conduct experiments on Task 0 and Task 2. Similarly, we select
Task 12 and Task 8 of ToxCast for experiments. Because those targets are more imbalanced. For all molecule datasets, we utilize 90\%/10\% train/test split as the provided in the previous research~\citep{wu2018moleculenet}. The statistics of all the benchmark datasets are summarized in Table~\ref{tab:data_stat}.
\begin{table}[h!]
    \centering
        \caption{Statistics of Datasets. PR means the ratio of the number of positive examples to the total number of examples.}
    \label{tab:data_stat}
    \resizebox{0.9\textwidth}{!}{
    \begin{tabular}{lllll}
       Dataset & Domain&\# train / test& dimension& PR(train) / (test) (\%)\\
        \hline
        STL10 &Image& 2777 / 8000  & 96$\times$96  & 10.0 / 50.0 \\
        CIFAR10 &Image& 27777 / 10000  & 32$\times$32  & 10.0 / 50.0  \\
      CIFAR100 &Image& 27777 / 10000  & 32$\times$32  & 10.0 / 50.0  \\
        Cat vs Dog&Image & 11128 / 5000  & 50$\times$50  &  10.0 / 50.0 \\
        Melanoma  &Image& 29814 / 3312  & 96$\times$96  & 1.8 / 1.5  \\
        Cardiomegaly &Image & 191027 / 202  & 224$\times$224 & 12.2 / 32.7  \\
        CheXpert/Edema  &Image &191027 / 202  & 224$\times$224 & 32.2 / 20.8  \\
        CheXpert/Consolidation &Image & 191027 / 202  & 224$\times$224 & 6.8 / 15.8  \\
        CheXpert/Atelectasis &Image & 191027 / 202  & 224$\times$224 & 31.2 / 37.1  \\
        CheXpert/Pleural Effusion  &Image& 191027 / 202  & 224$\times$224 & 40.3 / 31.7  \\
        HIV  &Graph&   37721 / 4192 & NA & 3.6 / 3.1 \\
        MUV(bioassay) &Graph &   13172 / 1562  & NA &  0.2 / 0.2\\
        Tox21(t0)  &Graph& 6687 / 752  & NA & 4.1 / 4.4\\
        Tox21(t2)  &Graph& 6022 / 669  & NA & 11.7 / 12.1 \\
        ToxCast(t8) &Graph& 936 / 97 & NA  & 8.8 / 9.3 \\
        ToxCast(t12) &Graph& 936 / 97 & NA  & 3.0 / 1.0 \\
        \hline
    \end{tabular}}
\end{table}

\subsubsection{Basic Settings}
We first introduce the basic settings of our experiments that are kept unchanged throughout this work. For all  experiments, we use 64 as the mini-batch size. We use 50 as the total training epochs for all datasets except for CheXpert. The learning rate is decreased by ten folds at the end of 30-th and 40-th epoch. For CheXpert, we train 2 epochs and decrease the learning rate by ten folds at the end of the first epoch similar as the previous research~\citep{yuan2020robust}. For each individual run, we utilize five-fold cross validation to choose the best hyper-parameter and the evaluation stopping point for each loss function to evaluate test AUROC. 

More specifically, for pairwise square loss (PSQ), pairwise squared hinge loss (PSH), pairwise hinge loss (PH), composite squared loss (CSQ), composite squared hinge loss (CSH), composite hinge loss (CH), we tune the margin parameter from \{0.1, 1.0, 10.0\} when there is no activation function for the output layer, and from \{0.1, 0.5, 1.0\} whenever there is an activation or normalization function (such as sigmoid, $\ell_1$ and $\ell_2$ normalization) that scale the magnitude of the final prediction less equal to 1.0. For pairwise logistic loss (PL), pairwise sigmoid loss (PSM), and composite logistic loss (CL), we tune the scaling parameter from \{0.1, 1.0, 10.0\}. For pairwise barrier hinge loss (PBH), we tune the margin parameter by the same way as PSQ, and we tune its scaling parameter by the same way as PL. After five-fold cross validation, we report mean and standard deviation of the desired measurements, such as testing AUROC, validation AUROC and training loss. 

For CheXpert and molecule datasets, we pre-train deep neural network with cross entropy loss; and we train from scratch for the other simpler benchmark datasets. For image datasets, we utilize densenet-121 for CheXpert, resnet-20 for the remaining image datasets~\citep{huang2017densely,he2016deep}. For molecule datasets, we utilize message passing neural network (MPNN) for MUV(bioassay), Tox21(t2) and ToxCast(t12) datasets~\citep{gilmer2017neural}; we utilize graph isomorphism network (GINE) for Tox21(t0) and ToxCast(t8) datasets~\citep{hu2019gine,xu2018powerful}; we utilize multi-level message passing neural network (ML-MPNN) for HIV dataset~\citep{wang2020advanced}.

\subsection{Hyper-parameter Tuning}
As a prior work proved, the PSQ loss and CSQ loss are equivalent but with different forms~\citep{ying2016stochastic}. Therefore, we choose them as representatives for pairwise loss and composite loss for parameter tuning to avoid tuning burdens. We conduct experiments for PSQ loss and CSQ loss from five aspects: sampling positive ratio (SPR) for mini-batch, consecutive epoch regularization (CER), weight decay (WD), normalization function at prediction (last) layer, and optimizer with different learning rates. In order to get more direct insights, we report the test AUROC in the following main content for each algorithmic or optimizer setting. It is worth noting that we use the validation AUROC for choosing the best hyper-parameter in our process and do final training/evaluation for each loss function by using the best combination of the algorithmic and optimizer settings; and we include the validation AUROC results at the appendix. Because CheXpert has a huge data size which would take tremendous cost for tuning, {we only tune the cardiomegaly target and adopt the same hyper-parameters for the other CheXpert targets}.

\begin{table}[h]
    \centering
    \caption{Tesing AUROC with different sampling positive ratios (SPR)}
    \label{tab:te_SPR}
    \resizebox{0.8\textwidth}{!}{ 
    \begin{tabular}{llllll}
    \toprule
       Loss &Dataset &SPR$=$origin & SPR$=5\%$& SPR$=25\%$& SPR$=50\%$\\
        \hline
        &STL10 &  0.742(0.020) & 0.704(0.017) & \textbf{0.771}(0.022) & 0.763(0.014)\\
        &CIFAR10 &  0.864(0.003) & 0.840(0.005) & 0.884(0.008) & \textbf{0.886}(0.003)\\
        &CIFAR100 &  \textbf{0.659}(0.008) & 0.640(0.007) & 0.658(0.015) & \textbf{0.659}(0.010)\\
        &Cat vs Dog &  0.900(0.006) & 0.874(0.008) & 0.918(0.005) & \textbf{0.921}(0.007)\\
        &Melanoma &  0.779(0.022) & \textbf{0.828}(0.014) & 0.821(0.021) & 0.809(0.018)\\
        &Cardiomegaly &  0.816(0.017)  &  0.829(0.005)  &  \textbf{0.835}(0.014)  & 0.820(0.009)  \\
        PSQ &HIV &   0.748(0.021) & 0.754(0.007) & \textbf{0.762}(0.014) & 0.759(0.014) \\
        &MUV(bioassay) &   0.617(0.130) & \textbf{0.716}(0.081) & 0.627(0.098) & 0.638(0.033)\\
        &Tox21(t0) &   0.777(0.014) & \textbf{0.781}(0.013) & 0.763(0.013) & 0.747(0.005)\\
        &Tox21(t2) &   0.881(0.007) & \textbf{0.883}(0.006) & \textbf{0.883}(0.014) & 0.880(0.016)\\
        &ToxCast(t8) &   0.455(0.019) & 0.450(0.012) & 0.491(0.040) & \textbf{0.497}(0.029)\\
        &ToxCast(t12) &   0.795(0.079) & 0.760(0.193) & \textbf{0.817}(0.095) & 0.676(0.236)\\
        \midrule
         &STL10 & 0.661(0.100) & 0.622(0.089) & \textbf{0.684}(0.131) & 0.676(0.121)\\
        &CIFAR10 & 0.826(0.056) & 0.809(0.038) & \textbf{0.851}(0.051) & 0.850(0.047)\\
       &CIFAR100 &  0.648(0.027) & 0.629(0.015) & 0.656(0.012) & \textbf{0.658}(0.017)\\
        &Cat vs Dog & 0.843(0.034) & 0.843(0.045) & 0.848(0.072) & \textbf{0.872}(0.049)\\
        &Melanoma &  0.765(0.016) & \textbf{0.835}(0.007) & 0.808(0.019) & 0.810(0.016)\\
        &Cardiomegaly &   0.840(0.006) &  \textbf{0.847}(0.004)  &  0.840(0.005)  & 0.834(0.003)  \\
        CSQ &HIV &   0.739(0.011) & \textbf{0.768}(0.011) & 0.764(0.012) & 0.747(0.012) \\
        &MUV(bioassay) &   0.642(0.074) & 0.686(0.066) & 0.638(0.069) & \textbf{0.702}(0.092)\\
        &Tox21(t0) &   0.769(0.013) & 0.762(0.017) & \textbf{0.773}(0.008) & 0.757(0.016)\\
        &Tox21(t2) &   \textbf{0.895}(0.009) & 0.887(0.006) & 0.893(0.011) & 0.890(0.006)\\
        &ToxCast(t8) &   0.445(0.014) & 0.446(0.017) & 0.469(0.018) & \textbf{0.470}(0.030)\\
        &ToxCast(t12) &   0.716(0.100) & 0.744(0.034) & \textbf{0.760}(0.100) & 0.740(0.109)\\
        \bottomrule
    \end{tabular}}
\end{table}
\begin{table}[h!]
    \centering
    \caption{Testing AUROC with different Consecutive epoch regularizations}
    \label{tab:te_CER}
    \resizebox{0.7\textwidth}{!}{
    \begin{tabular}{lllll}
    \toprule
       Loss &Dataset & $\gamma=0$& $\gamma=0.002$& $\gamma=0.02$\\
        \hline
        &STL10 & 0.763(0.014) & 0.761(0.014) & \textbf{0.809}(0.013)\\
        &CIFAR10 &  0.886(0.003) & \textbf{0.907}(0.002) & 0.890(0.005)\\
      &CIFAR100 & 0.659(0.010) & \textbf{0.669}(0.011) & 0.642(0.012)\\
        &Cat vs Dog & 0.921(0.007) & 0.930(0.004) & \textbf{0.931}(0.008)\\
        &Melanoma &  \textbf{0.809}(0.018) & 0.798(0.012) & 0.792(0.018)\\
        &Cardiomegaly &  0.820(0.007) & 0.832(0.007) & \textbf{0.835}(0.007)\\
       PSQ   &HIV &   \textbf{0.759}(0.014) & 0.757(0.015) & 0.758(0.014) \\
        &MUV(bioassay) &   \textbf{0.638}(0.033) & 0.629(0.052) & 0.525(0.091) \\
        &Tox21(t0) &   \textbf{0.747}(0.005) & 0.744(0.010) & 0.744(0.008) \\
        &Tox21(t2) &   0.880(0.016) & \textbf{0.881}(0.013) & 0.879(0.011)\\
        &ToxCast(t8) &   0.497(0.029) & 0.489(0.031) & \textbf{0.514}(0.050) \\
        &ToxCast(t12) &   0.676(0.236) & 0.677(0.235) & \textbf{0.743}(0.163)\\
        \midrule
         &STL10 &  0.676(0.121) & 0.727(0.063) & \textbf{0.808}(0.013)\\
        &CIFAR10 & 0.850(0.047) & \textbf{0.902}(0.005) & 0.883(0.008)\\
       &CIFAR100 & 0.658(0.017) & \textbf{0.680}(0.008) & 0.646(0.016)\\
        &Cat vs Dog & 0.872(0.049) & 0.926(0.008) & \textbf{0.930}(0.003)\\
        &Melanoma &  \textbf{0.810}(0.015) & 0.802(0.019) & 0.807(0.013)\\
        &Cardiomegaly & 0.834(0.003) & 0.833(0.005) & \textbf{0.835}(0.007)\\
       CSQ  &HIV &   0.747(0.012) & 0.752(0.010) & \textbf{0.763}(0.019) \\
        &MUV(bioassay) &   \textbf{0.702}(0.092) & 0.673(0.096) & 0.631(0.083) \\
        &Tox21(t0) &   0.757(0.016) & \textbf{0.758}(0.017) & \textbf{0.758}(0.017) \\
        &Tox21(t2) &   0.890(0.006) & \textbf{0.893}(0.007) & 0.890(0.007)\\
        &ToxCast(t8) &   0.470(0.030) & 0.489(0.024) & \textbf{0.491}(0.028) \\
        &ToxCast(t12) &   \textbf{0.740}(0.109) & 0.738(0.108) & 0.708(0.142)\\
        \bottomrule
    \end{tabular}}
\end{table}

{\bf The impact of SPR.} For SPR, i.e., the sampling positive ratio enforced for each mini-batch, we investigate four different values: origin, 5\%, 25\% and 50\%, which covers the original positive ratio of the dataset, low, medium, and high positive sampling ratios. If a SPR makes positive sample number below 1 for each mini-batch, we enforce it as 1 in order to make each loss function  well defined. We fix the $\gamma$ in CER as 0, WD as 1e-4, prediction activation function as sigmoid, optimizer as the momentum optimizer with a learning rate of 1e-2 for image datasets and the Adam optimizer with a learning rate of 1e-3 for molecule datasets.

\begin{table}[h]
    \centering
    \caption{Tesing AUROC with different output normalization/activation layers}
    \label{tab:te_activation_partial}
    \resizebox{0.7\textwidth}{!}{  
    \begin{tabular}{lllll}
       Loss &Dataset & None &sigmoid  & $\ell_2$-normalization \\
        \hline
        &STL10  & 0.726(0.030) & \textbf{0.772}(0.023)&  0.769(0.006) \\
        &CIFAR10 & 0.900(0.005) & \textbf{0.911}(0.001)&  0.893(0.006)\\  
      &CIFAR100 &  0.660(0.012) & 0.681(0.007)& \textbf{0.683}(0.006) \\
        &Cat vs Dog &  0.895(0.036) & \textbf{0.932}(0.003)&  0.919(0.004)\\
        &Melanoma &  0.793(0.035) & 0.794(0.013)&  \textbf{0.810}(0.011) \\
        &Cardiomegaly & 0.800(0.010) & \textbf{0.832}(0.007)&  0.815(0.030) \\
        PSQ  &HIV &   \textbf{0.760}(0.010) & 0.734(0.031) &  0.745(0.006)\\
        &MUV(bioassay) &   0.579(0.101) & 0.598(0.063) &  \textbf{0.604}(0.049)\\
        &Tox21(t0) &   0.744(0.008) & \textbf{0.782}(0.004) &  0.768(0.017) \\
        &Tox21(t2) &   0.883(0.009) & 0.883(0.012) &  \textbf{0.888}(0.008) \\
        &ToxCast(t8) &   \textbf{0.518}(0.051) & 0.500(0.040) &  0.477(0.068) \\
        &ToxCast(t12) &  0.735(0.170) & \textbf{0.840}(0.104) &  0.613(0.054) \\
        \midrule
         &STL10 &  0.656(0.079) & 0.751(0.046)&  \textbf{0.763}(0.011) \\
        &CIFAR10 & 0.781(0.048) & \textbf{0.908}(0.004)&  0.894(0.006)\\
       &CIFAR100 &  0.631(0.025) & \textbf{0.676}(0.010)&  0.675(0.007)\\
        &Cat vs Dog &  0.814(0.114) & \textbf{0.929}(0.006)&  0.916(0.009)\\
        &Melanoma & 0.801(0.015) & 0.798(0.013)&  \textbf{0.811}(0.013) \\
        &Cardiomegaly & 0.809(0.015) & \textbf{0.833}(0.005)&  0.814(0.024) \\
        CSQ &HIV &   \textbf{0.756}(0.019) & 0.744(0.017) &  0.741(0.008) \\
        &MUV(bioassay) &   \textbf{0.639}(0.091) & 0.599(0.063) &  0.578(0.107) \\
        &Tox21(t0) &   0.756(0.018) & 0.755(0.005) &  \textbf{0.773}(0.014) \\
        &Tox21(t2) &   0.889(0.004) & \textbf{0.895}(0.005) &  0.892(0.008) \\
        &ToxCast(t8) &   0.490(0.033) & \textbf{0.523}(0.065) &  0.483(0.047) \\
        &ToxCast(t12) &   0.721(0.150) & \textbf{0.985}(0.008) &  0.769(0.061) \\
        \bottomrule
    \end{tabular}}
\end{table}
\begin{table}[h]
\caption{Testing AUROC by using different optimizers for two objectives.}
    \label{tab:te_opt}
    \centering
    \resizebox{0.7\textwidth}{!}{
    \begin{tabular}{lllll}
       Loss &Dataset & SGD-style& Momentum-style& Adam-style  \\
        \hline
        &STL10 &   0.771(0.007) &  \textbf{0.803}(0.015) & 0.794(0.011) \\
        &CIFAR10 &   0.909(0.002) &  \textbf{0.911}(0.001) & 0.868(0.004)\\
      &CIFAR100 &    \textbf{0.688}(0.002) &  0.681(0.007) & 0.669(0.011)\\
        &Cat vs Dog &    \textbf{0.937}(0.004) &  0.932(0.003) & 0.902(0.010)\\
        &Melanoma &   0.804(0.024) &  0.801(0.009) &
        \textbf{0.810}(0.015) \\
        &Cardiomegaly & 0.848(0.005)   &  0.848(0.005)   & \textbf{0.853}(0.009) \\
        PSQ  &HIV &   \textbf{0.767}(0.005) & \textbf{0.767}(0.005) & 0.766(0.002) \\
        &MUV(bioassay) &   0.690(0.030) & \textbf{0.703}(0.030) & 0.702(0.053)\\
        &Tox21(t0) &   \textbf{0.788}(0.005) & \textbf{0.788}(0.005) & 0.752(0.011) \\
        &Tox21(t2) &   0.903(0.002) & \textbf{0.904}(0.001) & 0.893(0.004)\\
        &ToxCast(t8) &   0.444(0.018) & 0.460(0.104) & \textbf{0.553}(0.056) \\
        &ToxCast(t12) &   0.908(0.036) & 0.908(0.047) & \textbf{0.910}(0.040) \\   
        
        \end{tabular}}
       \resizebox{0.85\textwidth}{!}{ \begin{tabular}{llllll}
       \toprule
               Loss &Dataset & SGD-style& Momentum-style& Adam-style & PESG\\
               \midrule
         &STL10 &   0.796(0.013) &  \textbf{0.819}(0.003) & 0.785(0.016)  & 0.766(0.020)\\
        &CIFAR10 &    \textbf{0.910}(0.002) &  0.905(0.006) & 0.845(0.023)  & 0.886(0.007)\\
      &CIFAR100 &    0.678(0.010) &  \textbf{0.684}(0.005) & 0.652(0.007) & 0.664(0.006)\\
        &Cat vs Dog &   \textbf{0.934}(0.006) &  0.930(0.004) & 0.907(0.008) & 0.916(0.006)\\
        &Melanoma &   0.803(0.020) &  \textbf{0.818}(0.004) & 0.812(0.009)  & 0.815(0.016)\\
        &Cardiomegaly & 0.861(0.004)   &  \textbf{0.862}(0.004)  & 0.840(0.012) & 0.861(0.004)\\
        CSQ  &HIV &   0.770(0.003) & 0.770(0.003) & 0.763(0.005) & \textbf{0.775}(0.007) \\
        &MUV(bioassay) &   0.715(0.040) & 0.699(0.026) & 0.684(0.084) & \textbf{0.721}(0.049) \\
        &Tox21(t0) &   0.779(0.002) & 0.777(0.017) & \textbf{0.781}(0.012)  & 0.778(0.003) \\
        &Tox21(t2) &   0.898(0.003) & \textbf{0.899}(0.002) & 0.893(0.007)& 0.898(0.007)\\
        &ToxCast(t8) &   0.445(0.009) & \textbf{0.499}(0.002) & 0.467(0.055)& 0.443(0.006)\\
        &ToxCast(t12) &   0.875(0.084) & 0.883(0.047) & \textbf{0.900}(0.036) & 0.821(0.044)\\
        \bottomrule
    \end{tabular}}
\end{table}
\begin{table}[h]
\caption{Uncertain and missing rates on CheXpert dataset}
    \label{tab:chexpert_uncertain_rate}
    \centering
    \resizebox{0.5\textwidth}{!}{
    \begin{tabular}{lll}
      Target & Uncertain rate & Missing rate\\
      \hline
      Cardiomegaly & 0.031 & 0.722\\ Edema  & 0.063& 0.572\\ Consolidation & 0.121&0.664\\ Atelectasis  &0.157 & 0.684\\ Pleural Effusion &  0.051&  0.410\\
      \bottomrule
    \end{tabular}}
\end{table}
\begin{figure}
\caption{
		Loss functions comparisons on benchmark datasets (C.D for Cat vs Dog, Mel for Melanoma, T.C. for ToxCast, Card. for Cardiomegaly, Conso. for Consolidation, Ate. for Atelectasis, P.E. for Pleural Effusion. The percentages are normalized with the mean test AUROC of each data for all losses.)
    }\label{fig:losses-final}
    \includegraphics[scale=0.35]{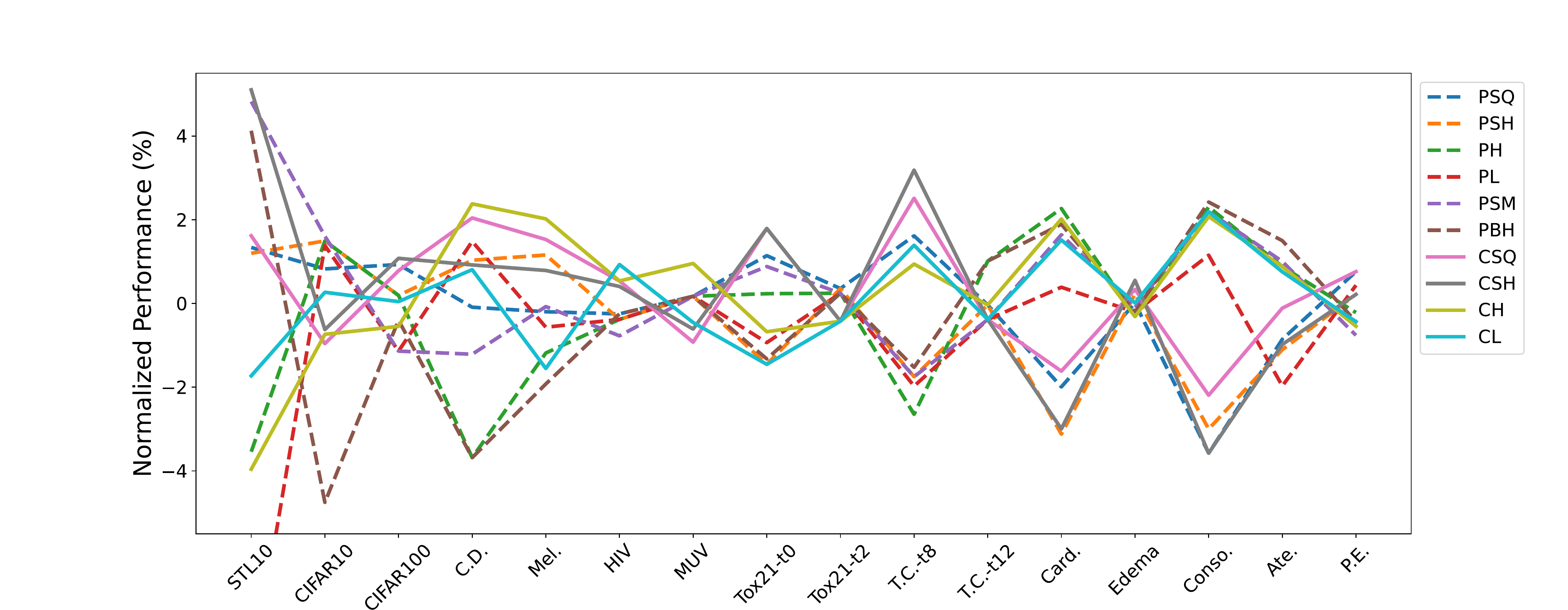}
\end{figure}


As we can see from the results in Table~\ref{tab:te_SPR}, most of the data favors higher SPR than lower SPR. The phenomenon is more obvious on the generated image dataset. The observation indicates over-sampling positive data in the mini-batch  could improve the AUROC optimization performance especially when testing data is more balanced.

{\bf The impact of CER.} For CER, following \citep{yuan2020robust}, we investigate three different values for $\gamma$: 0, 0.002, and 0.02. We fix other hyper-parameters as in the previous SPR setting and fix SPR as 50\%. The results are presented in the Table~\ref{tab:te_CER}. From the results, we can see CER generally improves the generalization performance for the image datasets. However, for some datasets that are more imbalanced, such as MUV(bioassay) with 0.2\% positive rate and ToxCast(t12) with 3.0\% positive rate, large CER could degrade model performance badly.

{\bf The impact of WD.} The weight decay (WD) is tuned at \{1e-1, 1e-2, 1e-3, 1e-4, 0\}. The other hyper-parameters are fixed as in previous CER setting, except we now fix $\gamma$ in CER as 0.002 for image datasets and fix $\gamma$ in CER as 0 for molecule datasets as these two extremely imbalanced datasets don't favor CER. The results are at appendix, as the observations are standard. Weight decay is generally believed to improve model generalization; it should be tuned in practice for different datasets.

{\bf The impact of activation/normalization.} Next, we further fix WD as 1e-4, and tune the activation/normalization function for the output (last) layer as: none, sigmoid, $\ell_1$ score normalization, $\ell_2$ score normalization, and batch normalization. Because $\ell_1$ score normalization and batch-normalization are not so competitive as others, we omit them in the main content and show partial results in Table~\ref{tab:te_activation_partial}, which indicates that normalization is effective. The full results are included at appendix.  We can see that the sigmoid activation and the $\ell_2$ norm score normalization are competitive and the former one wins more cases.

{\bf The impact of optimizers.} Next, we fix last normalization function as sigmoid and conduct experiments on different optimizers from \{SGD-style, Momentum-style, Adam-style\} with different learning rates tuned from \{1e-1, 1e-2, 1e-3, 1e-4, 1e-5\}. The other settings are inherited as previous choices, that is, 50\% for SPR, 0.002 for CER and 1e-4 for weight decay. The results are presented in Table~\ref{tab:te_opt}. We also include the best learning rate with the optimizers at the appendix. The SGD and Momentum method perform well for all datasets; Adam optimizer doesn't enjoy good testing performance, although it enjoys faster empirical training convergence speed, which we will elaborate shortly in subsection~\ref{sec:num_opt}.

We also include the performance of the method PESG for solving the CSQ in the min-max form as used in~\citep{yuan2020robust}. It is notable that PESG uses a primal-dual stochastic gradient style update for solving the min-max form. The SGD-style optimizer for solving the composite form is equivalent to a primal-dual style update according to~\citep{Zhang2020OptimalAF}. Neverthess, we can see that the SGD-style update for solving the composite form is better than PESG in more cases. This is probably due to another key difference between them, which is that  PESG adopts random sampling while other three optimizers adopts the oversampling strategy. The observations are consistent with the previous observation for CSQ loss for SPR (sampling positive rate) study Table~\ref{tab:te_SPR}. For example, MUV, HIV and Cardiomegaly don't necessarily require oversampling, where PESG enjoys better performance.

\subsection{Pairwise Losses vs Composite Losses}
Now that we get prior knowledge for the hyper-parameters, we conduct the final comparison for all loss functions of pairwise loss and composite loss classes. {For all datasets except CheXpert, we choose the best hyper-parameters by validation AUROC produced by the above simplified sequential hyper-parameter tuning steps (SPR$\rightarrow$CER$\rightarrow$WD$\rightarrow$normalization/activation\\$\rightarrow$optimizer) for PSQ and CSQ losses, whose details are included in the appendix. The results are presented at Figure~\ref{fig:losses-final}.}
From the results, although there is no clear winner, composite loss outperforms classical pairwise loss on most of the datasets.

{For all the targets at CheXpert dataset, we choose the hyper-parameters by prior knowledge on the other benchmark datasets, namely, SPR=50\%, CER with $\gamma=0.002$, WD=1e-4, last activation as $\ell_2$-normalization, momentum-style optimizer with a learning rate of 1e-3. The choices are also close to the best choices based on the validation results for CheXpert Cardiomegaly under the simplified sequential tuning process.} The results for CheXpert dataset are also presented at Figure~\ref{fig:losses-final}. We include the detailed numerical tabular results for all benchmark datasets at Table~\ref{tab:final_img},~\ref{tab:final_mole},~\ref{tab:final_chexpert}.

All the five targets for CheXpert have uncertain labels and missing labels. The rates are shown in Table~\ref{tab:chexpert_uncertain_rate}.
Consolidation and Atelectasis are the two targets with highest uncertain rate; and they also have high missing rate. From the results in Table~\ref{tab:final_chexpert} in the appendix, the pairwise barrier hinge loss achieves the best performance and pairwise sigmoid loss also achieves good performance, which confirms the point that a symmetric loss is better under the learning with corrupted labels~\citep{charoenphakdee2019symmetric}.

\begin{figure}[h!]
    \centering
     \begin{subfigure}[b]{0.3\textwidth}
         \centering
         \includegraphics[width=\textwidth]{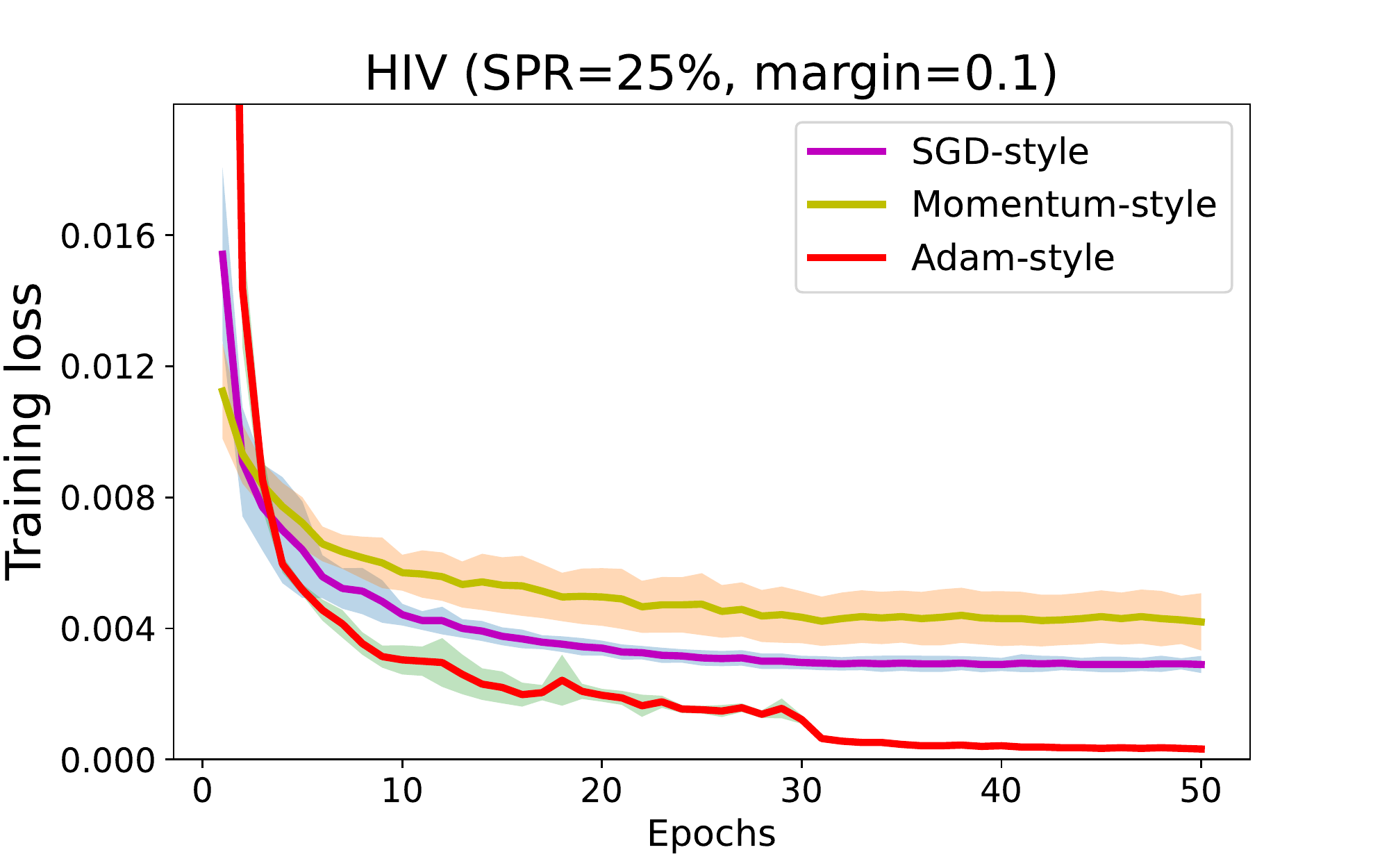}
         \label{fig:stl10-opt}
     \end{subfigure}
     \begin{subfigure}[b]{0.3\textwidth}
         \centering
         \includegraphics[width=\textwidth]{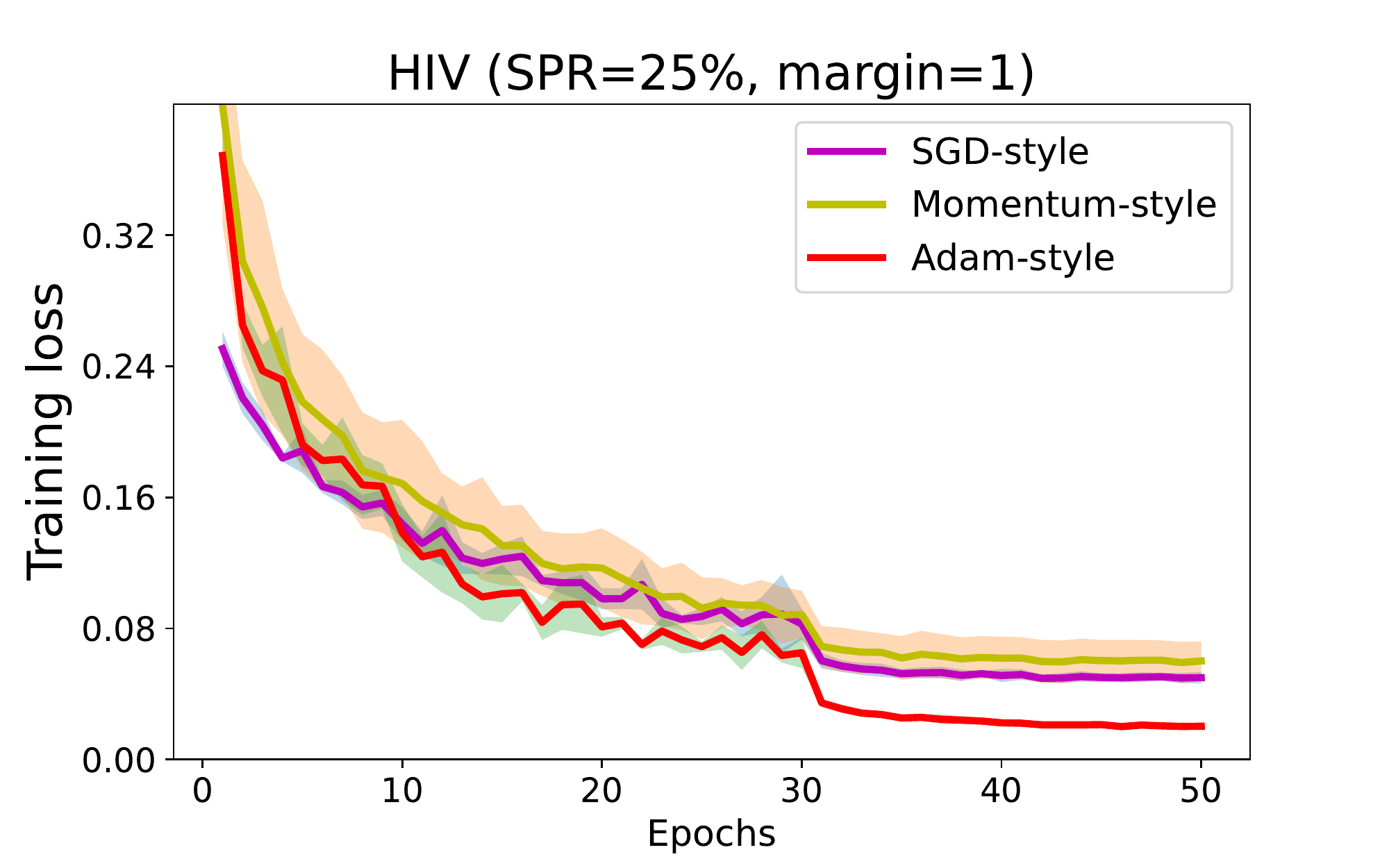}
         \label{fig:stl10-opt}
     \end{subfigure}
     \begin{subfigure}[b]{0.3\textwidth}
         \centering
         \includegraphics[width=\textwidth]{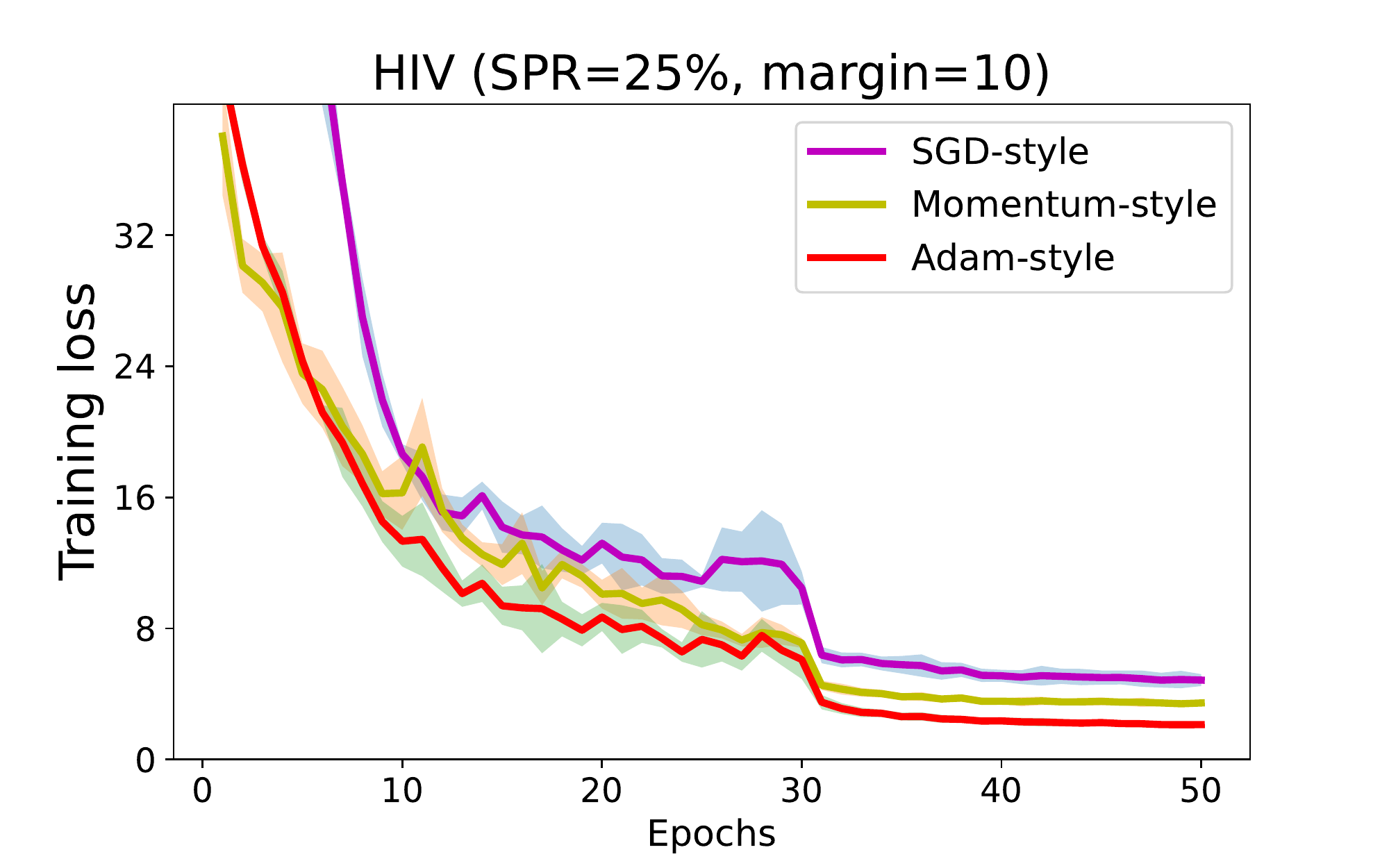}
         \label{fig:stl10-opt}
     \end{subfigure}
     \begin{subfigure}[b]{0.3\textwidth}
         \centering
         \includegraphics[width=\textwidth]{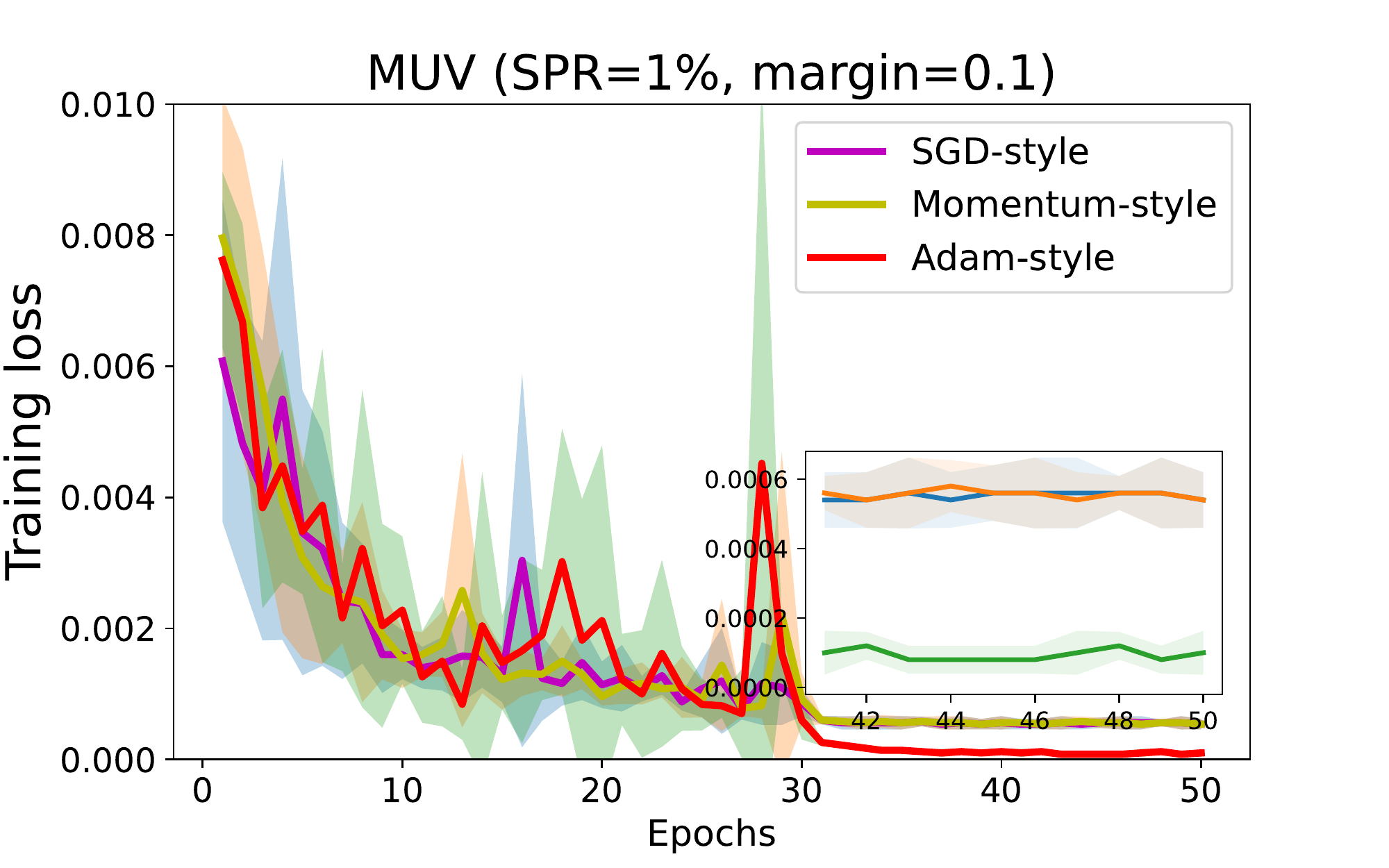}
         \label{fig:stl10-opt}
     \end{subfigure}
     \begin{subfigure}[b]{0.3\textwidth}
         \centering
         \includegraphics[width=\textwidth]{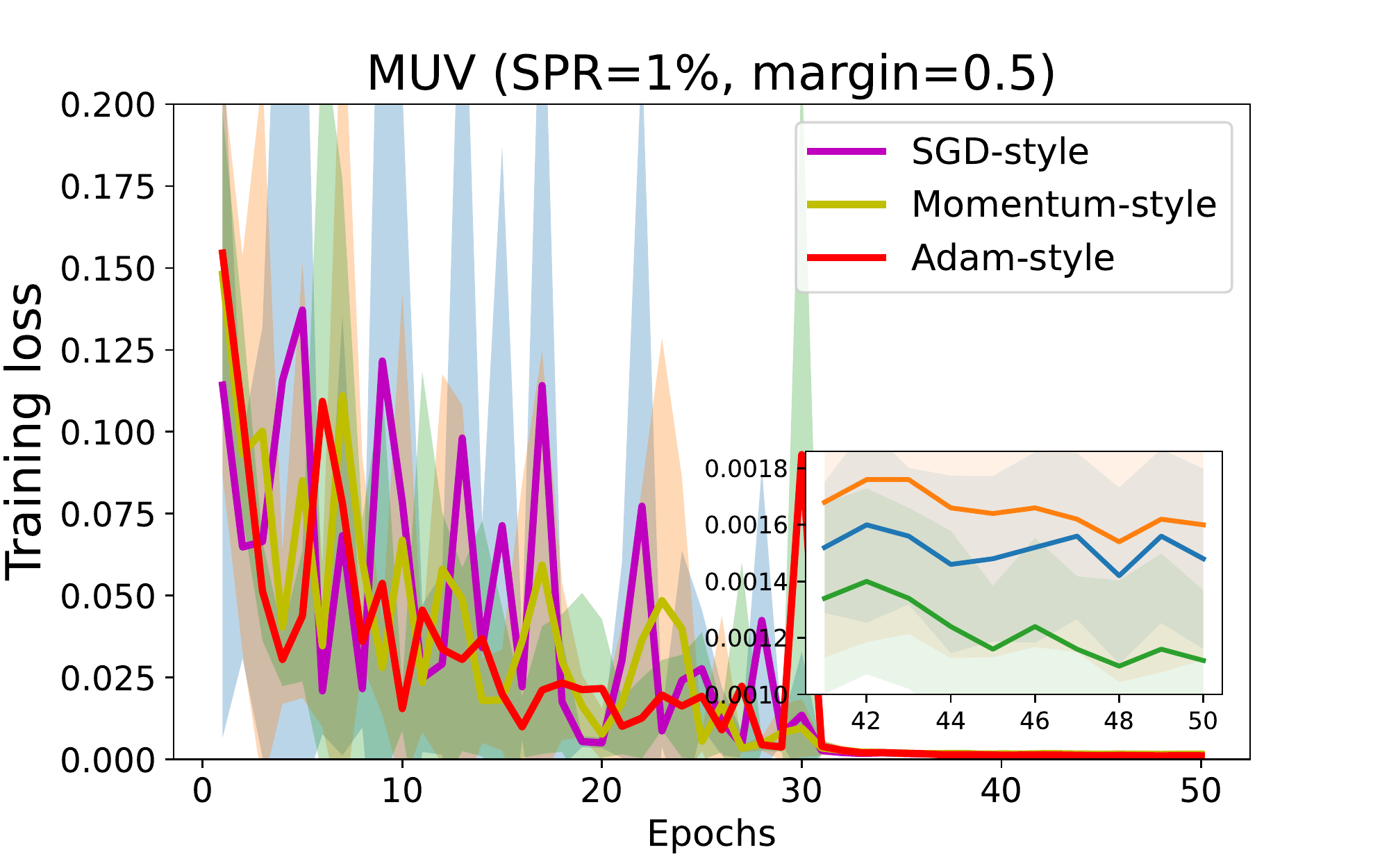}
         \label{fig:stl10-opt}
     \end{subfigure}
     \begin{subfigure}[b]{0.3\textwidth}
         \centering
         \includegraphics[width=\textwidth]{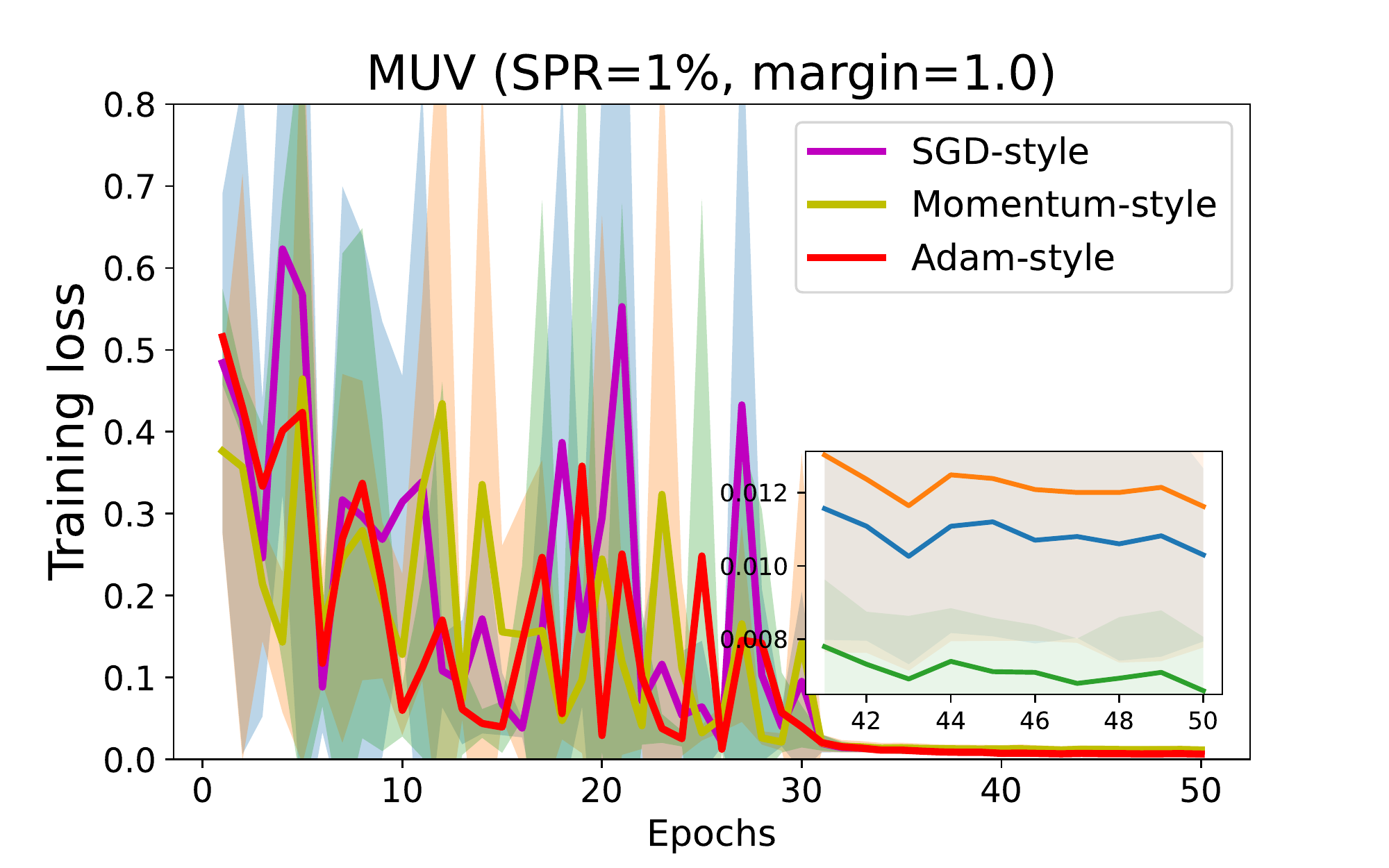}
         \label{fig:stl10-opt}
     \end{subfigure}
     \begin{subfigure}[b]{0.3\textwidth}
         \centering
         \includegraphics[width=\textwidth]{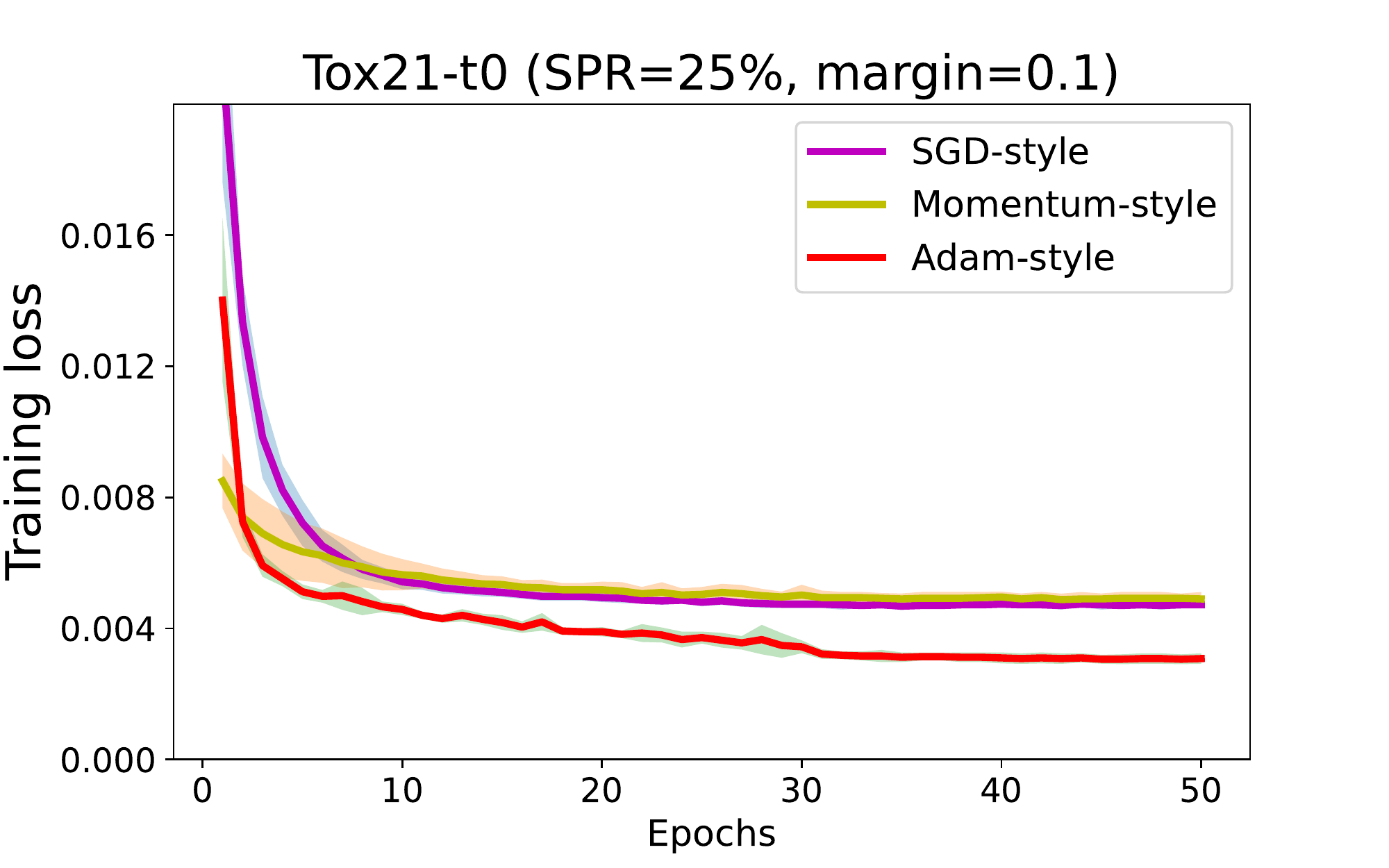}
         \label{fig:stl10-opt}
     \end{subfigure}
     \begin{subfigure}[b]{0.3\textwidth}
         \centering
         \includegraphics[width=\textwidth]{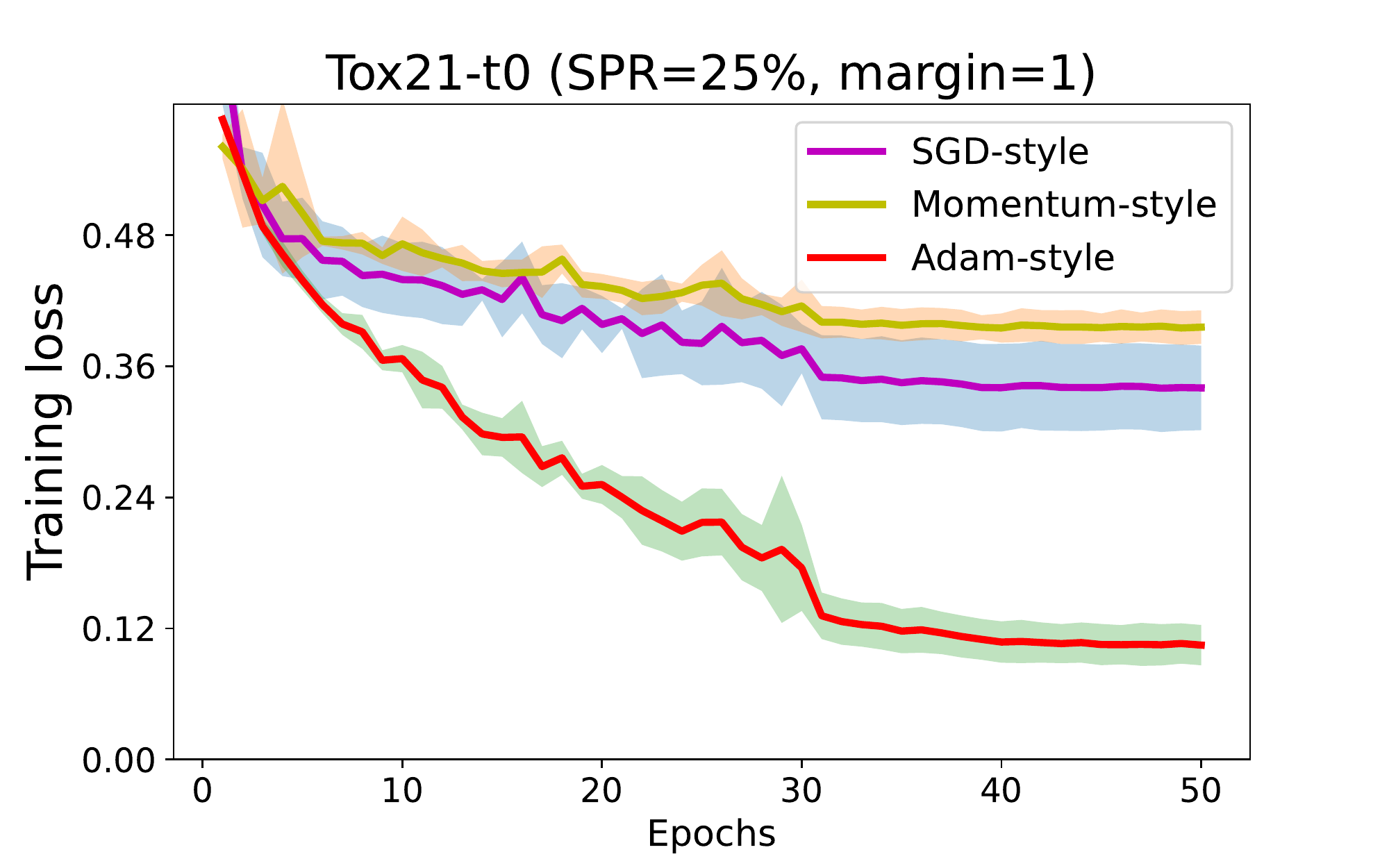}
         \label{fig:stl10-opt}
     \end{subfigure}
     \begin{subfigure}[b]{0.3\textwidth}
         \centering
         \includegraphics[width=\textwidth]{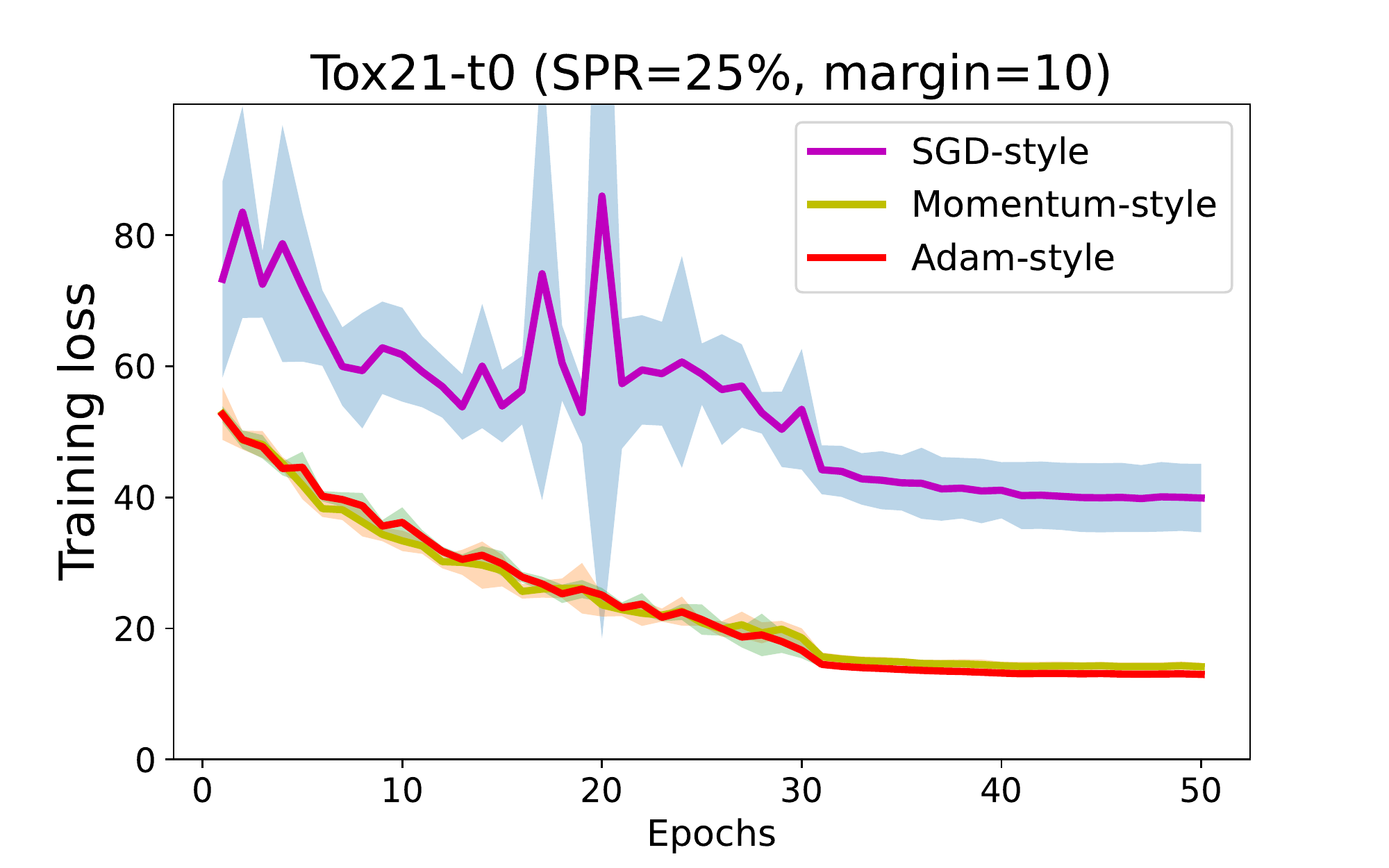}
         \label{fig:stl10-opt}
     \end{subfigure}
    \caption{Different Optimizers for Composite Squared Hinge (CSH) Loss on Benchmark Datasets}
    \label{fig:training_CSH}
\end{figure}

\begin{table}[h!]
    \centering
    \caption{Different loss functions on image datasets}
    \label{tab:final_img}
    \resizebox{1.0\textwidth}{!}{
    \begin{tabular}{lllllll}
      & Dataset &  STL10& CIFAR10& CIFAR100& Cat vs Dog& Melanoma\\
        \hline
          &PSQ &  0.727(0.031) & 0.904(0.004)  & 0.683(0.003) & 0.890(0.011)  &0.810(0.013) \\
          &PSH &  0.726(0.031) & 0.910(0.003)  &0.678(0.005)   & 0.900(0.002)  & 0.821(0.014)  \\
         Pairwise &PH &  0.692(0.056)  & 0.910(0.001)  &0.678(0.008) & 0.858(0.006) & 0.802(0.015) \\
         Loss &PL & 0.653(0.070)  & 0.909(0.012)  & 0.669(0.013)   & 0.904(0.012) & 0.807(0.016) \\
          &PSM &  0.752(0.025) & \textbf{0.911}(0.004) & 0.669(0.014) & 0.880(0.011)  & 0.811(0.014) \\
          &PBH & 0.747(0.017)  &  0.854(0.019) &0.674(0.014)  & 0.858(0.023) & 0.796(0.017)   \\
          \midrule
           &CSQ & 0.729(0.020)  &  0.888(0.014) & 0.682(0.009)  & 0.909(0.011) & \textbf{0.824}(0.011) \\
         Composite &CSH &  \textbf{0.754}(0.033)   &  0.891(0.006)  & \textbf{0.684}(0.005) & 0.899(0.007) &0.818(0.013)  \\
         Loss &CH & 0.689(0.052)  & 0.890(0.016) & 0.673(0.024)  & \textbf{0.912}(0.009)  & 0.828(0.002) \\
          &CL &  0.705(0.062) & 0.899(0.004)  & 0.677(0.012) & 0.898(0.010) & 0.799(0.015) \\
        \bottomrule
    \end{tabular}}
\end{table}
\begin{table}[h!]
\caption{Different loss functions on molecular datasets}
    \label{tab:final_mole}
    \centering
    \resizebox{1.0\textwidth}{!}{
    \begin{tabular}{llllllll}
      & Dataset &  HIV& MUV(bioassay)& Tox21(t0)& Tox21(t2)& ToxCast(t8) & ToxCast(t12)\\
        \hline
          &PSQ &   0.761(0.001) & 0.639(0.001) & 0.779(0.004) & \textbf{0.901}(0.002) & 0.453(0.027)& 0.573(0.016) \\
          &PSH &  0.760(0.002) & 0.639(0.001) & 0.759(0.002) & \textbf{0.901}(0.002) & 0.438(0.013)& 0.573(0.016) \\
         Pairwise &PH &   0.760(0.002) & 0.639(0.001) & 0.772(0.002) &0.900(0.001) & 0.434(0.003)& \textbf{0.579}(0.008) \\
         Loss &PL &  0.760(0.003) & 0.639(0.001) & 0.763(0.012) & 0.900(0.001) & 0.437(0.008)& 0.571(0.018) \\
          &PSM &   0.757(0.001) & 0.639(0.001) & 0.777(0.001) & 0.900(0.001) & 0.438(0.003)& 0.571(0.018) \\
          &PBH &  0.761(0.002) & 0.639(0.001) & 0.760(0.008) & 0.900(0.003) & 0.439(0.011)& \textbf{0.579}(0.014) \\
          \midrule
           &CSQ & 0.767(0.002) & 0.632(0.008) & \textbf{0.784}(0.007) & 0.894(0.001) & 0.457(0.012)& 0.571(0.017) \\
         Composite &CSH &  0.766(0.002) & 0.634(0.008) & \textbf{0.784}(0.007) & 0.894(0.001) & \textbf{0.460}(0.011)& 0.571(0.017)  \\
         Loss &CH & 0.767(0.007) & \textbf{0.644}(0.008) & 0.765(0.004) & 0.894(0.001) & 0.450(0.005)& 0.573(0.016) \\
          &CL &  \textbf{0.770}(0.004) & 0.635(0.009) & 0.759(0.001) & 0.894(0.002) & 0.452(0.005)& 0.571(0.017) \\
        \bottomrule
    \end{tabular}}
\end{table}
\begin{table}[h!]
\caption{Different loss functions on CheXpert dataset}
    \label{tab:final_chexpert}
    \centering
    \resizebox{1.0\textwidth}{!}{
    \begin{tabular}{lllllll}
      & Dataset & Cardiomegaly(t0) & Edema(t1)  & Consolidation(t2) &  Atelectasis(t3)  & Pleural Effusion(t4) \\
        \hline
          &PSQ &    0.782(0.011) &  0.922(0.006)  & 0.836(0.016) &  0.800(0.007)  & \textbf{0.927}(0.002) \\
          &PSH &     0.773(0.025) &  0.924(0.004)  & 0.841(0.028) &  0.798(0.021)  & 0.922(0.007) \\
         Pairwise &PH &     \textbf{0.816}(0.006) &  0.919(0.002)  & 0.887(0.010) &  0.814(0.004)  & 0.918(0.002) \\
         Loss &PL &   0.801(0.005) &  0.920(0.003)  & 0.877(0.016) &  0.791(0.002)  & 0.924(0.001) \\
          &PSM &     0.811(0.005) &  0.921(0.001)  & 0.886(0.013) &  0.815(0.005)  & 0.913(0.003) \\
          &PBH &    0.813(0.005) &  0.920(0.001)  & \textbf{0.888}(0.009) &  \textbf{0.819}(0.003)  & 0.916(0.001) \\
          \midrule
           &CSQ &    0.785(0.013) &  0.925(0.004)  & 0.848(0.022) &  0.806(0.006)  & \textbf{0.927}(0.005) \\
         Composite &CSH &     0.774(0.025) &  \textbf{0.927}(0.003)  & 0.836(0.019) &  0.799(0.015)  & 0.922(0.007) \\
         Loss &CH &   0.814(0.004) &  0.919(0.001)  & 0.885(0.010) &  0.814(0.004)  & 0.915(0.002) \\
          &CL &    0.810(0.005) &  0.922(0.001)  & 0.886(0.008) &  0.813(0.004)  & 0.916(0.002) \\
        \bottomrule
    \end{tabular}}
\end{table}

\subsection{Numerical Analysis of Optimizers}\label{sec:num_opt}
We compare  different styles of optimizers for optimizing CSH loss, namely PESG~\citep{yuan2020robust}, the momentum-style~\citep{ghadimi2020nasa}, and the Adam-style~\citep{wang2021momentum}. The learning rates are tuned from \{1e-1, 1e-2, 1e-3, 1e-4, 1e-5\}. Partial training convergence curves are presented in Figure~\ref{fig:training_CSH}. Full results are included in the appendix. The results show that the Adam-style optimizer is more competitive than other optimizers for solving the composite loss from the training perspective. The observation is especially strong for the molecule graph datasets. However, as we check the testing performance in Table~\ref{tab:te_opt}, the Adam-style optimizer does not show better testing performance. The findings confirm that adaptive optimizer doesn't have good generalization performance from the previous research~\citep{wilson2017marginal}. We also compare the training convergence for composite loss v.s. pairwise loss at appendix. Overall, composite loss is more competitive for training convergence versus iteration number. Moreover, composite loss consumes smaller running time for each iteration.

\section{Limitations and Conclusions}
The limitation for this study is that we did not exhaustively search all the settings of different algorithmic factors for deep AUROC optimization due to the computational cost. However, based on our carefully designed experiments, we gain three insights for deep AUROC maximization from the perspective of loss functions, algorithmic choices and optimizers: (i) composite loss performs well and  data with corrupted labels favors a symmetric pairwise loss; (2) some algorithmic choices such as over-sampling and normalization are helpful for deep AUROC maximization; (iii) Adam-type optimizer performs well on training data but not testing data. We  expect these findings will be useful for future research and development on deep AUROC maximization.

\vskip 0.2in
\bibliography{aucbenchmark}

\newpage
\appendix
\section{More Experiments for Test AUROC}
\subsection{Weight Decay}
All test AUROC results for different weight decays are included at Table~\ref{tab:te_WD}.
\subsection{Normalization for Last Layer}
All test AUROC results for normalization functions of last layer are included at Table~\ref{tab:te_activation}.
\subsection{Optimizers with Best Learning Rates}
All test AUROC results for optimiers (SGD, Momentum, Adam type) and their learning rates (from \{1e-1,1e-2,1e-3,1e-4,1e-5\}) are included at Table~\ref{tab:te_opt_and_lrs}.
\subsection{Test AUROC for Loss Functions}
The final comprehensive evaluation of all loss functions for image dataset, molecule dataset and specifically CheXpert dataset are included at Table~\ref{tab:final_img}, Table~\ref{tab:final_mole} and Table~\ref{tab:final_chexpert}.
\section{More Experiments for Validation AUROC}
\subsection{Sampling positive rate}
We include the sampling positive rate with validation AUROC at Table~\ref{tab:val-spr}. Similar with test AUROC, over-sampling also improve validation AUROC.

\subsection{Consecutive Epoch Regularization}
We include the consecutive epoch regularization with validation AUROC at Table~\ref{tab:val-cer}, where we see CER could improve validation AUROC as well.

\subsection{Weight Decay}
We include the weight decay with validation AUROC at Table~\ref{tab:val-wd}. Although most of the datasets favors WD with medium value for Validation AUROC. It is better to tune it for different datasets in practice.

\subsection{Normalization for Last Layer}
We include the normalization of last layer with validation AUROC at Table~\ref{tab:val-norm}. Sigmoid function is the most competitive normalization function, which is also consistant with the test AUROC results.

\subsection{Optimizers with Best Learning Rates}
We include the optimizers performance with their best learning rates for validation AUROC at Table~\ref{tab:val-opt}. As the test AUROC, Adam-type of optimizer doesn't perform well for validation AUROC either.

\section{More Experiments for Training Convergence}
\subsection{Optimizers Comparison for CSH Loss}
\begin{figure}[h]
    \centering
    \begin{subfigure}[b]{0.3\textwidth}
         \centering
         \includegraphics[width=\textwidth]{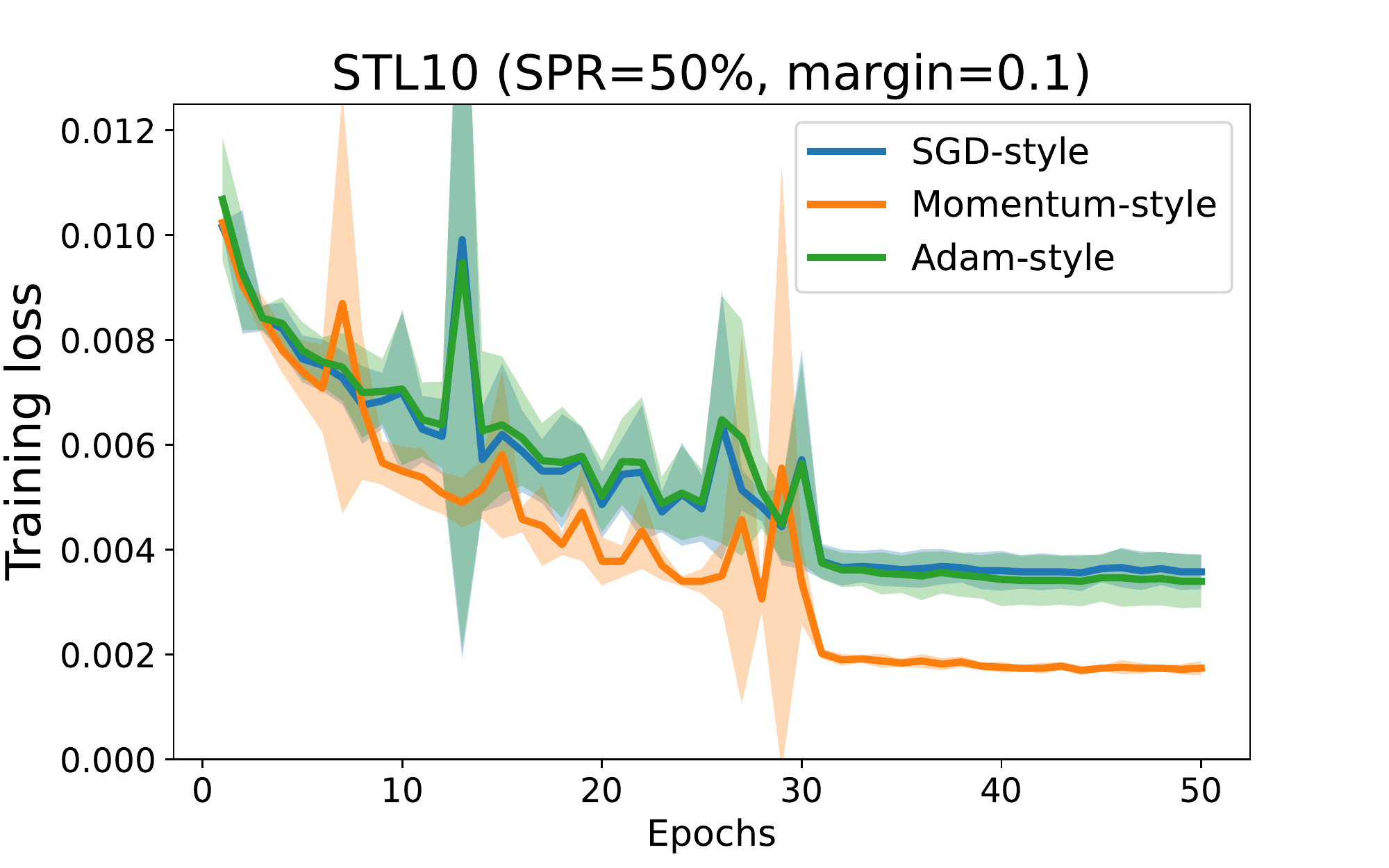}
         \label{fig:stl10-opt}
     \end{subfigure}
     \begin{subfigure}[b]{0.3\textwidth}
         \centering
         \includegraphics[width=\textwidth]{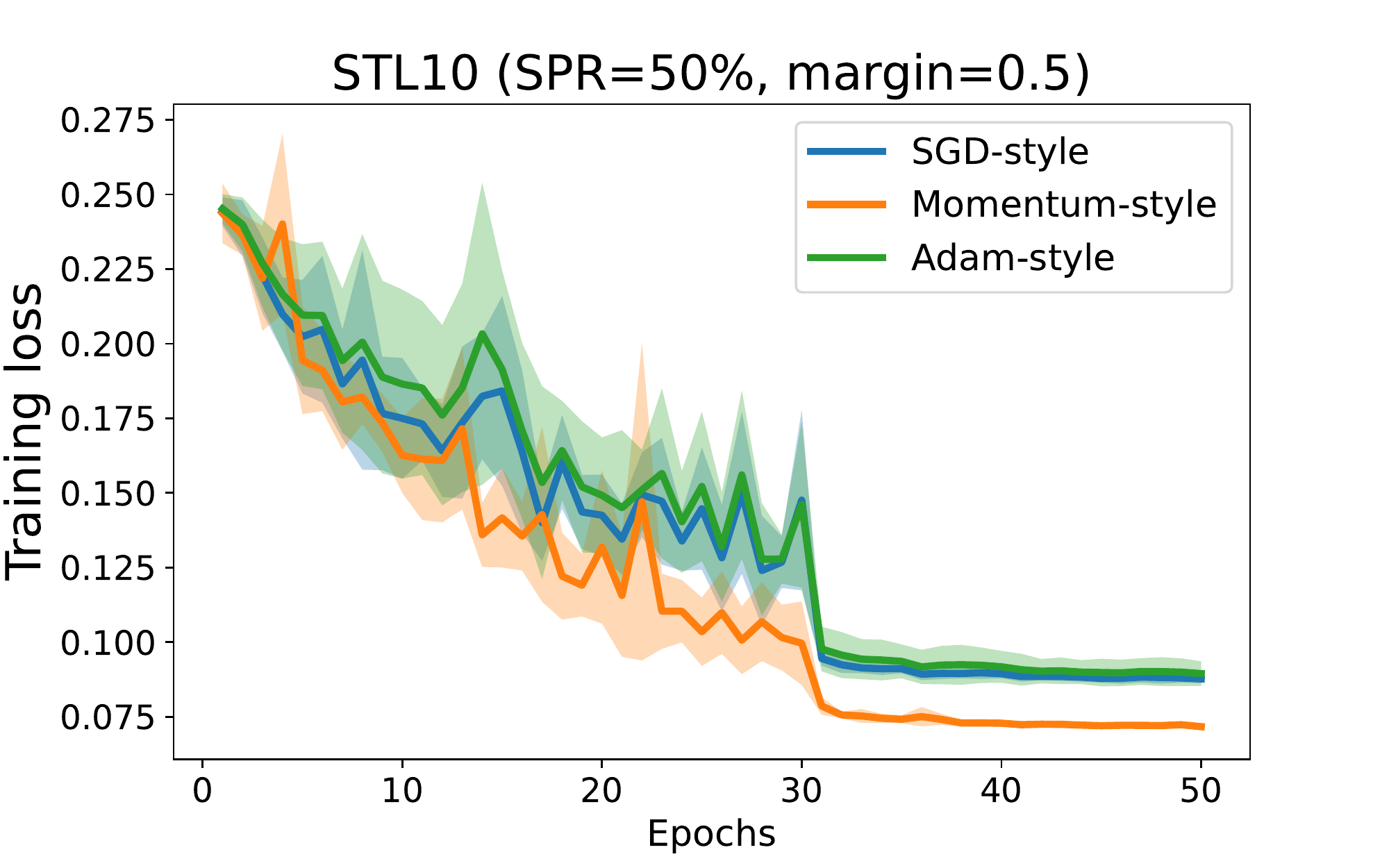}
         \label{fig:stl10-opt}
     \end{subfigure}
     \begin{subfigure}[b]{0.3\textwidth}
         \centering
         \includegraphics[width=\textwidth]{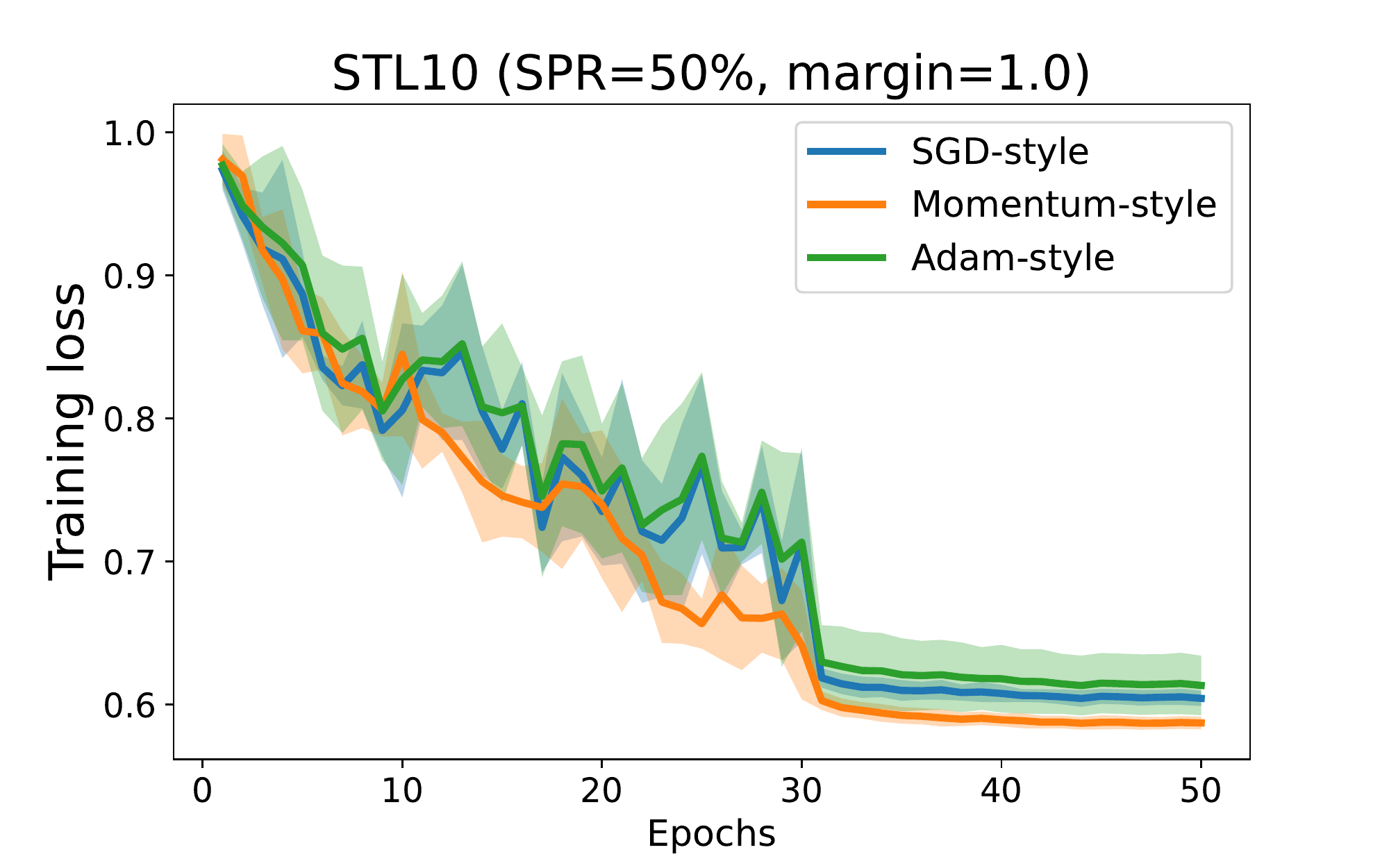}
         \label{fig:stl10-opt}
     \end{subfigure}
     \begin{subfigure}[b]{0.3\textwidth}
         \centering
         \includegraphics[width=\textwidth]{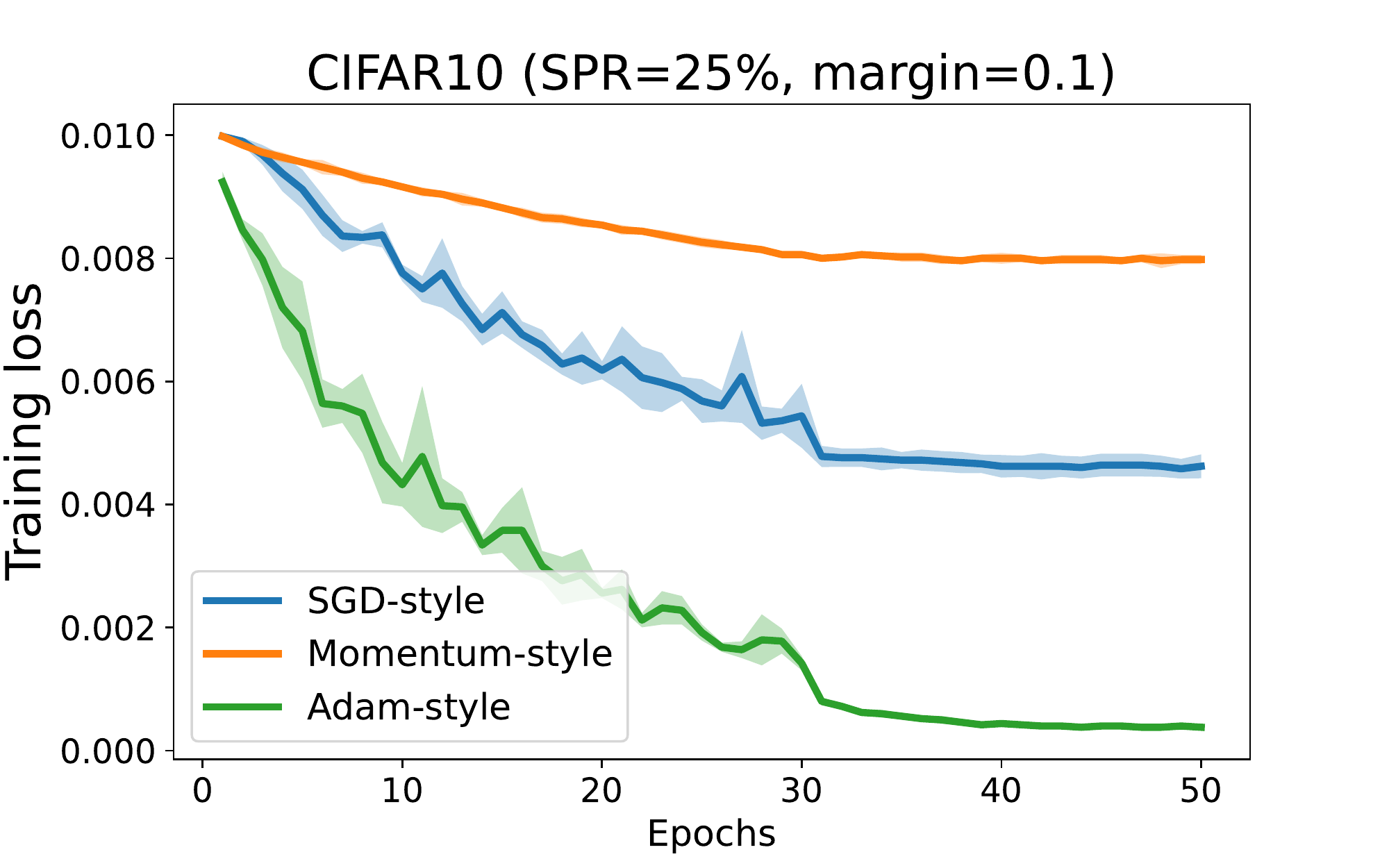}
         \label{fig:stl10-opt}
     \end{subfigure}
     \begin{subfigure}[b]{0.3\textwidth}
         \centering
         \includegraphics[width=\textwidth]{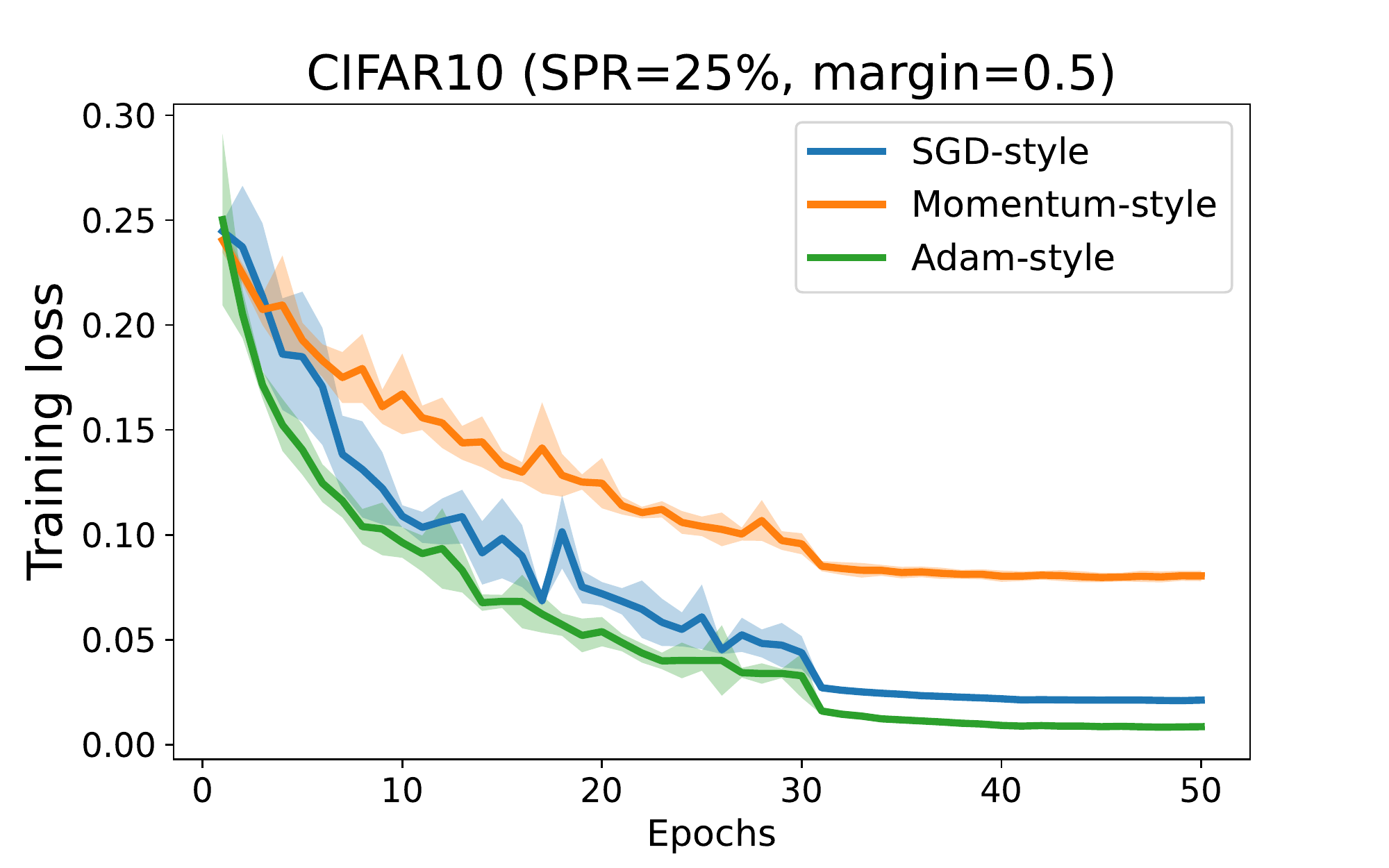}
         \label{fig:stl10-opt}
     \end{subfigure}
     \begin{subfigure}[b]{0.3\textwidth}
         \centering
         \includegraphics[width=\textwidth]{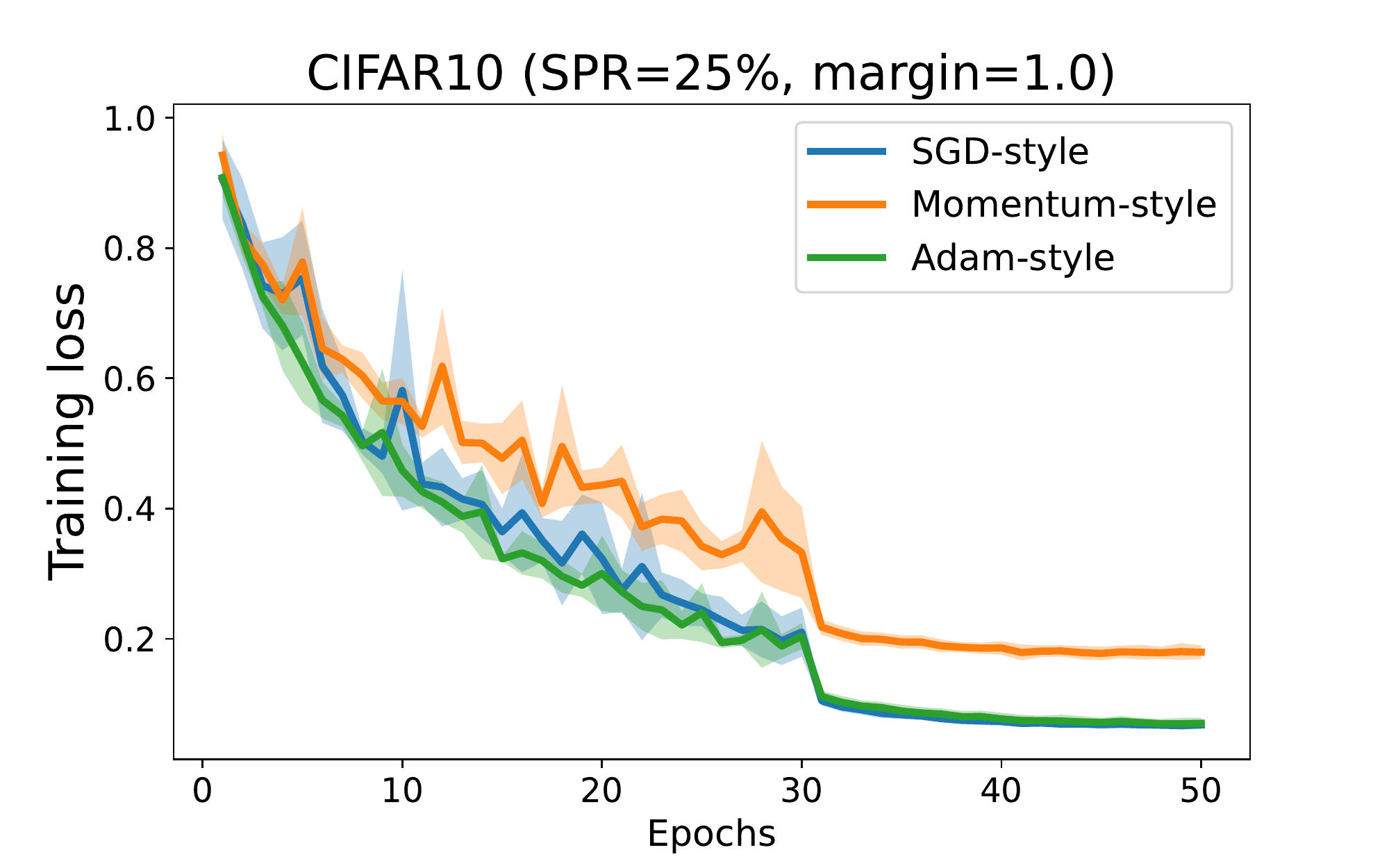}
         \label{fig:stl10-opt}
     \end{subfigure}
     \begin{subfigure}[b]{0.3\textwidth}
         \centering
         \includegraphics[width=\textwidth]{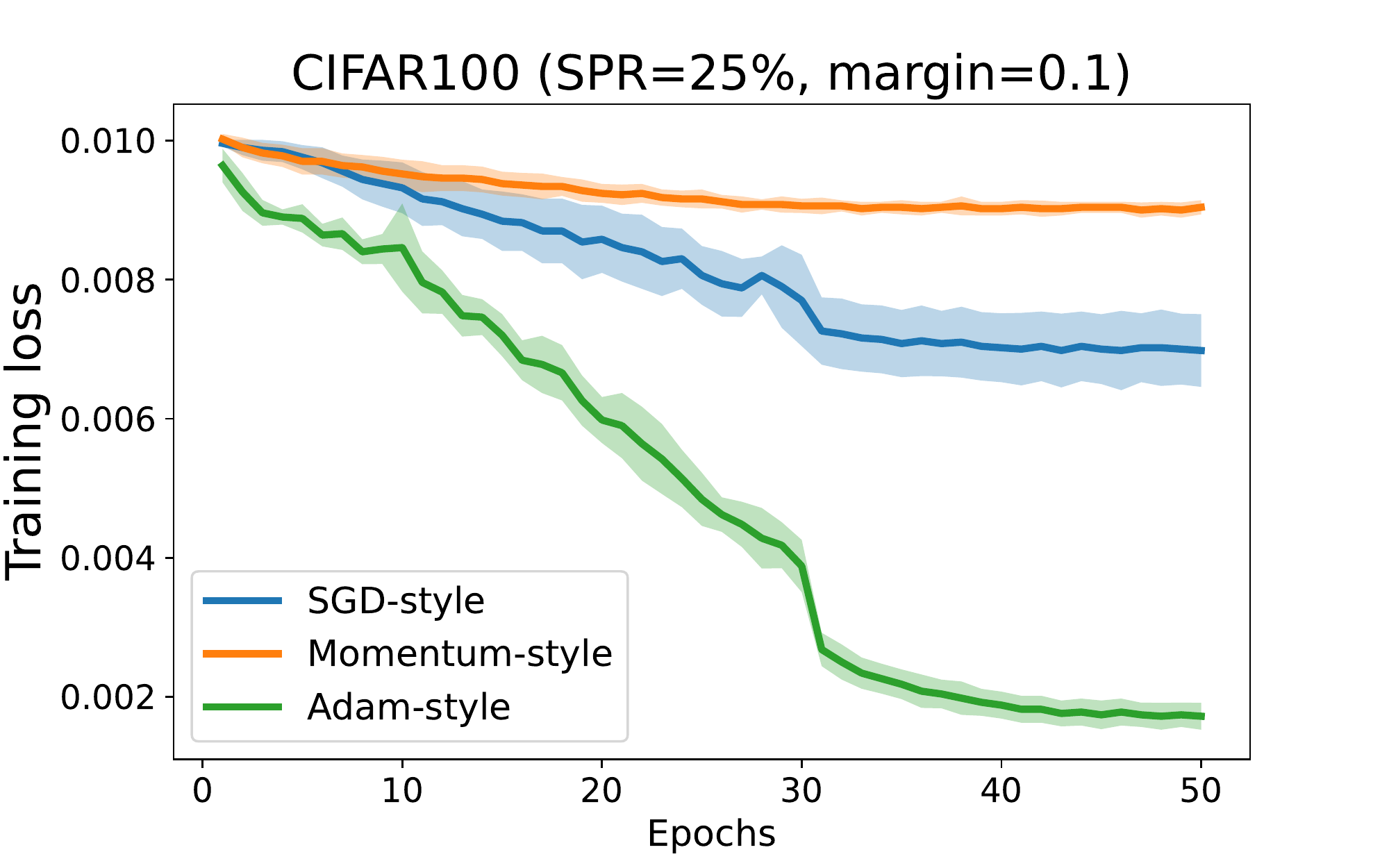}
         \label{fig:stl10-opt}
     \end{subfigure}
     \begin{subfigure}[b]{0.3\textwidth}
         \centering
         \includegraphics[width=\textwidth]{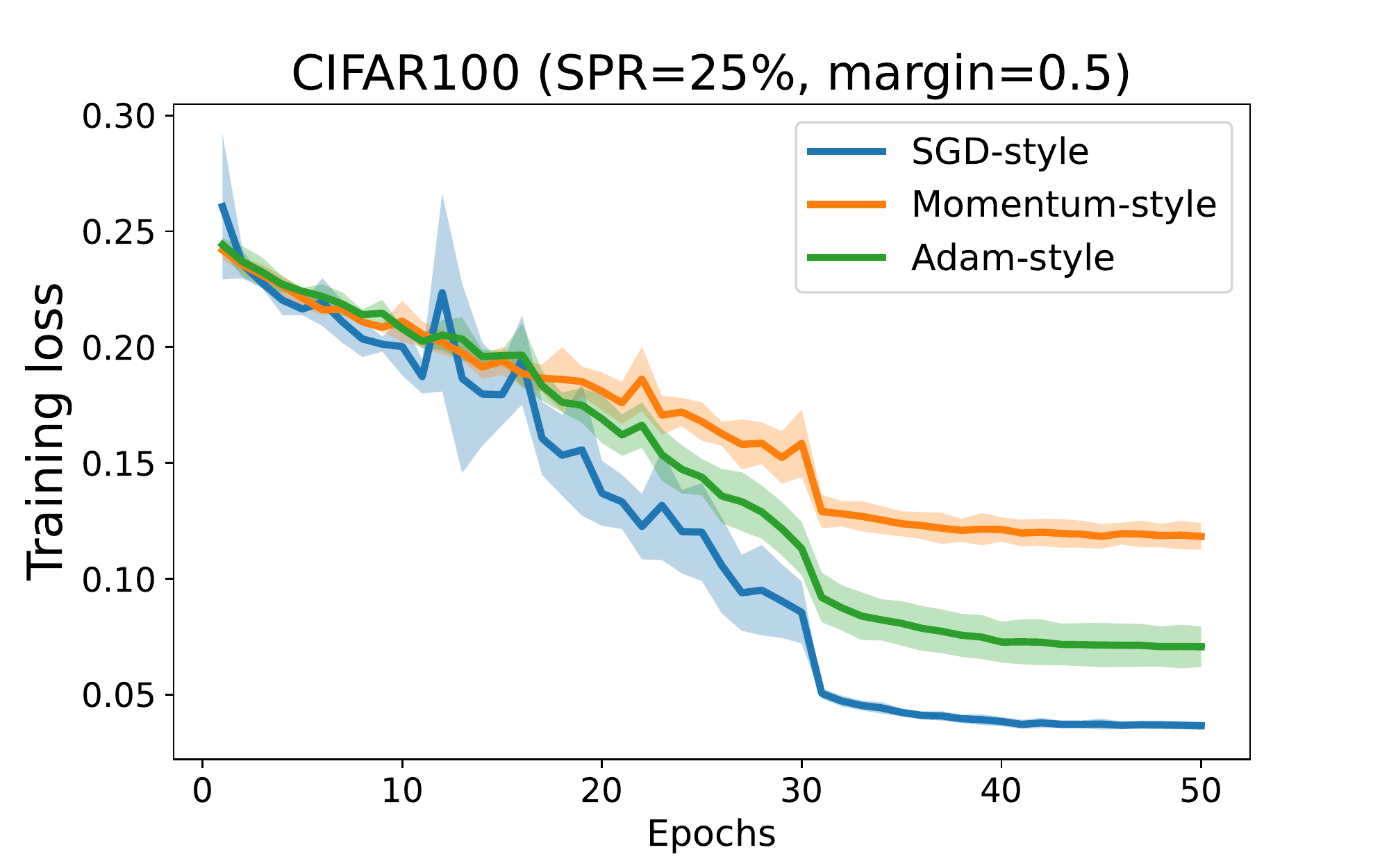}
         \label{fig:stl10-opt}
     \end{subfigure}
     \begin{subfigure}[b]{0.3\textwidth}
         \centering
         \includegraphics[width=\textwidth]{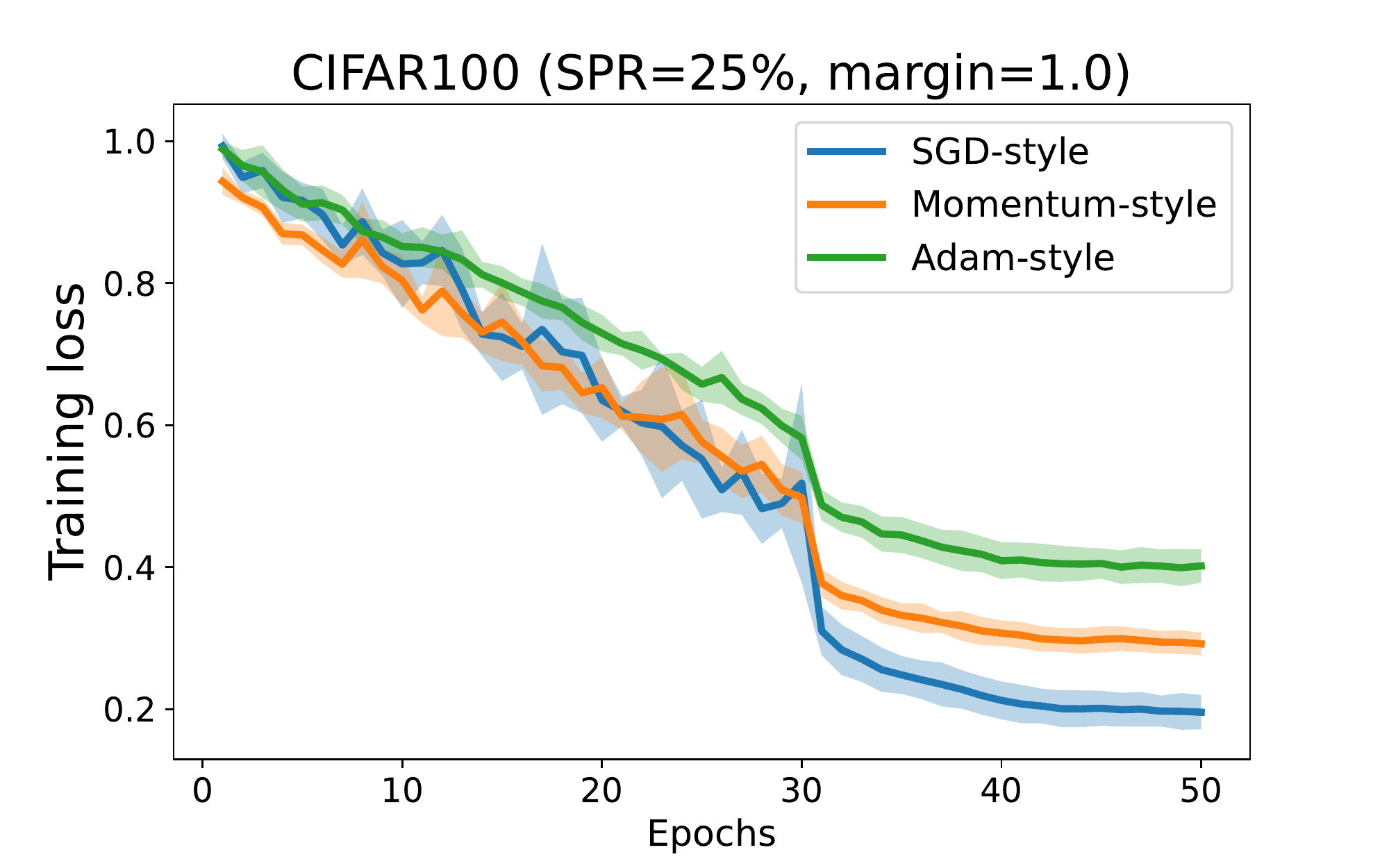}
         \label{fig:stl10-opt}
     \end{subfigure}
     \begin{subfigure}[b]{0.3\textwidth}
         \centering
         \includegraphics[width=\textwidth]{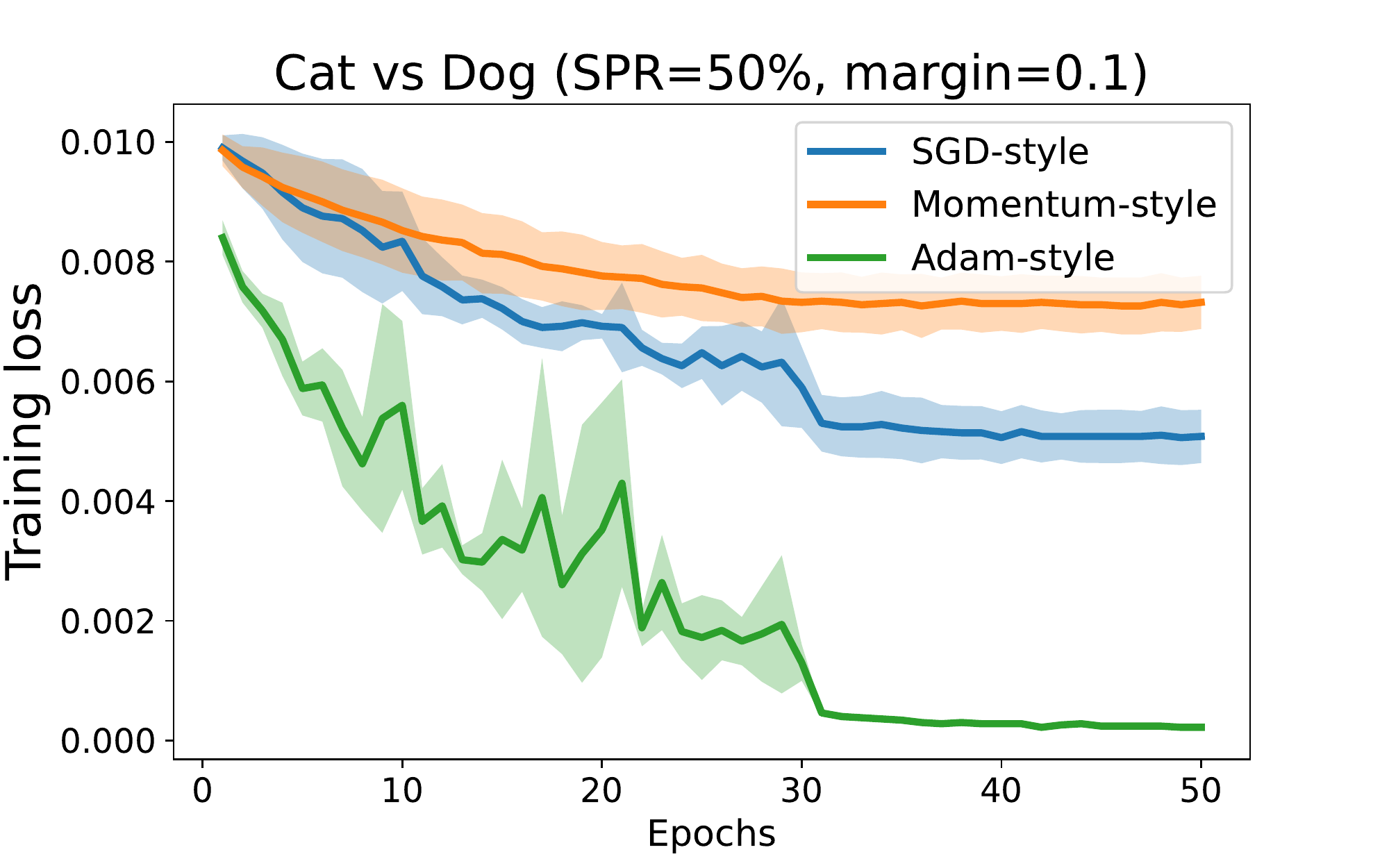}
         \label{fig:stl10-opt}
     \end{subfigure}
     \begin{subfigure}[b]{0.3\textwidth}
         \centering
         \includegraphics[width=\textwidth]{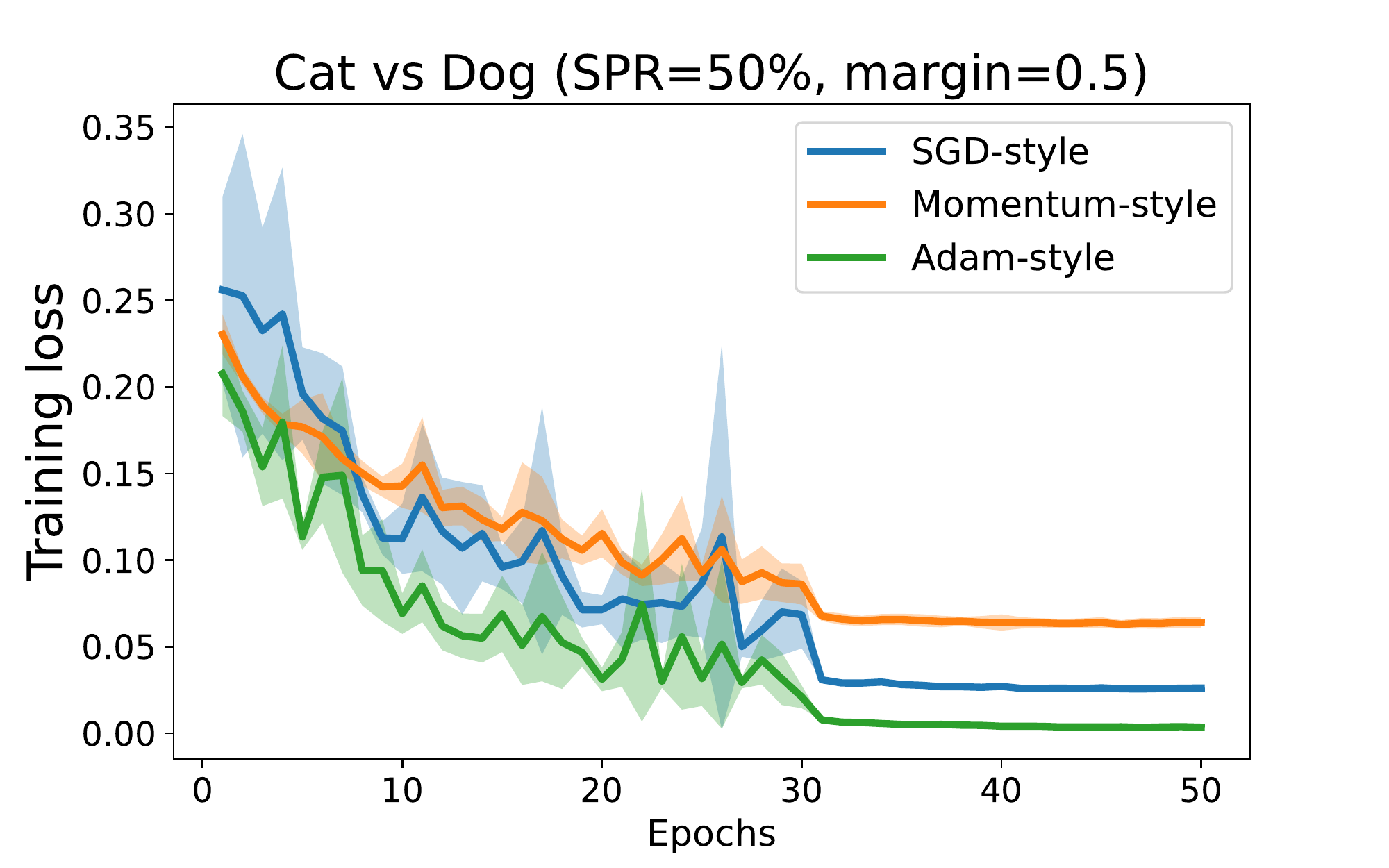}
         \label{fig:stl10-opt}
     \end{subfigure}
     \begin{subfigure}[b]{0.3\textwidth}
         \centering
         \includegraphics[width=\textwidth]{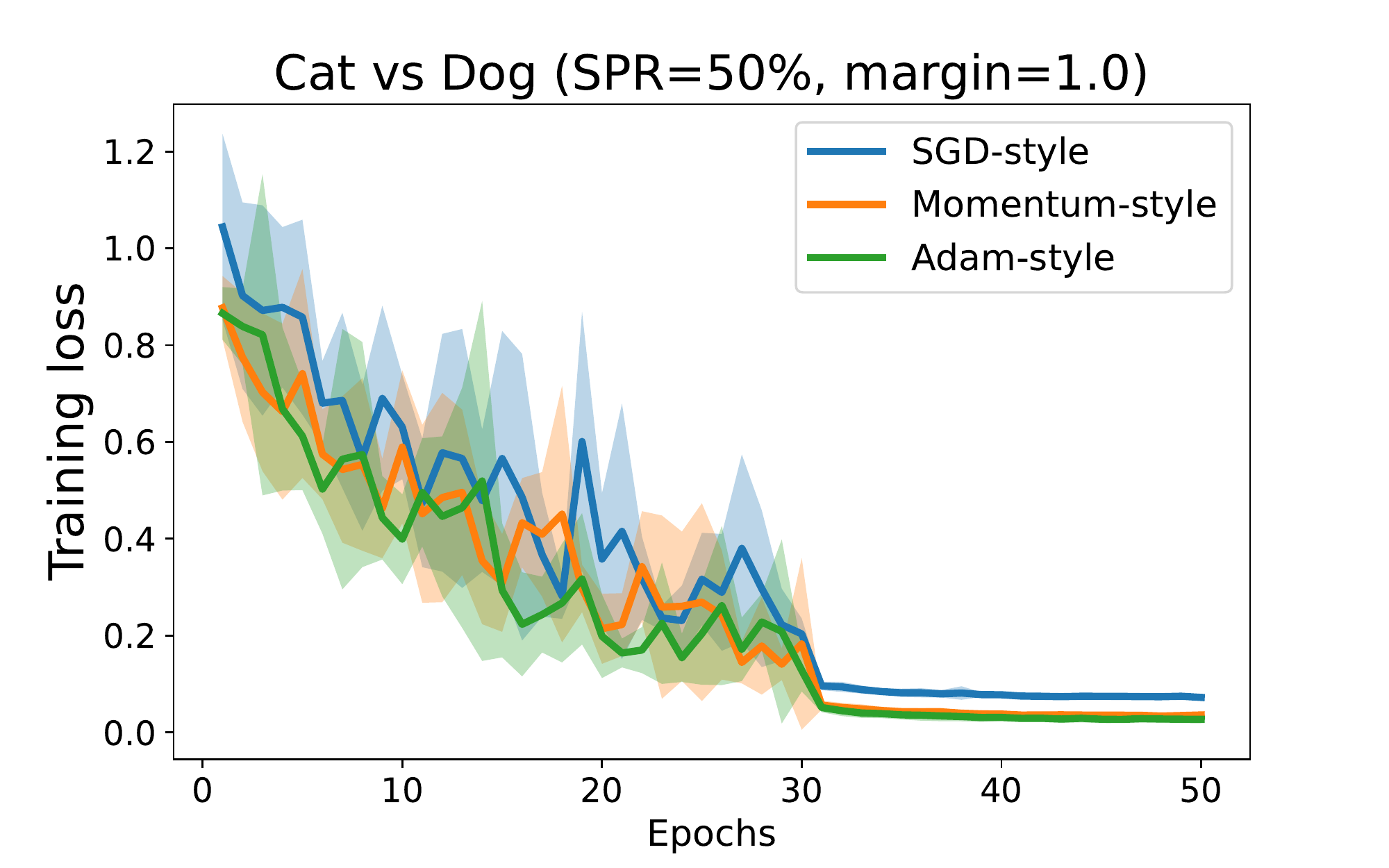}
         \label{fig:stl10-opt}
     \end{subfigure}
     \begin{subfigure}[b]{0.3\textwidth}
         \centering
         \includegraphics[width=\textwidth]{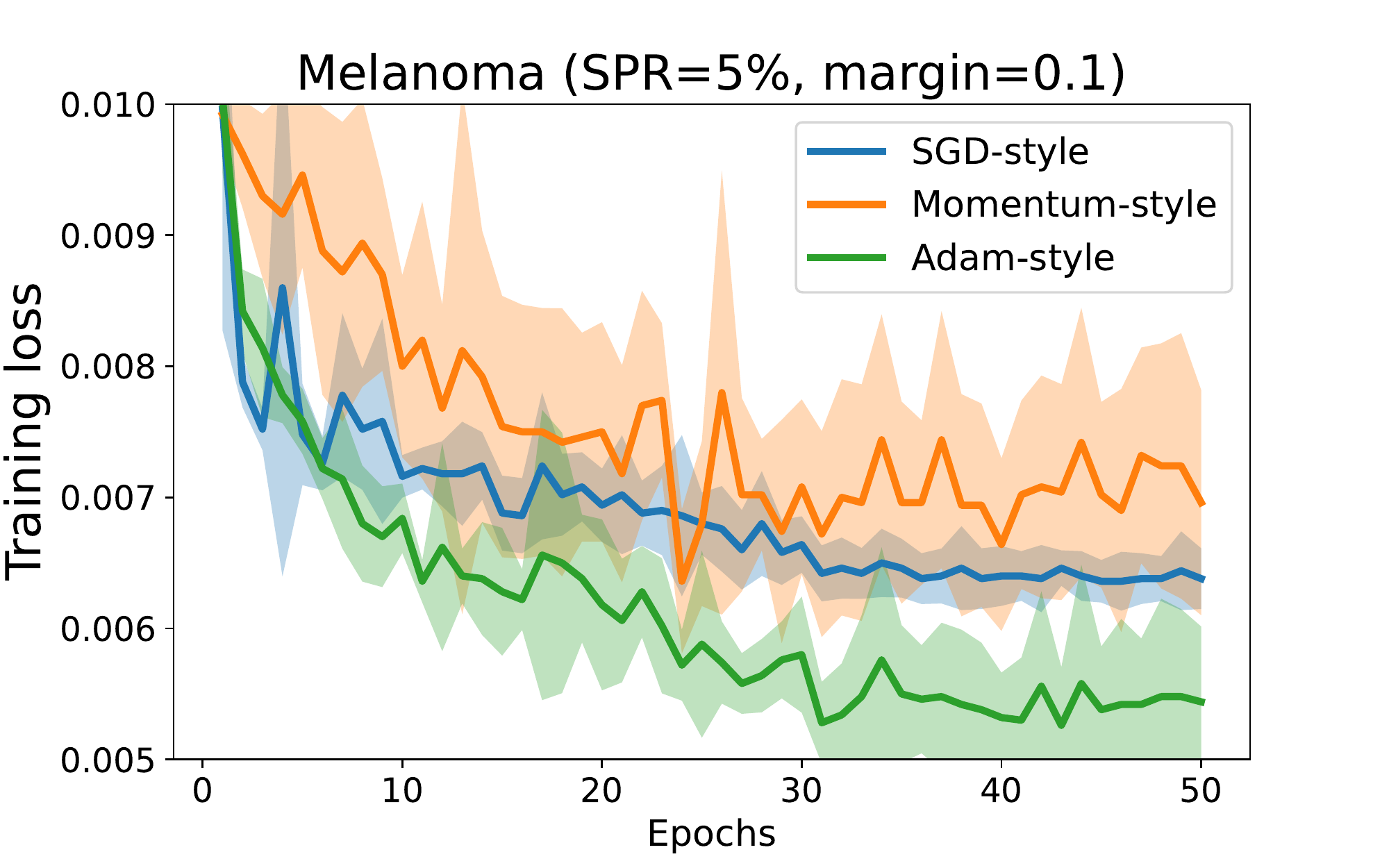}
         \label{fig:stl10-opt}
     \end{subfigure}
     \begin{subfigure}[b]{0.3\textwidth}
         \centering
         \includegraphics[width=\textwidth]{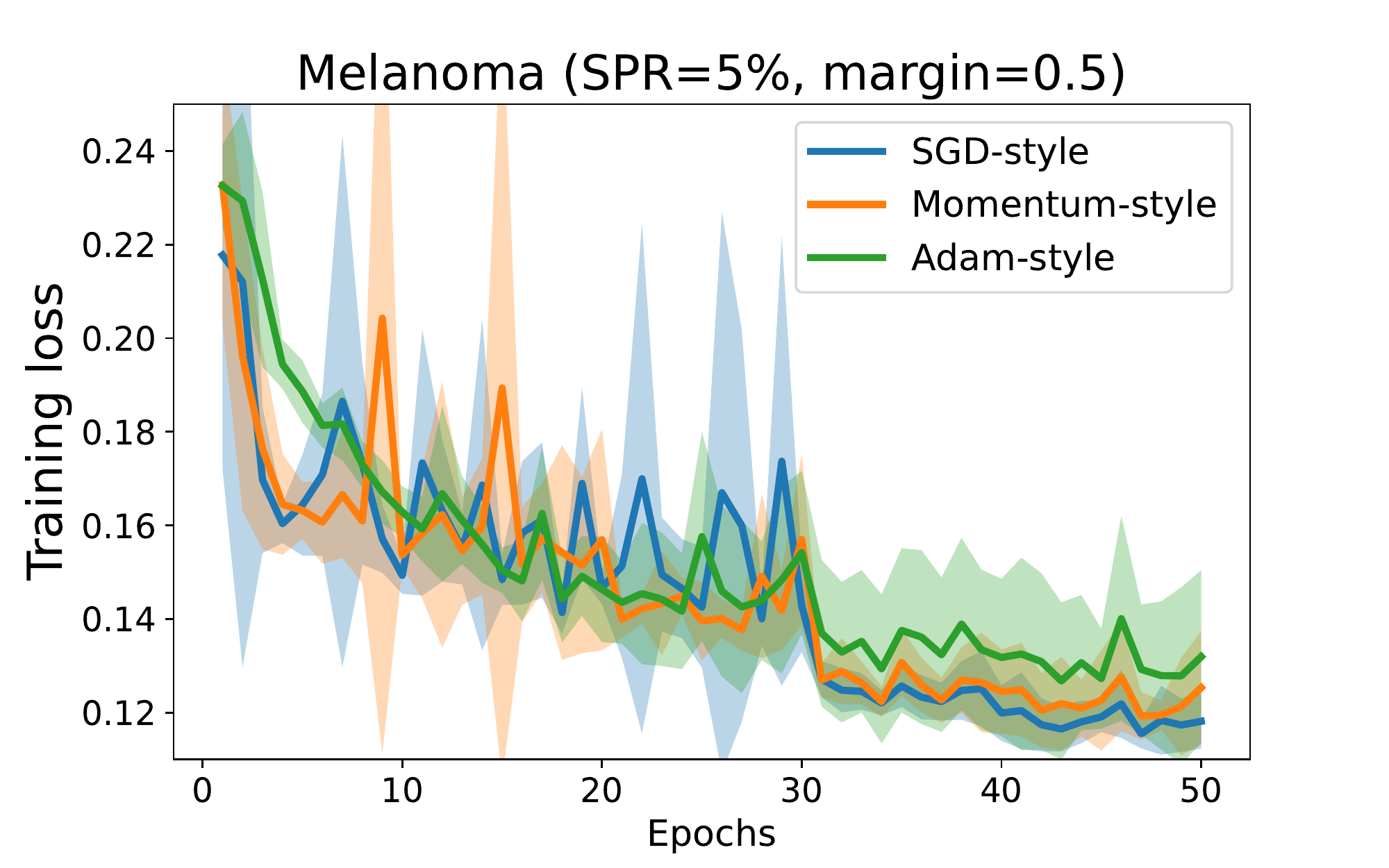}
         \label{fig:stl10-opt}
     \end{subfigure}
     \begin{subfigure}[b]{0.3\textwidth}
         \centering
         \includegraphics[width=\textwidth]{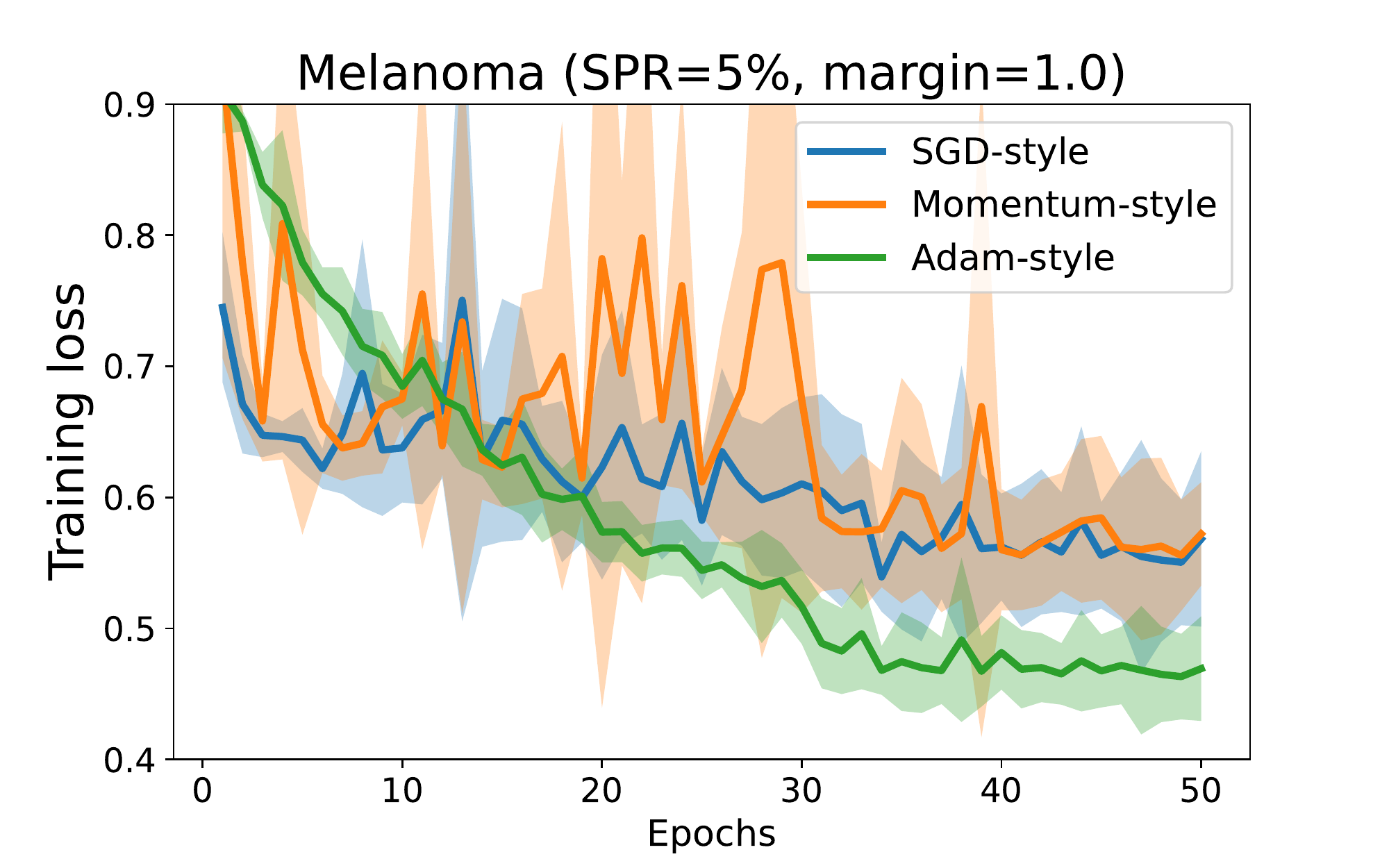}
         \label{fig:stl10-opt}
     \end{subfigure}
    \caption{Different Optimizers for Composite Squared Hinge Loss on Image datasets}
    \label{fig:training_CSH_img}
\end{figure}

\begin{figure}[h]
    \centering
    \begin{subfigure}[b]{0.3\textwidth}
         \centering
         \includegraphics[width=\textwidth]{Figures/hiv-spr025-margin=01.pdf}
         \label{fig:stl10-opt}
     \end{subfigure}
     \begin{subfigure}[b]{0.3\textwidth}
         \centering
         \includegraphics[width=\textwidth]{Figures/hiv-spr025-margin=1.pdf}
         \label{fig:stl10-opt}
     \end{subfigure}
     \begin{subfigure}[b]{0.3\textwidth}
         \centering
         \includegraphics[width=\textwidth]{Figures/hiv-spr025-margin=10.pdf}
         \label{fig:stl10-opt}
     \end{subfigure}
     \begin{subfigure}[b]{0.3\textwidth}
         \centering
         \includegraphics[width=\textwidth]{Figures/muv-spr001-margin=01.pdf}
         \label{fig:stl10-opt}
     \end{subfigure}
     \begin{subfigure}[b]{0.3\textwidth}
         \centering
         \includegraphics[width=\textwidth]{Figures/muv-spr001-margin=05.pdf}
         \label{fig:stl10-opt}
     \end{subfigure}
     \begin{subfigure}[b]{0.3\textwidth}
         \centering
         \includegraphics[width=\textwidth]{Figures/muv-spr001-margin=1.pdf}
         \label{fig:stl10-opt}
     \end{subfigure}
     \begin{subfigure}[b]{0.3\textwidth}
         \centering
         \includegraphics[width=\textwidth]{Figures/tox21-t0-spr025-margin=01.pdf}
         \label{fig:stl10-opt}
     \end{subfigure}
     \begin{subfigure}[b]{0.3\textwidth}
         \centering
         \includegraphics[width=\textwidth]{Figures/tox21-t0-spr025-margin=1.pdf}
         \label{fig:stl10-opt}
     \end{subfigure}
     \begin{subfigure}[b]{0.3\textwidth}
         \centering
         \includegraphics[width=\textwidth]{Figures/tox21-t0-spr025-margin=10.pdf}
         \label{fig:stl10-opt}
     \end{subfigure}
     \begin{subfigure}[b]{0.3\textwidth}
         \centering
         \includegraphics[width=\textwidth]{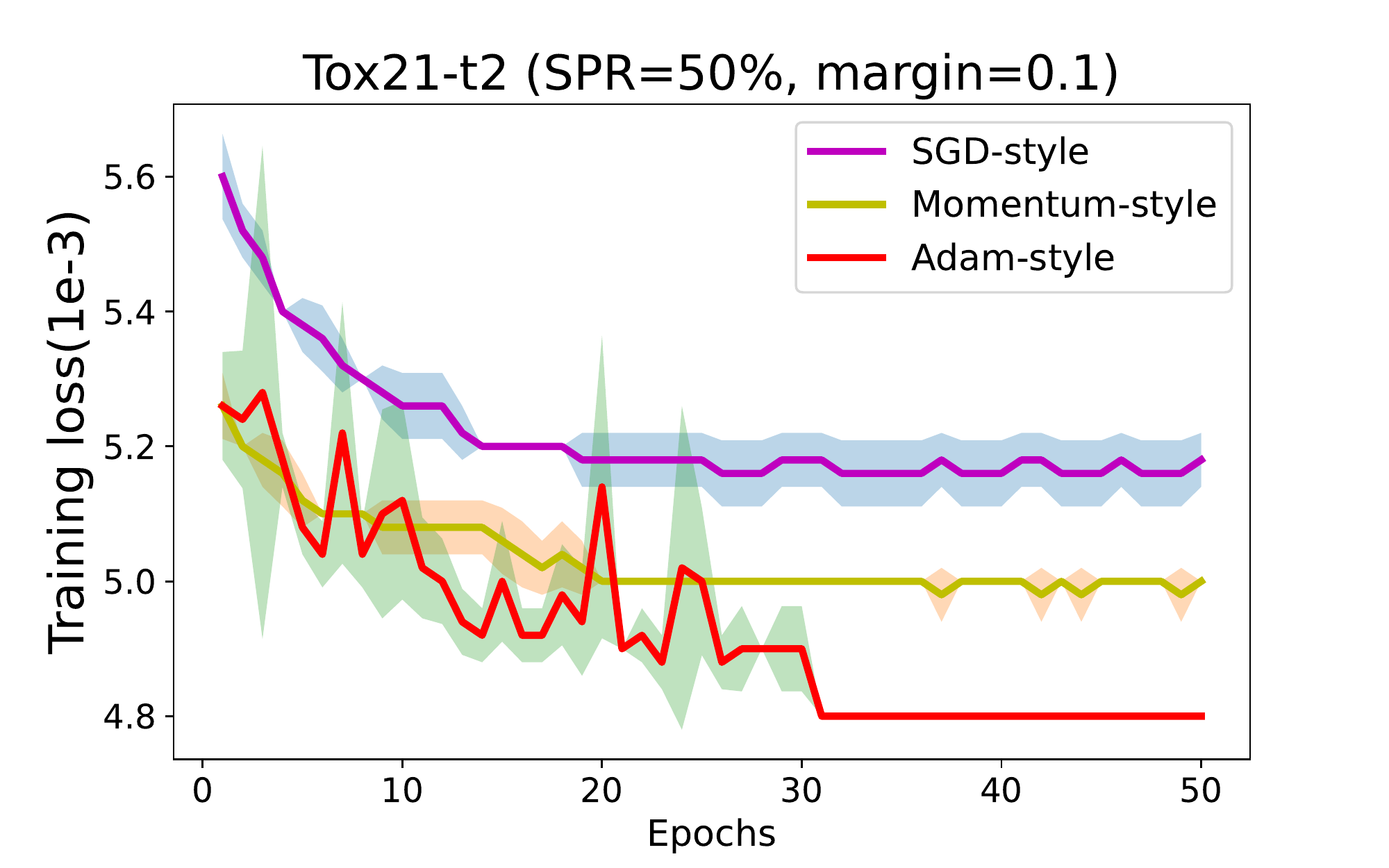}
         \label{fig:stl10-opt}
     \end{subfigure}
     \begin{subfigure}[b]{0.3\textwidth}
         \centering
         \includegraphics[width=\textwidth]{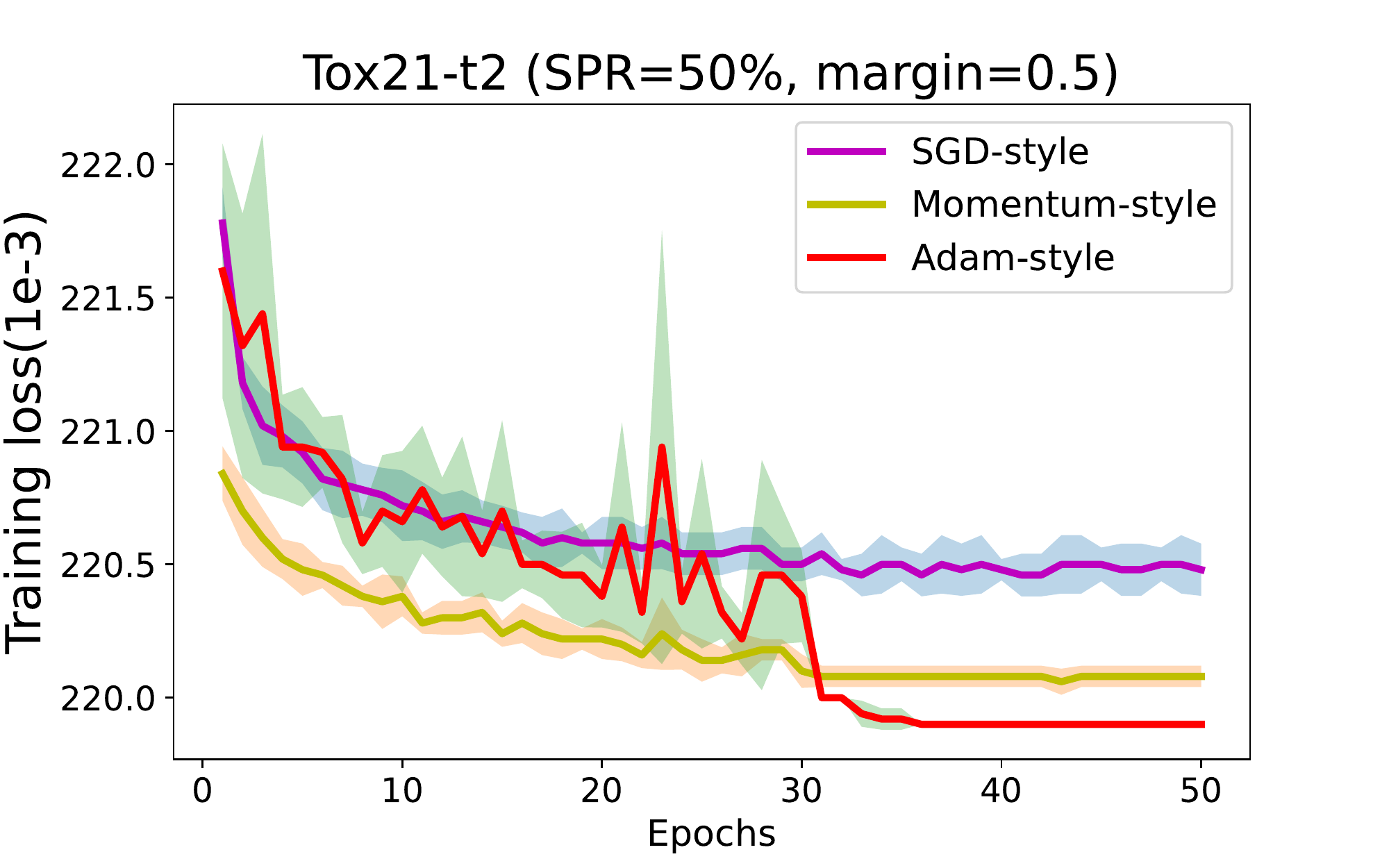}
         \label{fig:stl10-opt}
     \end{subfigure}
     \begin{subfigure}[b]{0.3\textwidth}
         \centering
         \includegraphics[width=\textwidth]{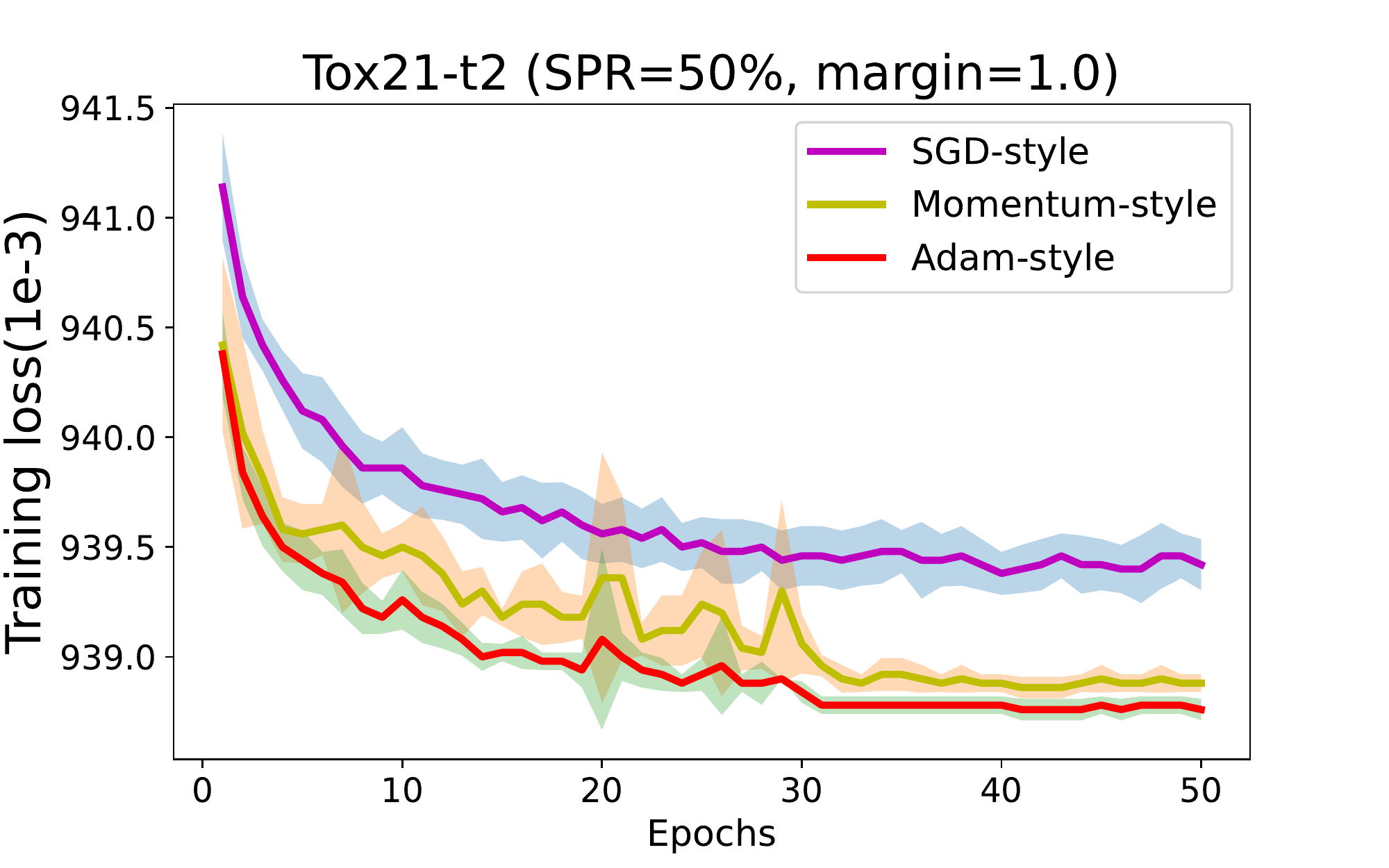}
         \label{fig:stl10-opt}
     \end{subfigure}
     \begin{subfigure}[b]{0.3\textwidth}
         \centering
         \includegraphics[width=\textwidth]{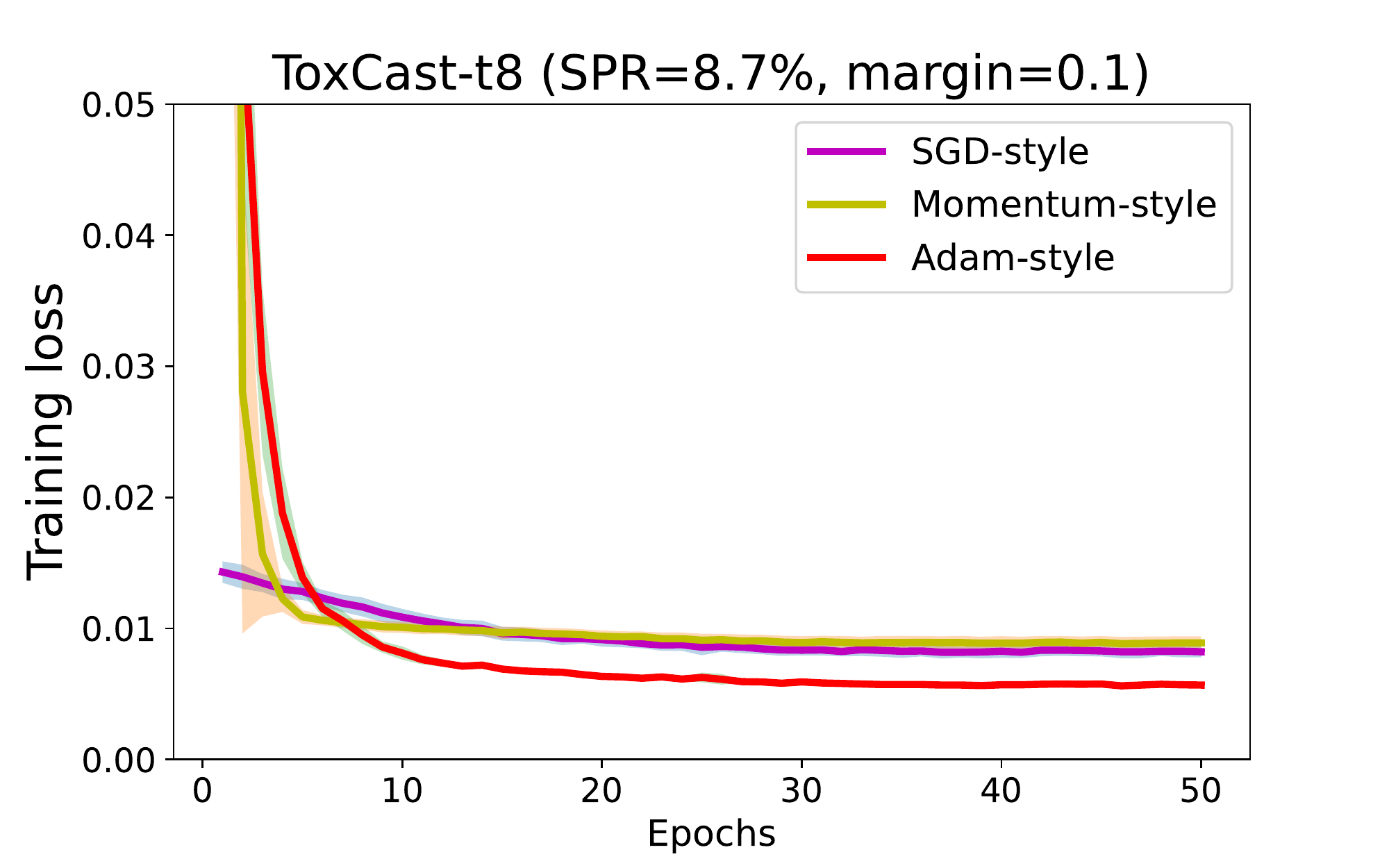}
         \label{fig:stl10-opt}
     \end{subfigure}
     \begin{subfigure}[b]{0.3\textwidth}
         \centering
         \includegraphics[width=\textwidth]{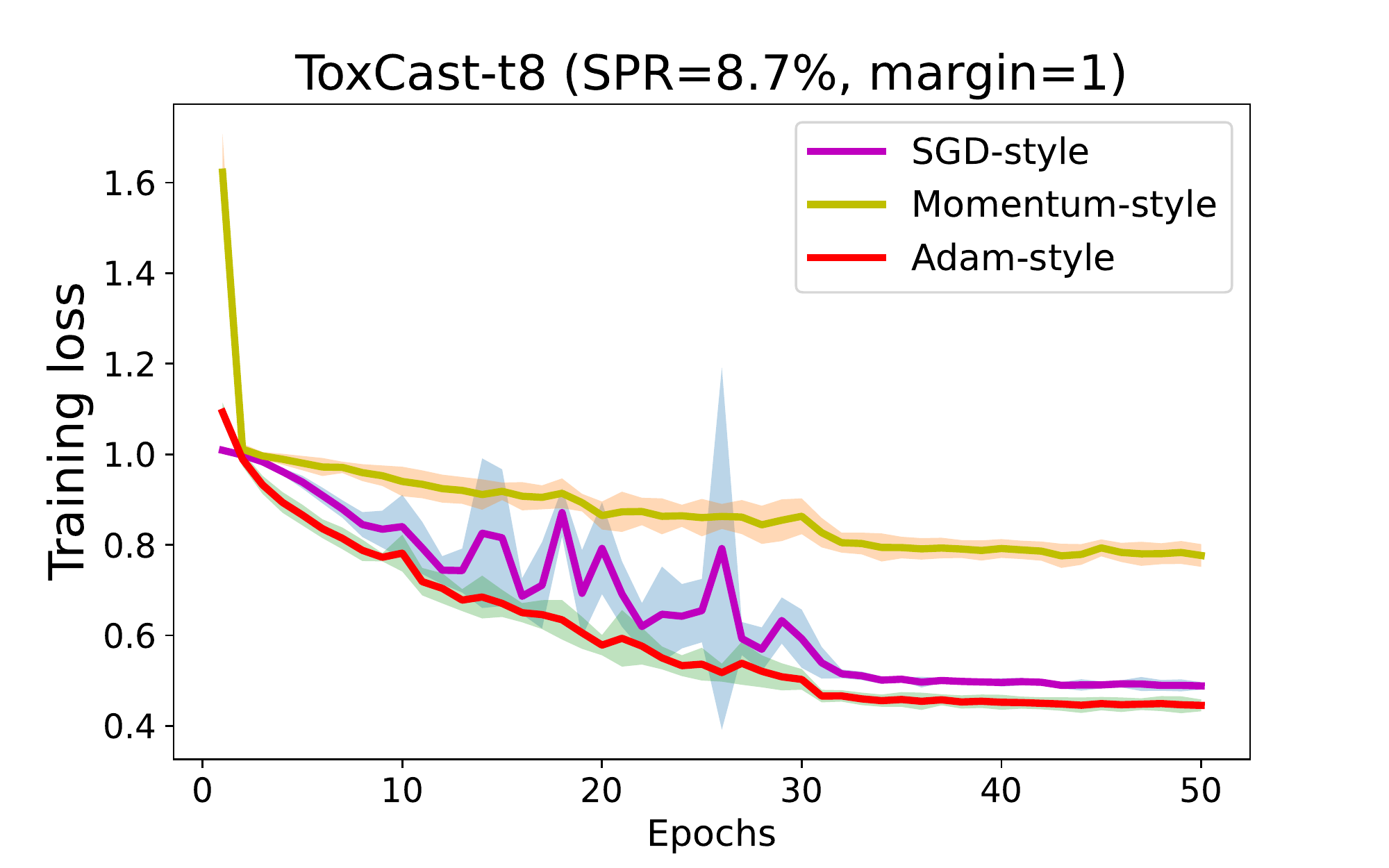}
         \label{fig:stl10-opt}
     \end{subfigure}
     \begin{subfigure}[b]{0.3\textwidth}
         \centering
         \includegraphics[width=\textwidth]{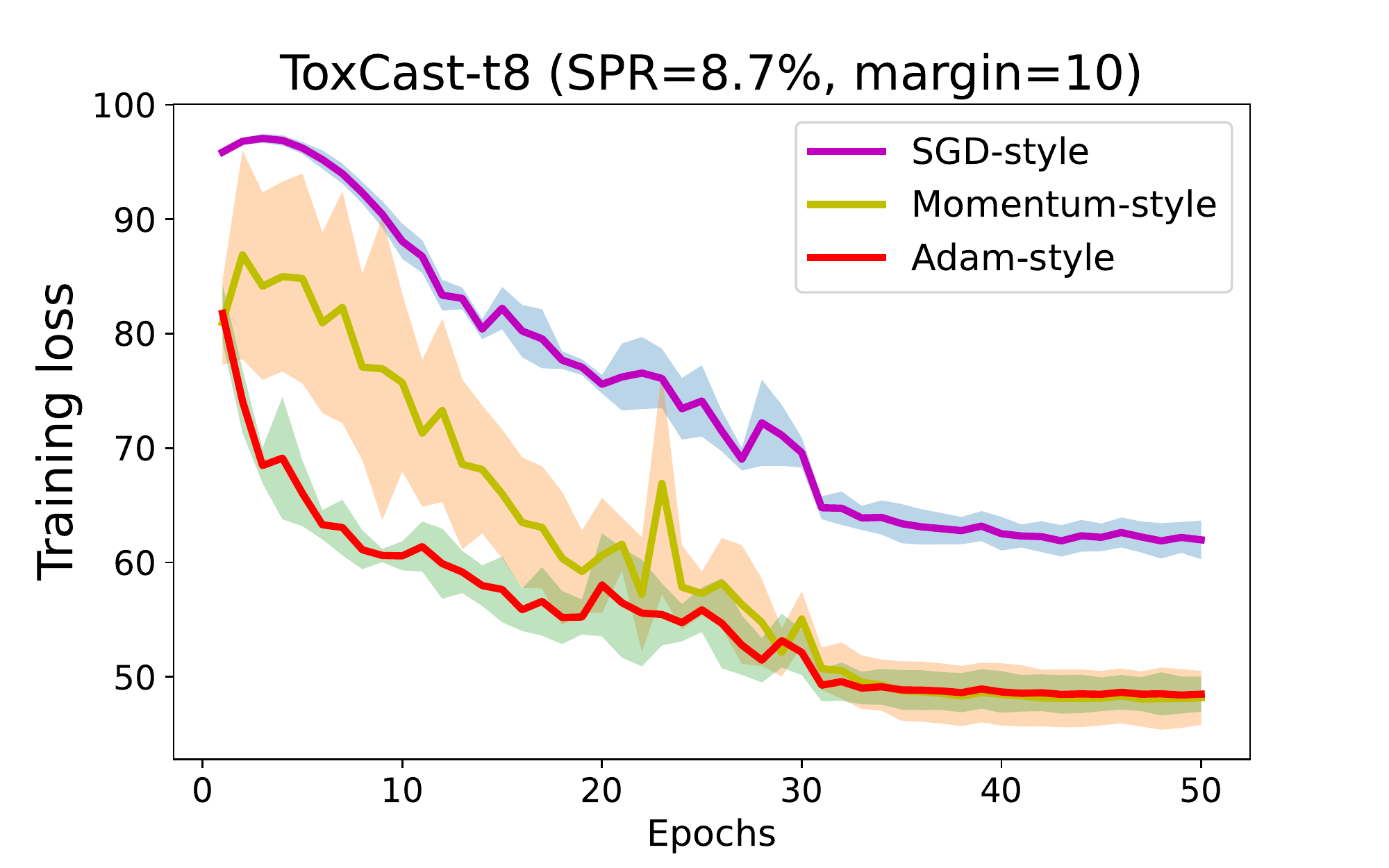}
         \label{fig:stl10-opt}
     \end{subfigure}
     \begin{subfigure}[b]{0.3\textwidth}
         \centering
         \includegraphics[width=\textwidth]{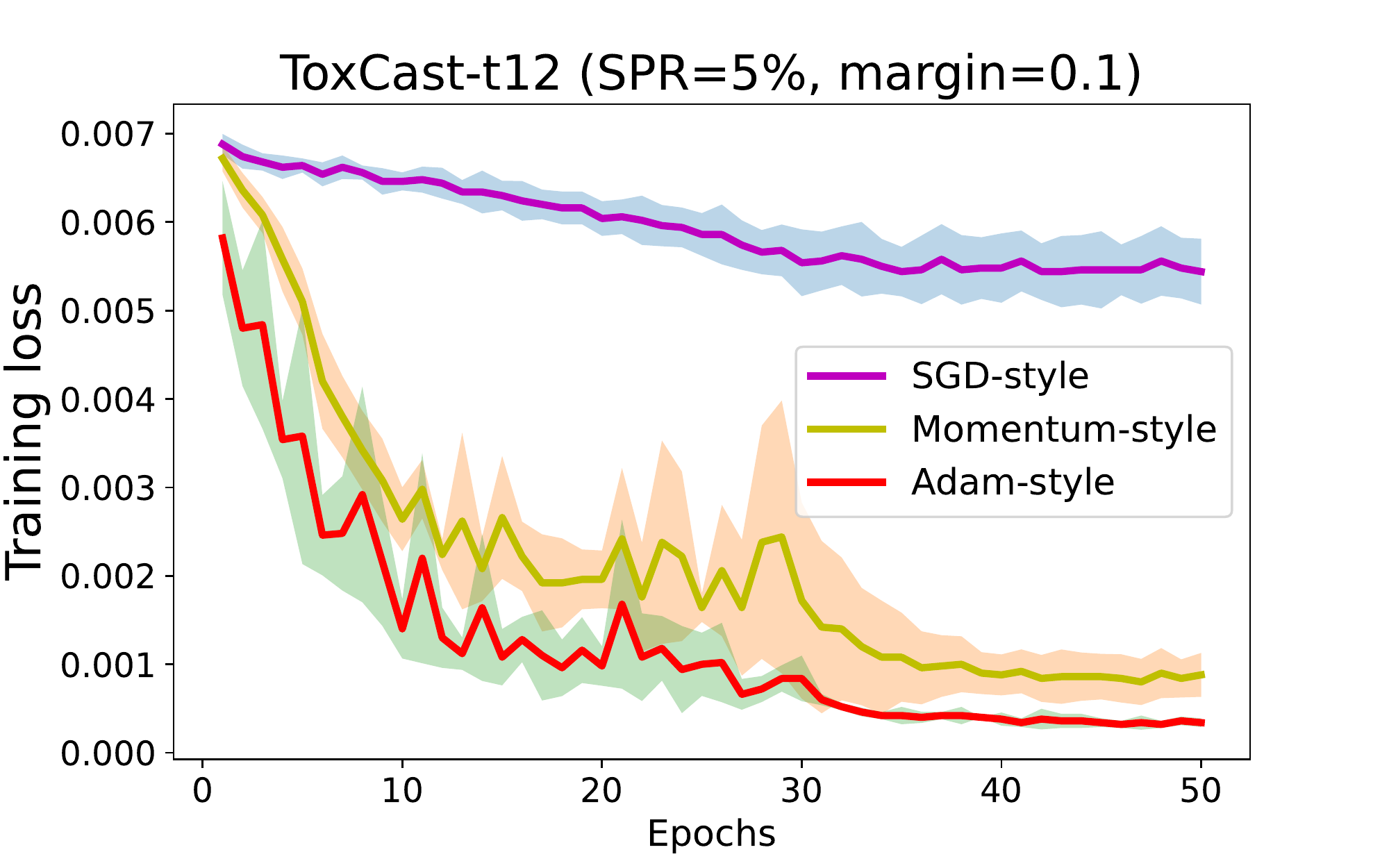}
         \label{fig:stl10-opt}
     \end{subfigure}
     \begin{subfigure}[b]{0.3\textwidth}
         \centering
         \includegraphics[width=\textwidth]{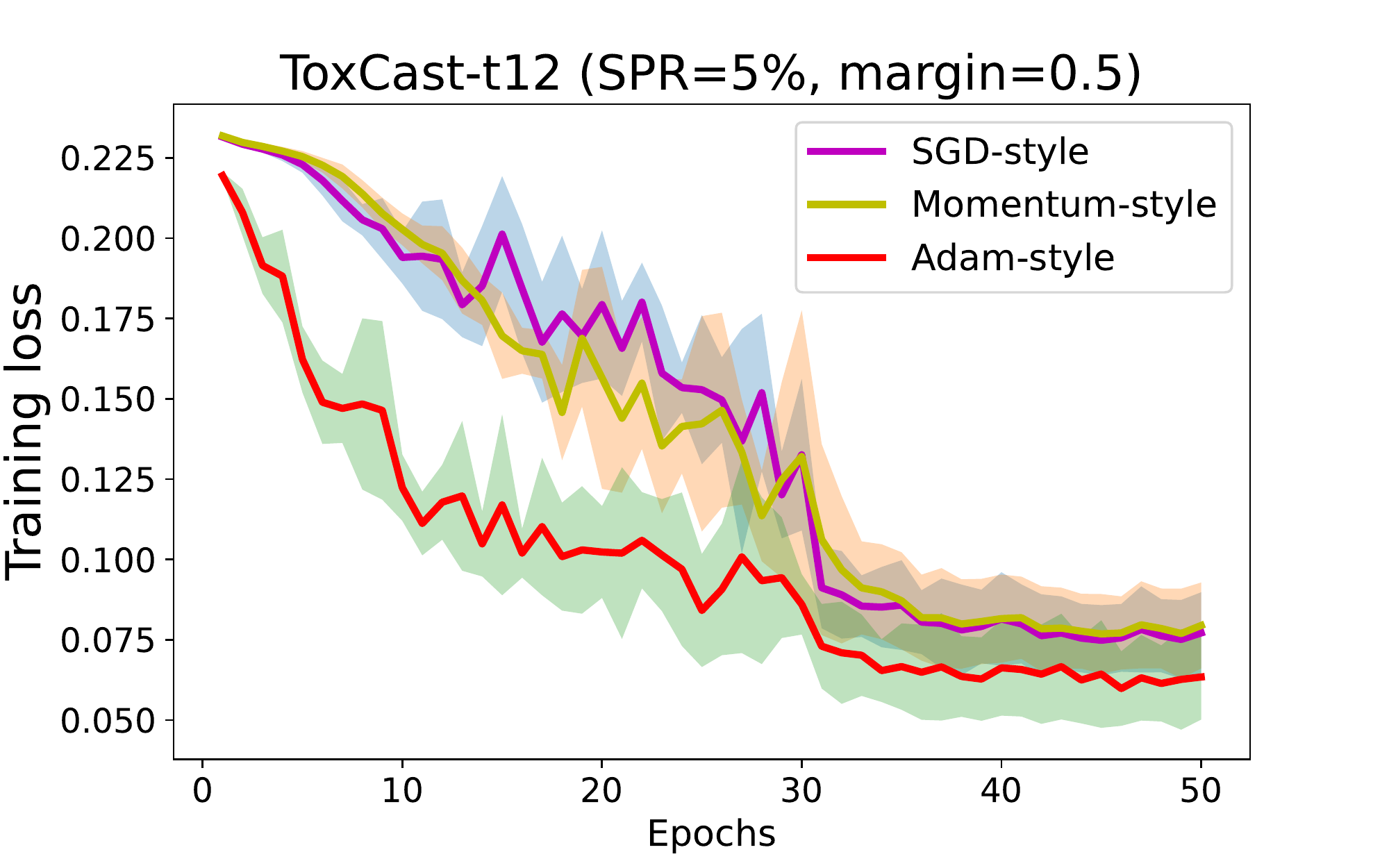}
         \label{fig:stl10-opt}
     \end{subfigure}
     \begin{subfigure}[b]{0.3\textwidth}
         \centering
         \includegraphics[width=\textwidth]{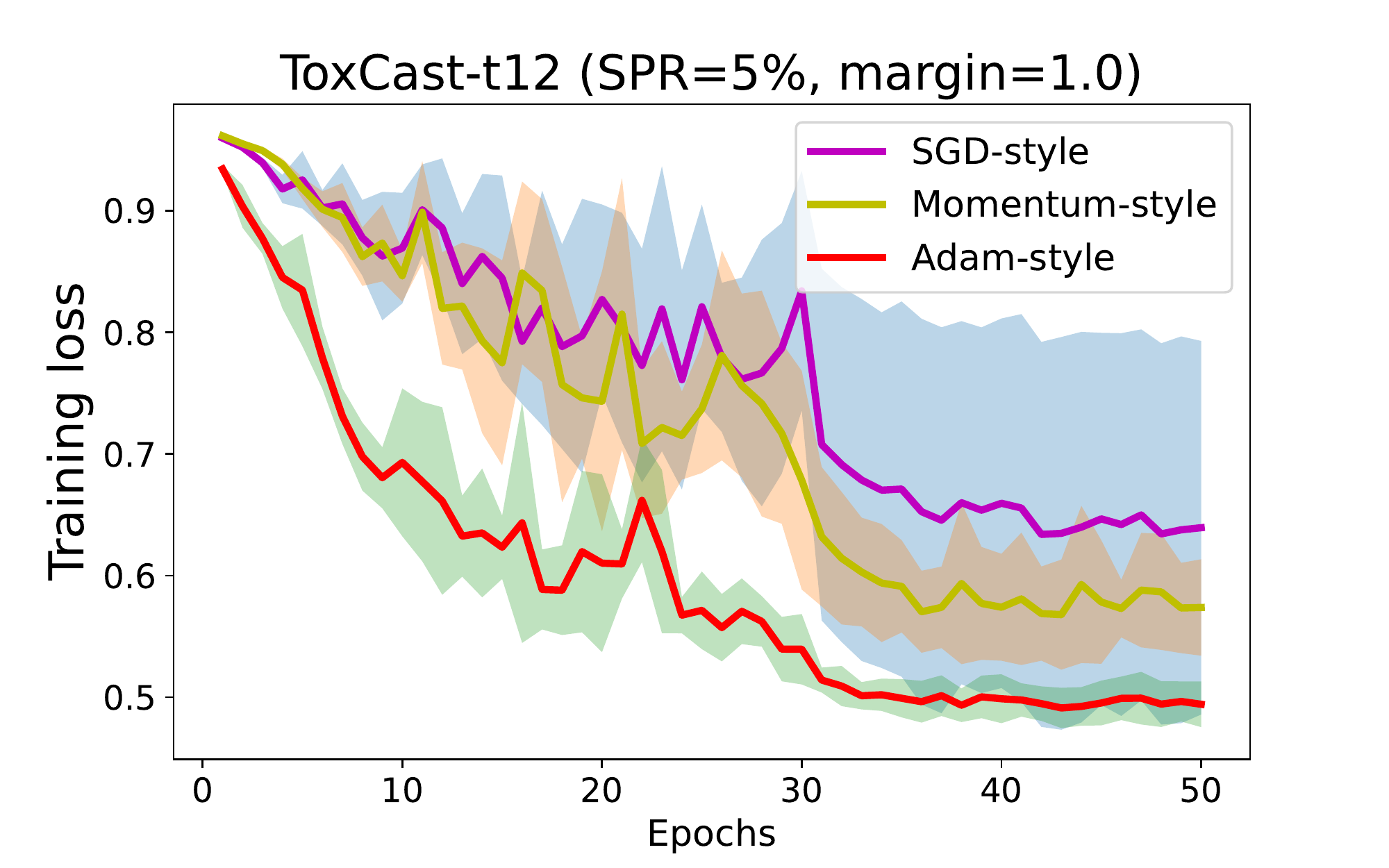}
         \label{fig:stl10-opt}
     \end{subfigure}
    \caption{Different Optimizers for Composite Squared Hinge Loss on Molecule datasets}
    \label{fig:training_CSH_mole}
\end{figure}

We include all the figures for optimizers (SGD, Momentum and Adam styles) for CSH loss at Figure~\ref{fig:training_CSH_img} for image datasets and Figure~\ref{fig:training_CSH_mole} for molecule datasets. CAdam is the most competitive optimizer for all the datasets, and even achieve dominating performance for molecule datasets from the training convergence perspective. However, as emphasized at the main content, the drawback for CAdam is it doesn't generalize well.
\subsection{Optimization: PSQ v.s. CSQ}
\begin{figure}[h]
    \centering
    \begin{subfigure}[b]{0.4\textwidth}
         \centering
     \includegraphics[width=\textwidth]{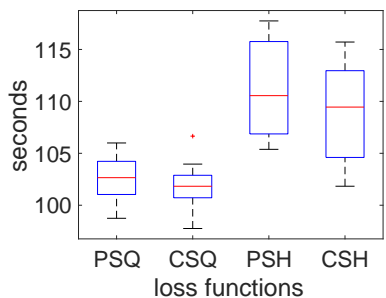}
     \end{subfigure}
     \caption{Time Consuming for every 40 iterations}
    \label{fig:time-cost}
\end{figure}

\begin{figure}[h]
    \centering
    \begin{subfigure}[b]{0.4\textwidth}
         \centering
         \includegraphics[width=\textwidth]{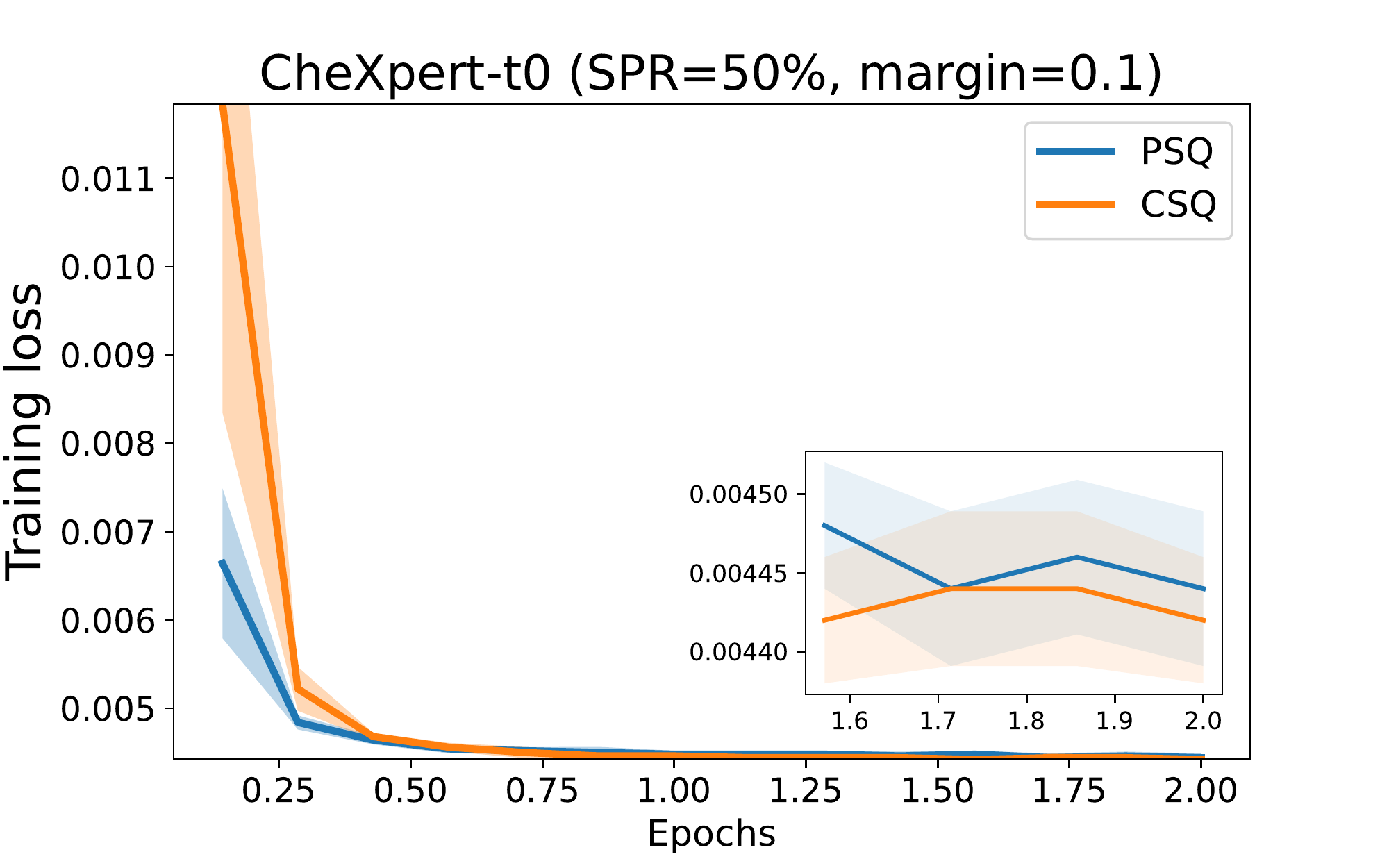}
     \end{subfigure}
     \begin{subfigure}[b]{0.4\textwidth}
         \centering
         \includegraphics[width=\textwidth]{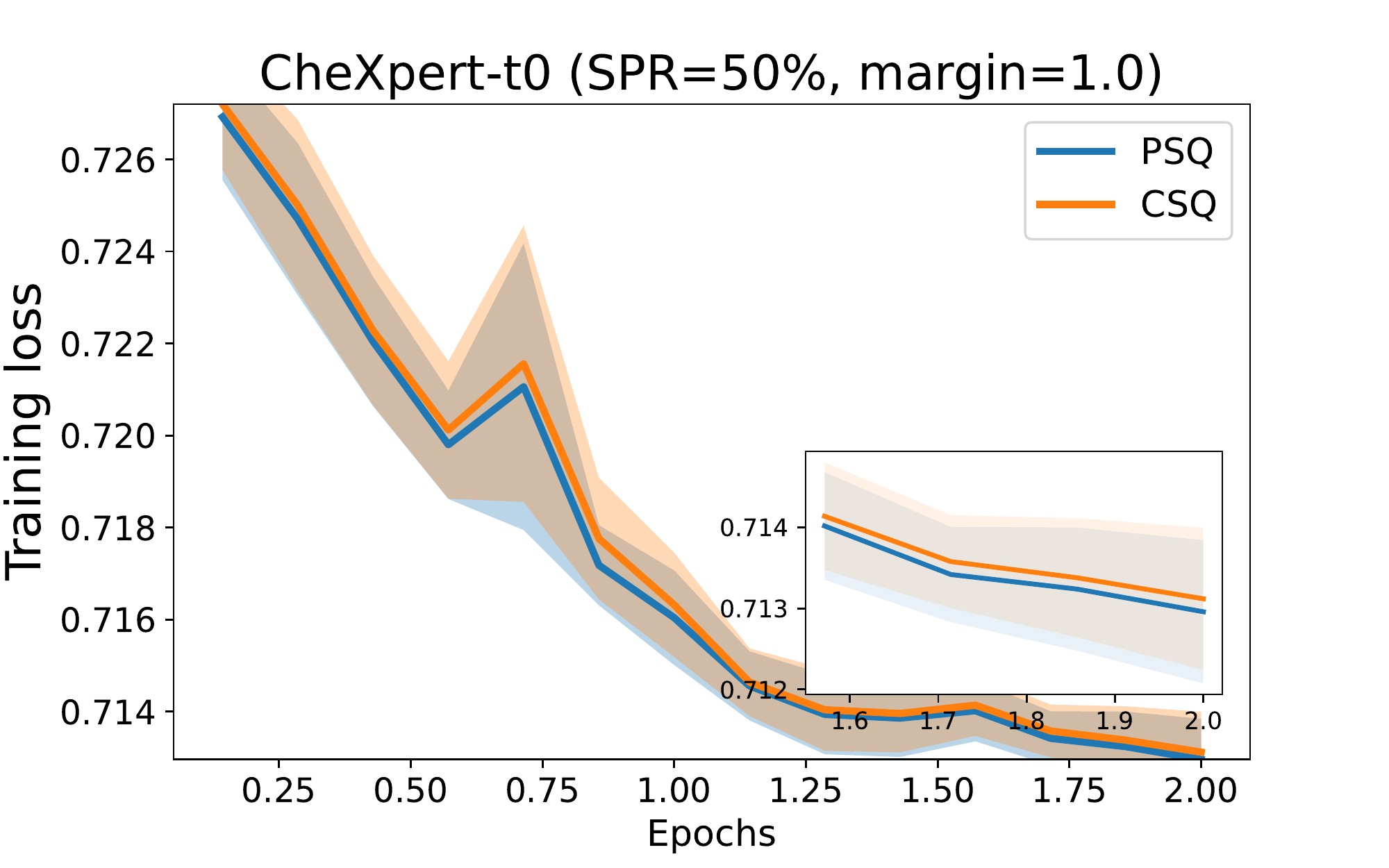}
     \end{subfigure}
     \begin{subfigure}[b]{0.4\textwidth}
         \centering
         \includegraphics[width=\textwidth]{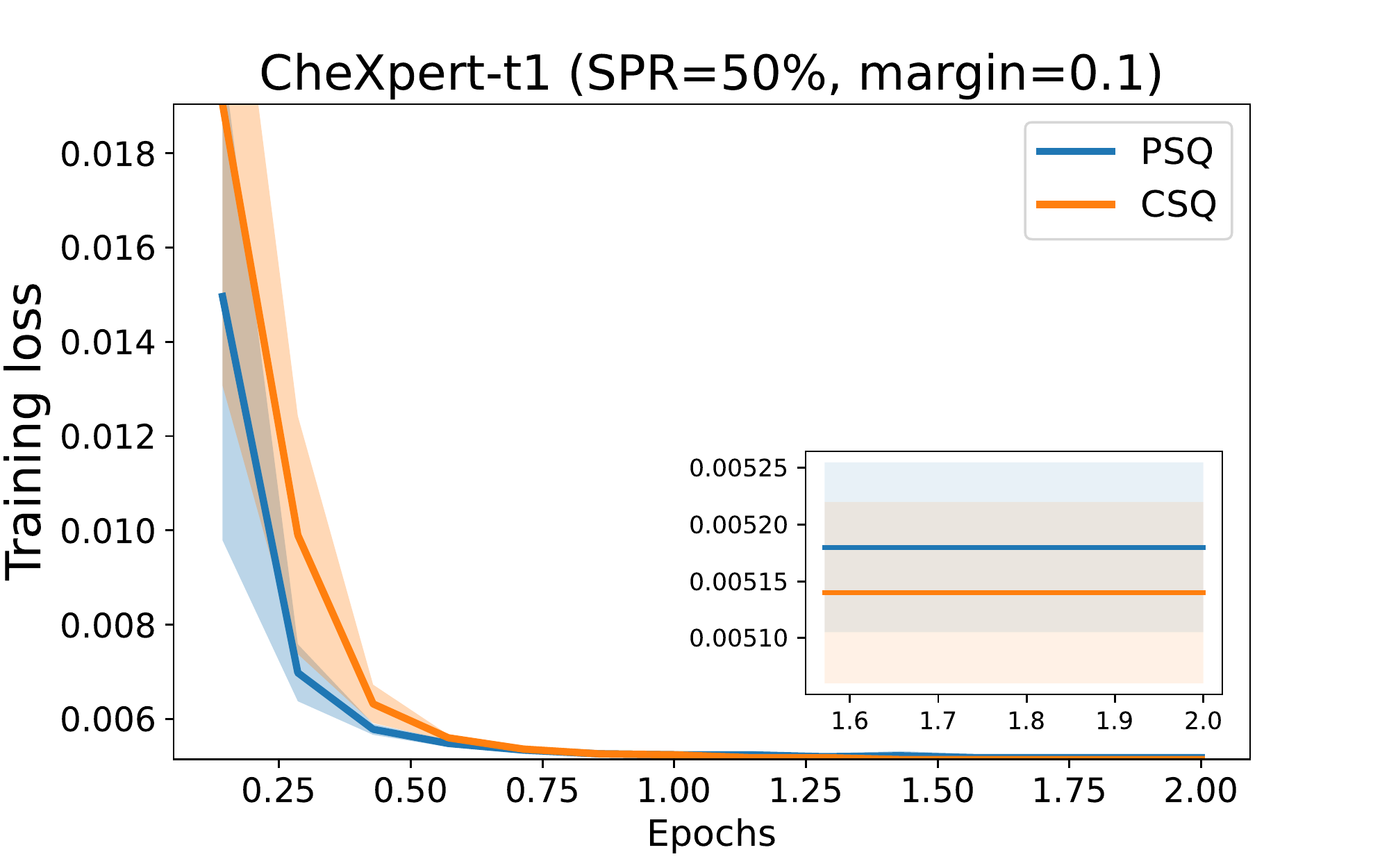}
     \end{subfigure}
     \begin{subfigure}[b]{0.4\textwidth}
         \centering
         \includegraphics[width=\textwidth]{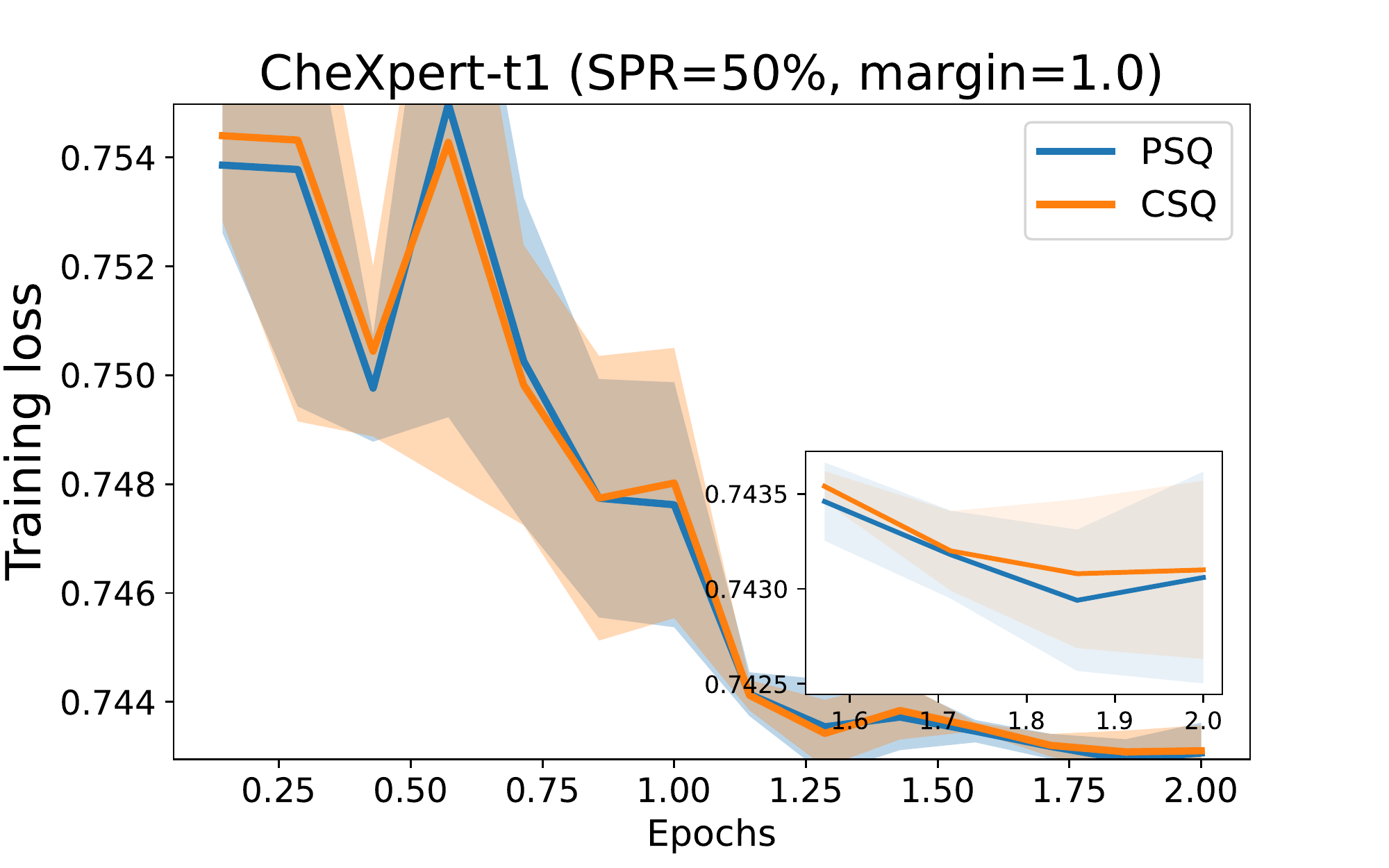}
     \end{subfigure}
     \begin{subfigure}[b]{0.4\textwidth}
         \centering
         \includegraphics[width=\textwidth]{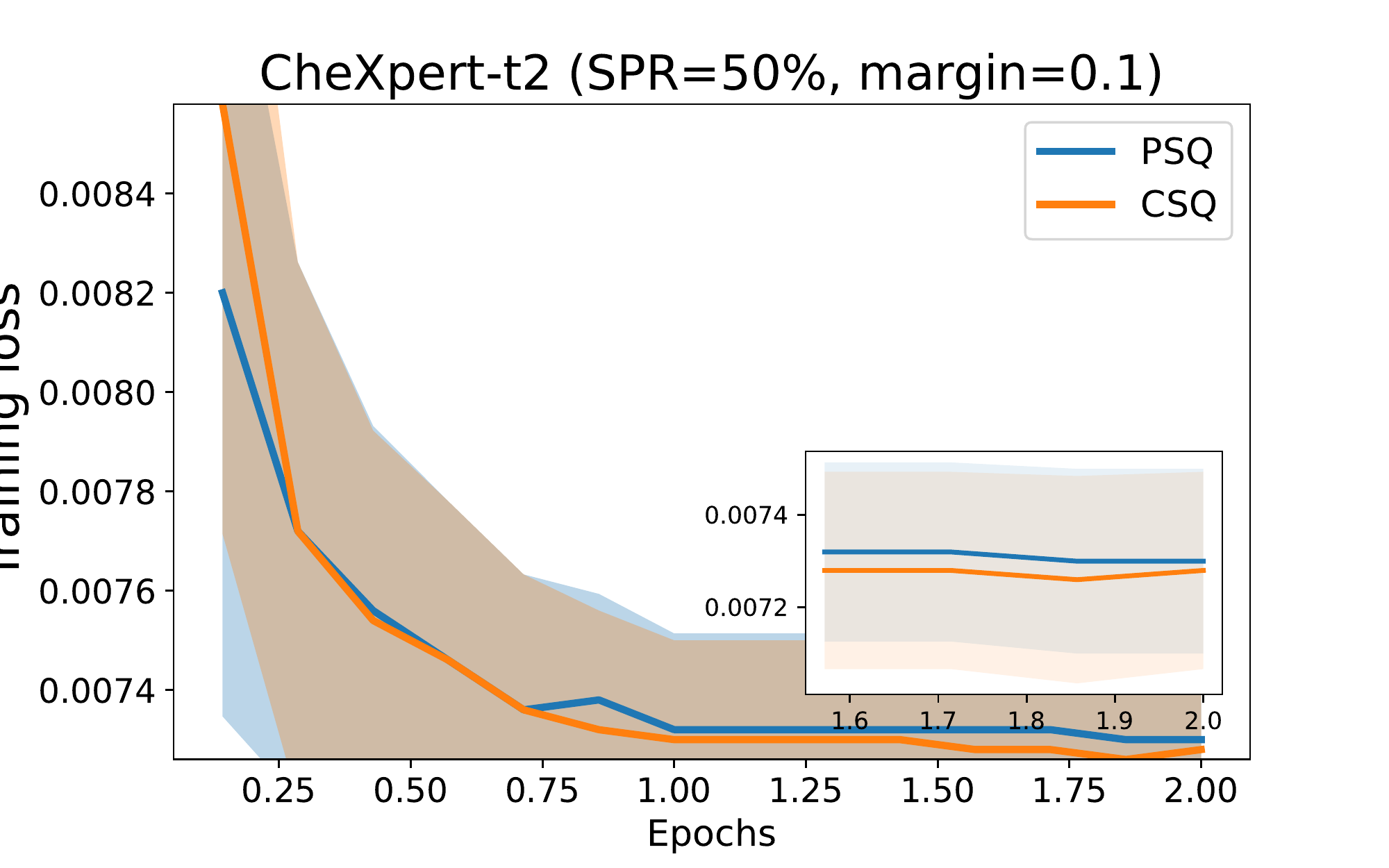}
     \end{subfigure}
     \begin{subfigure}[b]{0.4\textwidth}
         \centering
         \includegraphics[width=\textwidth]{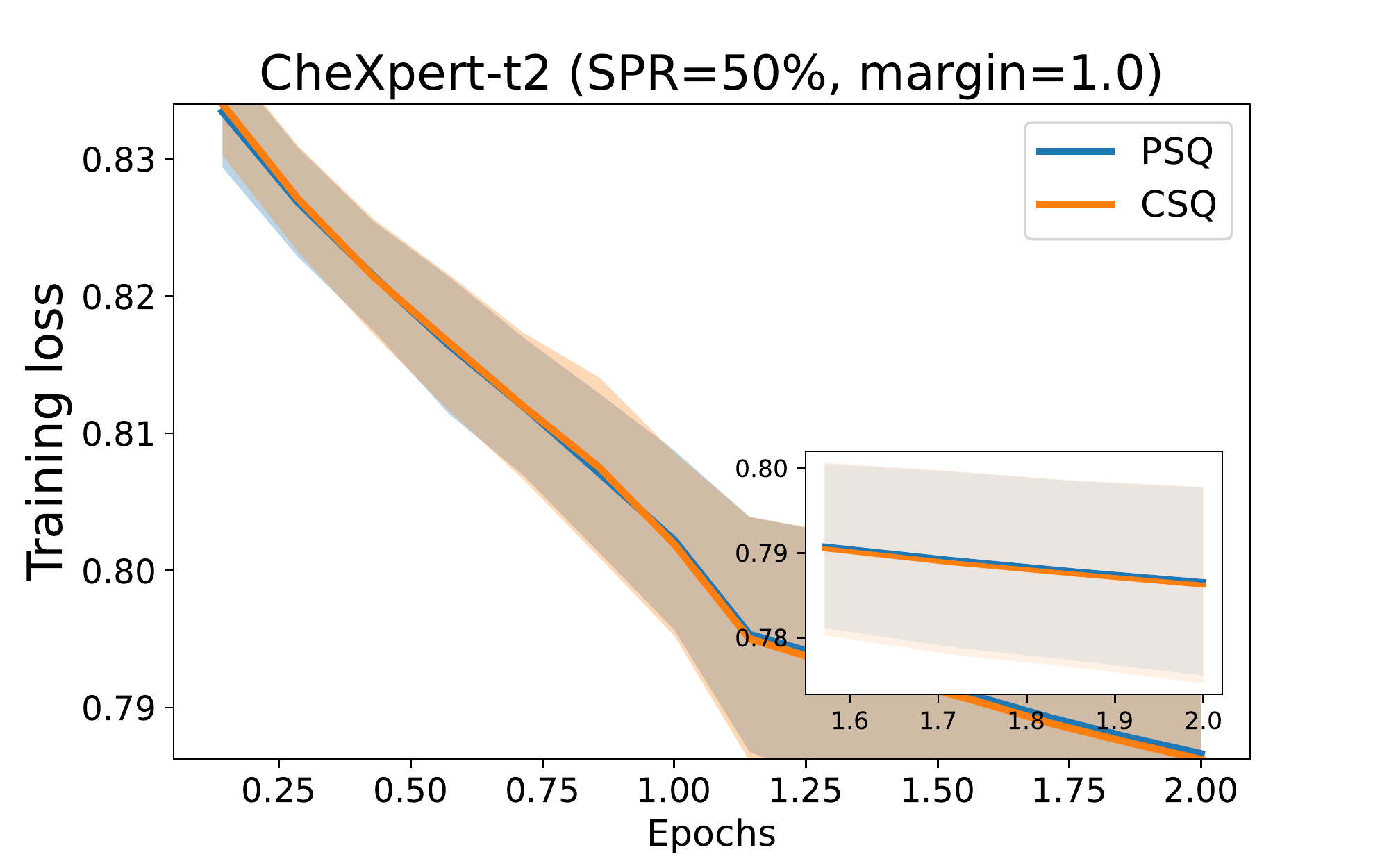}
     \end{subfigure}
     \begin{subfigure}[b]{0.4\textwidth}
         \centering
         \includegraphics[width=\textwidth]{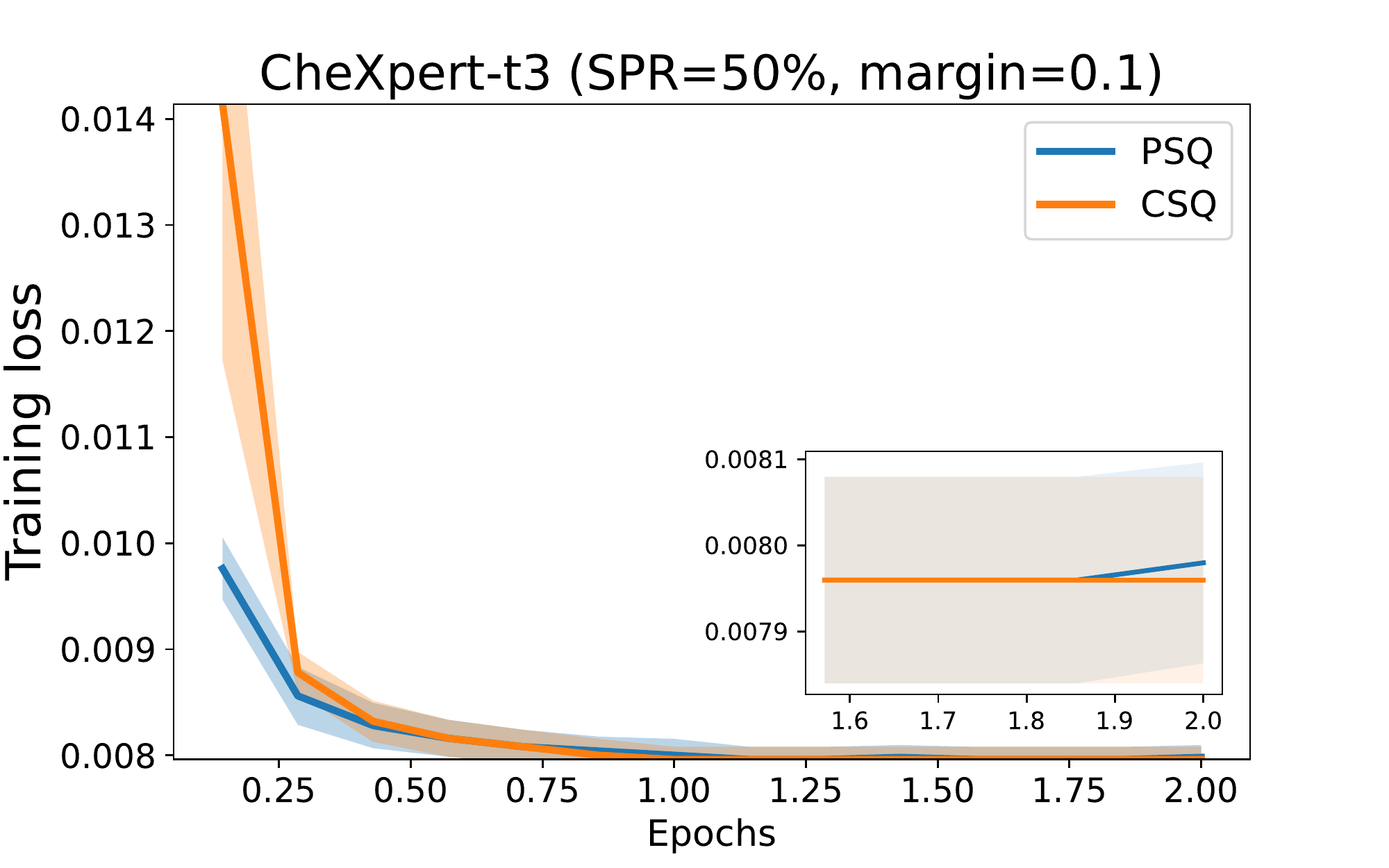}
     \end{subfigure}
     \begin{subfigure}[b]{0.4\textwidth}
         \centering
         \includegraphics[width=\textwidth]{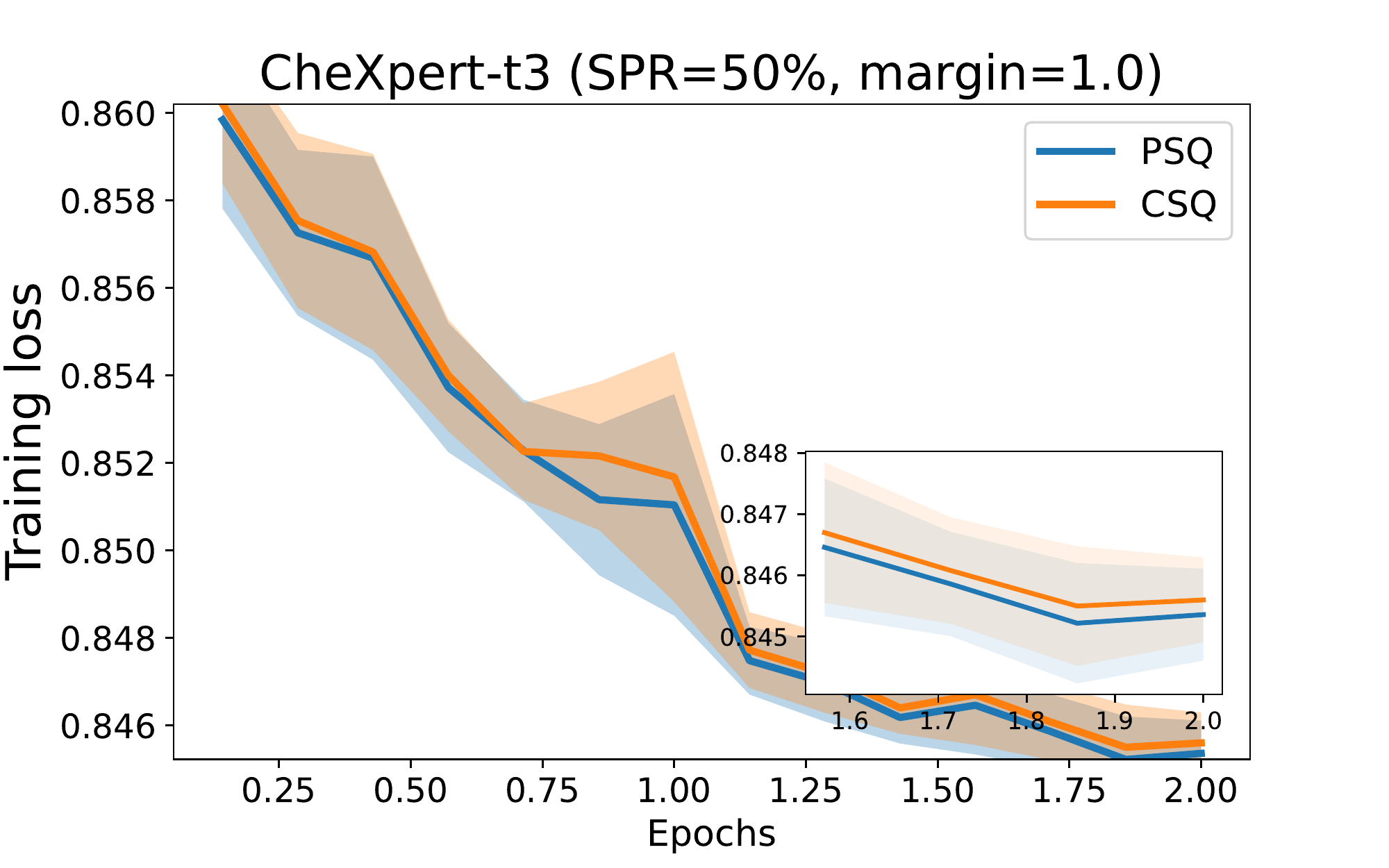}
     \end{subfigure}
     \begin{subfigure}[b]{0.4\textwidth}
         \centering
         \includegraphics[width=\textwidth]{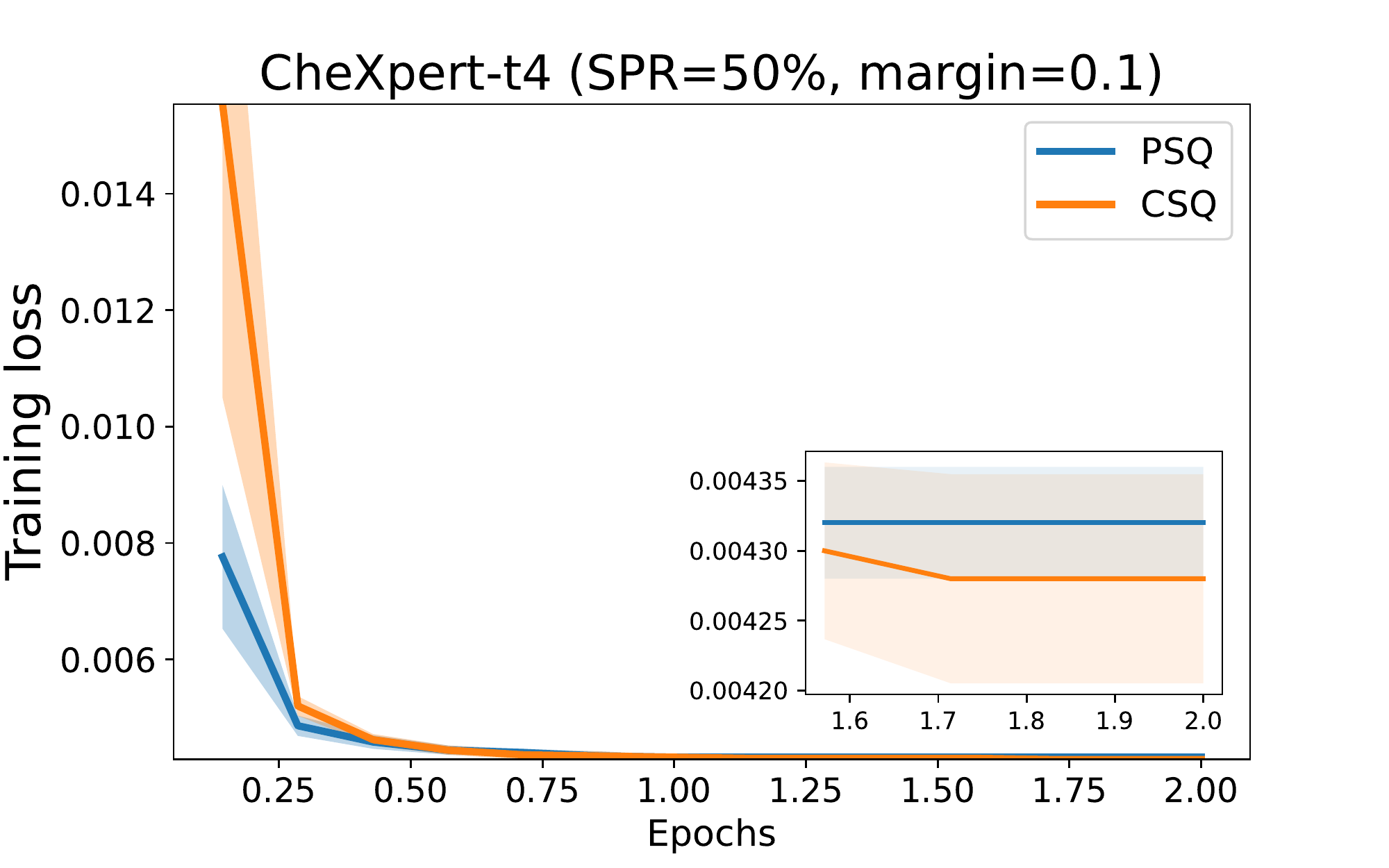}
     \end{subfigure}
     \begin{subfigure}[b]{0.4\textwidth}
         \centering
         \includegraphics[width=\textwidth]{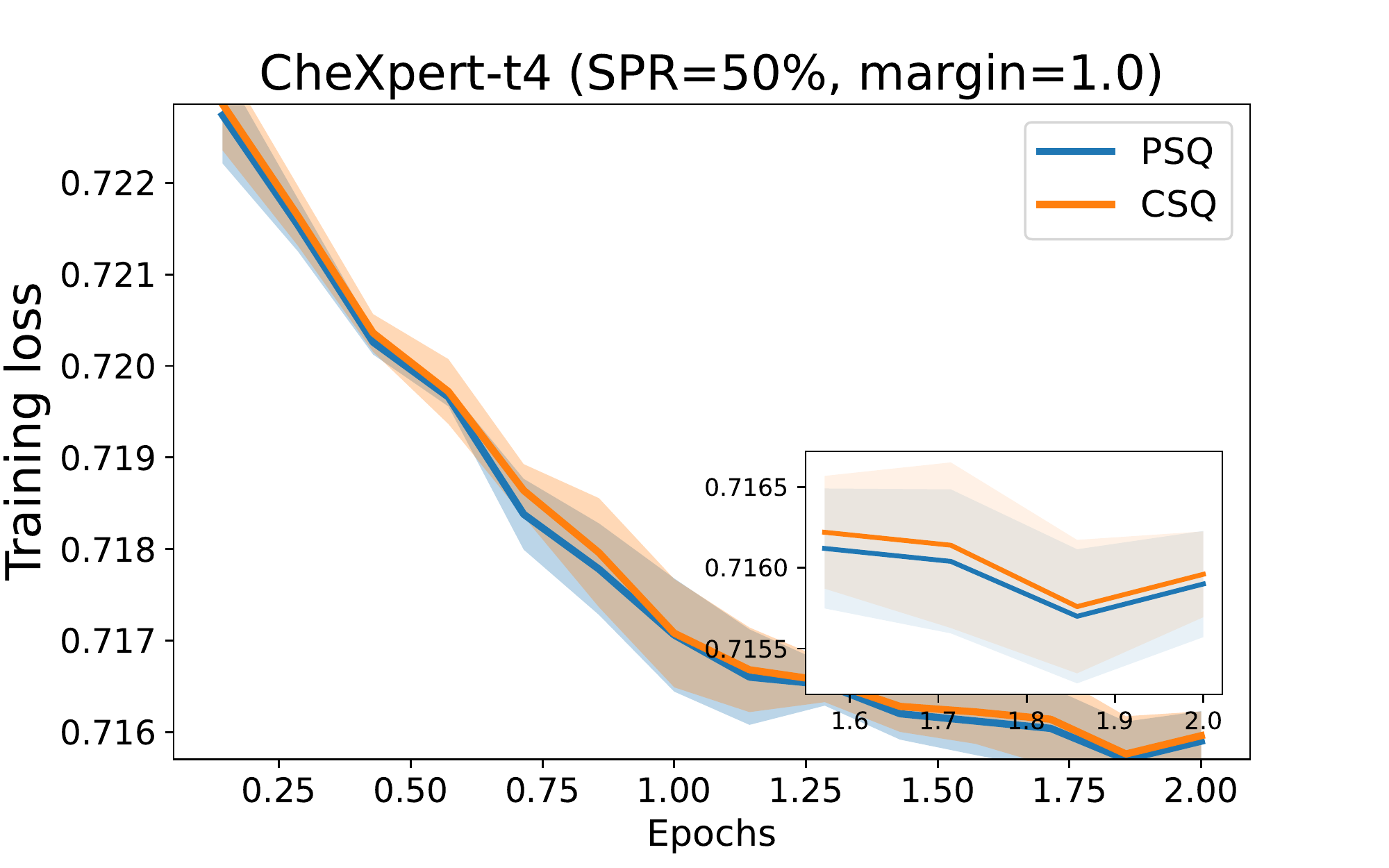}
     \end{subfigure}
    \caption{PSQ v.s. CSQ on CheXpert dataset}
    \label{fig:psq-csq}
\end{figure}

Because PSQ and CSQ are proved equivalent~\cite{ying2016stochastic}, we may wonder which one is better only from the optimization perspective. To this purpose, we compare them under sigmoid normalization for last layer, momentum optimizer with learning rate as 1e-3. We include the training convergence for PSQ and CSQ loss on CheXpert dataset for the five targets, t0: Cardiomegaly, t1: Edema, t2: Consolidation, t3: Atelectasis, t4: Pleural Effusion, at Figure~\ref{fig:psq-csq}. From the results, we can see CSQ is clearly better than PSQ for smaller margin parameter (0.1), but slightly worse when margin is larger (1.0). To this end, we investigate the running time for PSQ v.s. CSQ, and PSH v.s. CSH because they are  equivalent as well. We count the running time under GeForce GTX 1080 Ti Graphics Card for every 40 iterations of mini-batch of 64 at Figure~\ref{fig:time-cost}, where We repeat the counting independently for 20 times. Both CSQ and CSH are faster than PSQ and PSH, which could due to pairwise loss need generate all pairs inside of mini-batch. 


\begin{table}[h]
    \centering
    \caption{Weight decay with testing AUROC}
    \label{tab:te_WD}
    \resizebox{0.9\textwidth}{!}{
    \begin{tabular}{lllllll}
    \toprule
       Loss &Dataset & WD$=1e-1$& WD$=1e-2$& WD$=1e-3$& WD$=1e-4$& WD$=0$\\
        \hline
        &STL10 &  0.761(0.040) &  \textbf{0.793}(0.013) &  0.772(0.023)&  0.761(0.014)&0.780(0.019)\\
        &CIFAR10 &  0.620(0.010) &  0.899(0.002) &  \textbf{0.911}(0.001)&  0.907(0.002)&0.907(0.002)\\
       &CIFAR100 & 0.549(0.007) &  0.652(0.025) &  \textbf{0.681}(0.006)&  0.669(0.011)&0.678(0.008)\\
        &Cat vs Dog &  0.688(0.010) &  \textbf{0.945}(0.004) &  0.932(0.003)&  0.930(0.004)&0.927(0.007)\\
        &Melanoma &  0.719(0.014) &  0.802(0.012) &  0.794(0.013)&  0.798(0.012)&\textbf{0.806}(0.019)\\
        &Cardiomegaly & \textbf{0.841}(0.007) &  0.835(0.006) & 0.826(0.005)&  0.832(0.007)	&0.828(0.006)	\\
        PSQ &HIV &   0.753(0.021) & 0.753(0.015) & 0.758(0.007) & \textbf{0.759}(0.014)&0.750(0.008)\\
        &MUV(bioassay) &   0.518(0.060) & 0.563(0.089) & 0.651(0.102) & \textbf{0.638}(0.033)&0.627(0.050)\\
        &tox21(t0) &   0.743(0.011) & \textbf{0.748}(0.015) & 0.742(0.009) & 0.747(0.005)&0.746(0.008)\\
        &tox21(t2) &   0.881(0.009) & 0.885(0.013) & 0.881(0.012) & 0.880(0.016) &\textbf{0.889}(0.006)\\
        &toxcast(t8) &   0.490(0.024) & 0.484(0.025) & 0.492(0.025) & \textbf{0.497}(0.029) &0.496(0.032)\\
        &toxcast(t12) &   \textbf{0.734}(0.167) & 0.640(0.214) & 0.673(0.243) & 0.676(0.236) &0.642(0.217)\\
        \midrule
          &STL10 &  0.669(0.099) &  \textbf{0.795}(0.010) &  0.751(0.045)&  0.727(0.064)&0.729(0.078)\\
        &CIFAR10 & 0.564(0.037) &  0.893(0.010) &  \textbf{0.908}(0.004)&  0.902(0.005)&0.907(0.005)\\
       &CIFAR100 &  0.551(0.012) &  0.665(0.006) &  0.676(0.010)&  0.680(0.008)&\textbf{0.684}(0.006)\\
        &Cat vs Dog & 0.668(0.029) &  \textbf{0.943}(0.004) &  0.930(0.006)&  0.926(0.008)&0.920(0.004)\\
        &Melanoma &  0.714(0.008) &  0.782(0.022) &  0.798(0.013)&  \textbf{0.802}(0.019)&0.783(0.017)\\
        &Cardiomegaly & \textbf{0.838}(0.015) &  0.833(0.005) & 0.832(0.005)&  0.833(0.005)	&0.832(0.006)\\
        CSQ &HIV &   0.763(0.015) & \textbf{0.766}(0.016) & 0.763(0.015) & 0.747(0.012)&0.741(0.012)\\
        &MUV(bioassay) &   0.579(0.092) & 0.629(0.073) & 0.689(0.056) & \textbf{0.702}(0.092)&0.659(0.071)\\
        &tox21(t0) &   \textbf{0.761}(0.016) & 0.756(0.016) & 0.758(0.019) & 0.757(0.016)&0.760(0.020)\\
        &tox21(t2) &   0.887(0.009) & 0.887(0.009) & 0.881(0.012) & 0.890(0.006)&\textbf{0.892}(0.005)\\
        &toxcast(t8) &   \textbf{0.487}(0.030) & 0.479(0.027) & \textbf{0.487}(0.023) & 0.470(0.030)&0.476(0.022)\\
        &toxcast(t12) &   0.708(0.160) & 0.727(0.152) & \textbf{0.746}(0.111) & 0.740(0.109)&0.723(0.152)\\
        \bottomrule
    \end{tabular}}
\end{table}
\begin{table}[h!]
    \centering
    \caption{Normalization function for prediction layer with testing AUROC}
    \label{tab:te_activation}
    \resizebox{1.0\textwidth}{!}{  
    \begin{tabular}{llllllll}
       Loss &Dataset & None &sigmoid &$\ell_1$-normalization & $\ell_2$-normalization &batch normalization\\
        \hline
        &STL10  & 0.726(0.030) & \textbf{0.772}(0.023)& 0.734(0.012)& 0.769(0.006) & 0.747(0.020)\\
        &CIFAR10 & 0.900(0.005) & \textbf{0.911}(0.001)& 0.853(0.004)& 0.893(0.006) & 0.893(0.009)\\  
      &CIFAR100 &  0.660(0.012) & 0.681(0.007)& 0.669(0.014)& \textbf{0.683}(0.006) &  0.667(0.002)\\
        &Cat vs Dog &  0.895(0.036) & \textbf{0.932}(0.003)& 0.824(0.030)& 0.919(0.004) &  0.922(0.005)\\
        &Melanoma &  0.793(0.035) & 0.794(0.013)& 0.780(0.011)& \textbf{0.810}(0.011) &  0.805(0.010)\\
        &Cardiomegaly & 0.800(0.010) & \textbf{0.832}(0.007)& \textbf{0.832}(0.007)& 0.815(0.030) &  \textbf{0.832}(0.007)\\
        PSQ  &HIV &   \textbf{0.760}(0.010) & 0.734(0.031) & 0.731(0.013) & 0.745(0.006) & 0.755(0.019)\\
        &MUV(bioassay) &   0.579(0.101) & 0.598(0.063) & 0.603(0.034) & \textbf{0.604}(0.049) & 0.580(0.194)\\
        &Tox21(t0) &   0.744(0.008) & \textbf{0.782}(0.004) & 0.772(0.017) & 0.768(0.017) &0.738(0.014)\\
        &Tox21(t2) &   0.883(0.009) & 0.883(0.012) & 0.886(0.006) & \textbf{0.888}(0.008) &0.886(0.015)\\
        &ToxCast(t8) &   \textbf{0.518}(0.051) & 0.500(0.040) & 0.469(0.025) & 0.477(0.068) & 0.481(0.031)\\
        &ToxCast(t12) &  0.735(0.170) & \textbf{0.840}(0.104) & 0.590(0.052) & 0.613(0.054) &0.617(0.105)\\
        \midrule
         &STL10 &  0.656(0.079) & 0.751(0.046)& 0.698(0.019)& \textbf{0.763}(0.011) &  0.746(0.021)\\
        &CIFAR10 & 0.781(0.048) & \textbf{0.908}(0.004)& 0.853(0.006)& 0.894(0.006) &  0.898(0.003)\\
       &CIFAR100 &  0.631(0.025) & \textbf{0.676}(0.010)& \textbf{0.676}(0.006)& 0.675(0.007) &  0.670(0.004)\\
        &Cat vs Dog &  0.814(0.114) & \textbf{0.929}(0.006)& 0.842(0.023)& 0.916(0.009) & 0.917(0.010)\\
        &Melanoma & 0.801(0.015) & 0.798(0.013)& 0.788(0.015)& \textbf{0.811}(0.013) &  0.799(0.029)\\
        &Cardiomegaly & 0.809(0.015) & \textbf{0.833}(0.005)& \textbf{0.833}(0.005)& 0.814(0.024) &  0.830(0.005)\\
        CSQ &HIV &   \textbf{0.756}(0.019) & 0.744(0.017) & 0.734(0.012) & 0.741(0.008) & \textbf{0.756}(0.006)\\
        &MUV(bioassay) &   0.639(0.091) & 0.599(0.063) & \textbf{0.641}(0.072) & 0.578(0.107) & 0.597(0.103)\\
        &Tox21(t0) &   0.756(0.018) & 0.755(0.005) & 0.754(0.035) & \textbf{0.773}(0.014) &0.750(0.023)\\
        &Tox21(t2) &   0.889(0.004) & \textbf{0.895}(0.005) & 0.887(0.003) & 0.892(0.008) &0.886(0.006)\\
        &ToxCast(t8) &   0.490(0.033) & \textbf{0.523}(0.065) & 0.456(0.034) & 0.483(0.047) &0.499(0.055)\\
        &ToxCast(t12) &   0.721(0.150) & \textbf{0.985}(0.008) & 0.585(0.042) & 0.769(0.061) &0.660(0.092)\\
        \bottomrule
    \end{tabular}}
\end{table}

\begin{table}[h]
    \centering
    \caption{Optimizers with testing AUROC}
    \label{tab:te_opt_and_lrs}
    \resizebox{0.7\textwidth}{!}{
    \begin{tabular}{lllll}
       Loss &Dataset & SGD-style& Momentum-style& Adam-style\\
        \hline
        &STL10 &   0.771(0.007) &  \textbf{0.803}(0.015) & 0.794(0.011)\\
        &CIFAR10 &   0.909(0.002) &  \textbf{0.911}(0.001) & 0.868(0.004)\\
      &CIFAR100 &    \textbf{0.688}(0.002) &  0.681(0.007) & 0.669(0.011)\\
        &Cat vs Dog &    \textbf{0.937}(0.004) &  0.932(0.003) & 0.902(0.010)\\
        &Melanoma &   0.804(0.024) &  0.801(0.009) &
        \textbf{0.810}(0.015)\\
        &Cardiomegaly & 0.848(0.005)   &  0.848(0.005)   & \textbf{0.853}(0.009) \\
        PSQ  &HIV &   \textbf{0.767}(0.005) & \textbf{0.767}(0.005) & 0.766(0.002) \\
        &MUV(bioassay) &   0.690(0.030) & \textbf{0.703}(0.030) & 0.702(0.053) \\
        &Tox21(t0) &   \textbf{0.788}(0.005) & \textbf{0.788}(0.005) & 0.752(0.011) \\
        &Tox21(t2) &   0.903(0.002) & \textbf{0.904}(0.001) & 0.893(0.004)\\
        &ToxCast(t8) &   0.444(0.018) & 0.460(0.104) & \textbf{0.553}(0.056) \\
        &ToxCast(t12) &   0.908(0.036) & 0.908(0.047) & \textbf{0.910}(0.040)\\
        \hdashline
        &STL10 &      1e-1& 1e-1    & 1e-3  \\
        &CIFAR10 &     1e-1& 1e-2     & 1e-3 \\
       &CIFAR100 &    1e-1& 1e-2    & 1e-4  \\
        &Cat vs Dog & 1e-1    & 1e-3    & 1e-3  \\
        &Melanoma & 1e-2   &  1e-3   & 1e-4 \\
        &Cardiomegaly & 1e-4   &  1e-5   & 1e-3 \\
        PSQ &HIV &   1e-4 & 1e-5 & 1e-4  \\
        &MUV(bioassay) &    1e-1 & 1e-2 & 1e-3 \\
        &Tox21(t0) &   1e-4 &  1e-5 & 1e-3  \\
        &Tox21(t2) &   1e-3 & 1e-4 &  1e-4 \\
        &ToxCast(t8) &   1e-2 &  1e-1 &  1e-2 \\
        &ToxCast(t12) &  1e-2 &   1e-3&  1e-4 \\
    \end{tabular}}
\end{table}

\begin{table}[h]
\vspace{-0.3in}
    \centering
    \resizebox{0.8\textwidth}{!}{
    \begin{tabular}{llllll}
    \toprule
    Loss &Dataset & SGD-style& Momentum-style& Adam-style & PESG\\
    \hline
         &STL10 &   0.796(0.013) &  \textbf{0.819}(0.003) & 0.785(0.016)  & 0.766(0.020)\\
        &CIFAR10 &    \textbf{0.910}(0.002) &  0.905(0.006) & 0.845(0.023)  & 0.886(0.007)\\
      &CIFAR100 &    0.678(0.010) &  \textbf{0.684}(0.005) & 0.652(0.007) & 0.664(0.006)\\
        &Cat vs Dog &   \textbf{0.934}(0.006) &  0.930(0.004) & 0.907(0.008) & 0.916(0.006)\\
        &Melanoma &   0.803(0.020) &  \textbf{0.818}(0.004) & 0.812(0.009)  & 0.815(0.016)\\
        &Cardiomegaly & 0.861(0.004)   &  \textbf{0.862}(0.004)  & 0.840(0.012) & 0.861(0.004)\\
        CSQ  &HIV &   0.770(0.003) & 0.770(0.003) & 0.763(0.005) & \textbf{0.775}(0.007) \\
        &MUV(bioassay) &   0.715(0.040) & 0.699(0.026) & 0.684(0.084) & \textbf{0.721}(0.049) \\
        &Tox21(t0) &   0.779(0.002) & 0.777(0.017) & \textbf{0.781}(0.012)  & 0.778(0.003) \\
        &Tox21(t2) &   0.898(0.003) & \textbf{0.899}(0.002) & 0.893(0.007)& 0.898(0.007)\\
        &ToxCast(t8) &   0.445(0.009) & \textbf{0.499}(0.002) & 0.467(0.055)& 0.443(0.006)\\
        &ToxCast(t12) &   0.875(0.084) & 0.883(0.047) & \textbf{0.900}(0.036) & 0.821(0.044)\\
        \hdashline
         &STL10 &    1e-1& 1e-1   &1e-3 & 1e-1 \\
        &CIFAR10 &    1e-1& 1e-2    &1e-3 & 1e-1 \\
        &CIFAR100 &   1e-1& 1e-2   &1e-4 & 1e-2 \\
         &Cat vs Dog &   1e-1 & 1e-3    & 1e-3 & 1e-1\\
         &Melanoma &  1e-1 &  1e-3   & 1e-5  & 1e-2\\
        &Cardiomegaly & 1e-4   &  1e-5   & 1e-4 & 1e-4\\
         CSQ &HIV &    1e-4&  1e-5&  1e-4 & 1e-4\\
         &MUV(bioassay) &    1e-1& 1e-2 & 1e-3  & 1e-2\\
         &Tox21(t0) &   1e-5  & 1e-4  &  1e-3 & 1e-5\\
         &Tox21(t2) &    1e-3&  1e-4& 1e-3  & 1e-3\\
         &ToxCast(t8) &   1e-3 &  1e-1 & 1e-1 & 1e-4 \\
         &ToxCast(t12) &  1e-2 &  1e-3 & 1e-4 & 1e-2\\
        \bottomrule
    \end{tabular}}    
\end{table}

\begin{table}[h]
    \centering
    \caption{Sampling positive rate with validation AUROC}
    \label{tab:val-spr}
    \resizebox{0.9\textwidth}{!}{
    \begin{tabular}{llllll}
       Loss &Dataset &PR$=$origin & PR$=5\%$& PR$=25\%$& PR$=50\%$\\
        \hline
        &STL10 &  0.776(0.030) &  0.724(0.014) &  0.809(0.027) & \textbf{0.810}(0.027)\\
        &CIFAR10 &  0.916(0.018) &  0.884(0.008) &  \textbf{0.930}(0.015) & 0.928(0.013)\\
       &CIFAR100 &  0.725(0.011) &  0.715(0.021) &  \textbf{0.749}(0.030) & 0.740(0.018)\\
        &Cat vs Dog &  0.912(0.015) &  0.887(0.015) &  \textbf{0.928}(0.012) & 0.924(0.012)\\
        &Melanoma &    0.764(0.014) & \textbf{0.803}(0.010)& 0.802(0.021)& 0.802(0.015)\\
        &Cardiomegaly &    0.873(0.002) & 0.872(0.002) & 0.874(0.002)& \textbf{0.875}(0.002)\\
        PSQ &HIV &   0.875(0.028) & 0.879(0.034) & 0.890(0.034) & \textbf{0.891}(0.036)\\
        &MUV &   \textbf{0.929}(0.050) & 0.902(0.046) & 0.914(0.053) & 0.912(0.060)\\
        &tox21(t0) &   0.861(0.013) & 0.862(0.015) & 0.868(0.020) & \textbf{0.870}(0.020)\\
        &tox21(t2) &   0.954(0.009) & 0.952(0.010) & \textbf{0.957}(0.011) & 0.955(0.007)\\
        &toxcast(t8) &   0.840(0.031) & 0.838(0.040) & \textbf{0.847}(0.036) & 0.844(0.030)\\
        &toxcast(t12) &   \textbf{0.950}(0.029) & 0.948(0.040) & 0.933(0.055) & 0.935(0.056)\\
        \midrule
         &STL10 &  0.680(0.088) &  0.634(0.077) &  0.699(0.113) & \textbf{0.705}(0.107)\\
        &CIFAR10 &  0.871(0.062) &  0.849(0.047) &  \textbf{0.898}(0.048) & 0.893(0.052)\\
          &CIFAR100 &   0.726(0.020) &  0.692(0.028) &  \textbf{0.731}(0.017) & 0.721(0.023)\\
        &Cat vs Dog &   0.812(0.065) & 0.864(0.040) & 0.775(0.094) & \textbf{0.891}(0.030)\\
        &Melanoma &  0.771(0.023) & \textbf{0.802}(0.021) & 0.795(0.018) & 0.796(0.020)\\
        &Cardiomegaly &  0.872(0.002) & 0.871(0.002)& 0.874(0.002) & \textbf{0.876}(0.002)\\
        CSQ&HIV &   0.862(0.035) & 0.874(0.033) & \textbf{0.886}(0.032) & 0.873(0.025)\\
        &MUV &   \textbf{0.918}(0.058) & 0.906(0.068) & 0.899(0.068) & 0.893(0.064)\\
        &tox21(t0) &   0.857(0.016) & 0.859(0.015) & \textbf{0.876}(0.016) & 0.874(0.022)\\
        &tox21(t2) &   0.957(0.009) & 0.951(0.010) & 0.958(0.009) & \textbf{0.960}(0.009)\\
        &toxcast(t8) &   \textbf{0.839}(0.031) & 0.837(0.039) & 0.834(0.034) & 0.837(0.042)\\
        &toxcast(t12) &   0.965(0.043) & \textbf{0.972}(0.038) & 0.964(0.045) & 0.962(0.052)\\
        \bottomrule
    \end{tabular}}
\end{table}

\begin{table}[h]
    \centering
    \caption{Consecutive epoch regularization with validation AUROC}
    \label{tab:val-cer}
    \resizebox{0.7\textwidth}{!}{
    \begin{tabular}{lllll}
       Loss &Dataset & $\gamma=0$& $\gamma=0.002$& $\gamma=0.02$\\
        \hline
        &STL10 & 0.810(0.027) & 0.817(0.033) & \textbf{0.829}(0.019)\\
        &CIFAR10 &  0.928(0.013) & \textbf{0.945}(0.012) & 0.937(0.005)\\
       &CIFAR100 & 0.740(0.018) & \textbf{0.744}(0.020) & 0.706(0.013)\\
        &Cat vs Dog & 0.924(0.012) & 0.940(0.009) & \textbf{0.950}(0.008)\\
        &Melanoma &  0.802(0.015) & 0.797(0.012) & \textbf{0.806}(0.019)\\
      PSQ   &Cardiomegaly &  0.874(0.002) & 0.876(0.002) & \textbf{0.877}(0.002)\\
        &HIV &   \textbf{0.891}(0.036) & 0.885(0.035) & 0.873(0.023) \\
        &MUV &   0.912(0.061) & \textbf{0.915}(0.053) & 0.909(0.046) \\
        &tox21(t0) &   \textbf{0.870}(0.020) & 0.868(0.022) & 0.864(0.027) \\
        &tox21(t2) &   \textbf{0.955}(0.007) & 0.954(0.006) & 0.951(0.007)\\
        &toxcast(t8) &   0.844(0.030) & 0.839(0.030) & \textbf{0.847}(0.039) \\
        &toxcast(t12) &   0.935(0.056) & 0.936(0.057) & \textbf{0.937}(0.058)\\
        \midrule
         &STL10 &  0.705(0.108) & 0.741(0.068) & \textbf{0.829}(0.032)\\
        &CIFAR10 & 0.893(0.052) & \textbf{0.942}(0.011) & 0.932(0.007)\\
       &CIFAR100 & 0.721(0.023) & \textbf{0.741}(0.012) & 0.735(0.014)\\
        &Cat vs Dog & 0.891(0.030) & 0.934(0.008) & \textbf{0.947}(0.007)\\
        &Melanoma &  0.796(0.020) & 0.800(0.018) & \textbf{0.805}(0.018)\\
        &Cardiomegaly &  0.876(0.002) & 0.876(0.002) & \textbf{0.877}(0.002)\\
      CSQ   &HIV &   0.873(0.025) & 0.875(0.026) & \textbf{0.878}(0.027) \\
        &MUV &   0.893(0.064) & 0.901(0.064) & \textbf{0.905}(0.061) \\
        &tox21(t0) &   \textbf{0.874}(0.022) & 0.873(0.022) & 0.873(0.023) \\
        &tox21(t2) &   \textbf{0.960}(0.009) & 0.958(0.007) & 0.956(0.006)\\
        &toxcast(t8) &   0.837(0.042) & 0.831(0.043) & \textbf{0.842}(0.024) \\
        &toxcast(t12) &   0.962(0.052) & 0.961(0.053) & \textbf{0.963}(0.050)\\
        \bottomrule
    \end{tabular}}
\end{table}

\begin{table}[h!]
    \centering
    \caption{Weight decay with validation AUROC}
    \label{tab:val-wd}
    \resizebox{1.0\textwidth}{!}{
    \begin{tabular}{lllllll}
       Loss &Dataset & $WD=1e-1$& $WD=1e-2$& $WD=1e-3$& $WD=1e-4$& $WD=0$\\
        \hline
        &STL10 &  0.780(0.038) &  \textbf{0.834}(0.019) &  0.816(0.035)&  0.817(0.033) &0.810(0.020)\\
        &CIFAR10 &  0.645(0.012) &  \textbf{0.946}(0.003) &  \textbf{0.946}(0.008)&  0.945(0.012) &0.943(0.008) \\
      &CIFAR100 & 0.588(0.015) &  0.735(0.022) &  0.747(0.009)&  0.745(0.020)& \textbf{0.754}(0.006)\\
        &Cat vs Dog &  0.724(0.008) &  \textbf{0.954}(0.005) &  0.944(0.011)&  0.940(0.009)& 0.939(0.011)\\
        &Melanoma &  0.728(0.017) &  \textbf{0.804}(0.018) &  0.795(0.018)&  0.797(0.012) &0.803(0.017)\\
        &Cardiomegaly &  0.838(0.001) &  \textbf{0.877}(0.002) &  0.876(0.002) &  0.876(0.002) &0.876(0.002)\\
       PSQ   &HIV &   0.857(0.032) & 0.877(0.032) & 0.889(0.034) & \textbf{0.891}(0.036)& \textbf{0.891}(0.034)\\
        &MUV &   \textbf{0.923}(0.051) & 0.918(0.049) & 0.921(0.057) & 0.912(0.061) &0.905(0.061)\\
        &tox21(t0) &   0.862(0.025) & 0.865(0.025) & 0.864(0.026) & \textbf{0.870}(0.020) & 0.864(0.026)\\
        &tox21(t2) &   0.949(0.007) & 0.953(0.006) & 0.954(0.007) & \textbf{0.955}(0.007)& 0.954(0.008)\\
        &toxcast(t8) &   0.841(0.031) & \textbf{0.849}(0.037) & 0.843(0.036) & 0.844(0.030) & 0.848(0.036)\\
        &toxcast(t12) &   0.933(0.055) & \textbf{0.937}(0.058) & 0.934(0.056) & 0.935(0.056)&0.935(0.056)\\
        \midrule
          &STL10 &  0.712(0.082) & \textbf{0.815}(0.022) &  0.779(0.049)&  0.741(0.068)&0.771(0.080)\\
        &CIFAR10 & 0.586(0.033) &  \textbf{0.951}(0.007) &  0.946(0.008)&  0.942(0.011) &0.949(0.010)\\
       &CIFAR100 &  0.616(0.016) &  0.727(0.015) &  0.747(0.015)&  0.741(0.012) &\textbf{0.753}(0.011)\\
        &Cat vs Dog & 0.702(0.022) &  \textbf{0.954}(0.008) &  0.936(0.013)&  0.934(0.008)& 0.927(0.016)\\
        &Melanoma &  0.728(0.022) &  0.801(0.013) &  \textbf{0.808}(0.015)&  0.800(0.018) & 0.798(0.014)\\
        &Cardiomegaly &  0.839(0.001) &  \textbf{0.877}(0.002) &  0.876(0.002)&  0.876(0.002) &0.876(0.002)\\
       CSQ  &HIV &   0.870(0.026) & 0.872(0.029) & \textbf{0.880}(0.027) & 0.872(0.025) &0.877(0.029)\\
        &MUV &   \textbf{0.902}(0.066) & 0.901(0.062) & 0.898(0.063) & 0.893(0.064) & 0.899(0.067)\\
        &tox21(t0) &   0.870(0.019) & \textbf{0.874}(0.022) & \textbf{0.874}(0.024) & \textbf{0.874}(0.022) & 0.873(0.022)\\
        &tox21(t2) &   0.954(0.007) & 0.959(0.010) & 0.959(0.009) & \textbf{0.960}(0.009)& \textbf{0.960}(0.009)\\
        &toxcast(t8) &   0.830(0.033) & 0.831(0.025) & \textbf{0.838}(0.032) & 0.837(0.042) & \textbf{0.838}(0.037)\\
        &toxcast(t12) &   0.959(0.054) & 0.961(0.052) & \textbf{0.962}(0.052) & \textbf{0.962}(0.052) & 0.961(0.051)\\
        \bottomrule
    \end{tabular}}
\end{table}

\begin{table}[h]
    \centering
    \caption{Normalization with validation AUROC}
    \label{tab:val-norm} 
    \resizebox{1.0\textwidth}{!}{
    \begin{tabular}{lllllll}
       Loss &Dataset & None &sigmoid &$\ell_1$-normalization & $\ell_2$-normalization& batch normalization\\
        \hline
        &STL10  & 0.769(0.027) & \textbf{0.816}(0.035)& 0.771(0.020)& 0.812(0.023) & 0.793(0.026)\\
        &CIFAR10 &   0.932(0.014) & \textbf{0.946}(0.008)& 0.898(0.005)& 0.933(0.009) & 0.931(0.012)\\
        &CIFAR100 &  0.727(0.015) & 0.747(0.009)& 0.728(0.011)& \textbf{0.754}(0.008) & 0.736(0.011)\\
        &Cat vs Dog &  0.899(0.028) & \textbf{0.944}(0.011)& 0.856(0.023)& 0.922(0.010)& 0.926(0.011)\\
        &Melanoma &  \textbf{0.796}(0.016) & 0.795(0.018)& 0.790(0.014)& 0.794(0.014) & 0.789(0.016)\\
        &Cardiomegaly & \textbf{0.876}(0.001) & \textbf{0.876}(0.002)& \textbf{0.876}(0.002)& 0.873(0.002) & \textbf{0.876}(0.002)\\
        PSQ &HIV &   0.874(0.025) & 0.809(0.016) & 0.770(0.021) & 0.798(0.022) & \textbf{0.876}(0.024)\\
        &MUV &   0.910(0.047) & 0.895(0.054) & \textbf{0.918}(0.043) & 0.887(0.064) & 0.888(0.067)\\
        &tox21(t0) &   \textbf{0.864}(0.025) & 0.844(0.018) & 0.845(0.020) & 0.846(0.015) & 0.863(0.027)\\
        &tox21(t2) &   0.953(0.007) & 0.953(0.010) & \textbf{0.957}(0.008) & 0.948(0.009) & 0.953(0.010)\\
        &toxcast(t8) &   \textbf{0.841}(0.034) & 0.836(0.033) & 0.835(0.032) & 0.837(0.037) & 0.832(0.040)\\
        &toxcast(t12) &   0.935(0.056) & 0.983(0.028) & \textbf{0.987}(0.018) & 0.985(0.026) &0.966(0.036)\\
        \midrule
         &STL10 &  0.664(0.077) & 0.779(0.049)& 0.724(0.030)& \textbf{0.802}(0.022) & 0.787(0.020)\\
        &CIFAR10 &   0.829(0.043) & \textbf{0.946}(0.008)& 0.896(0.015)& 0.935(0.014)& 0.933(0.013)\\
        &CIFAR100 &  0.689(0.025) & \textbf{0.747}(0.015)& 0.717(0.010)& 0.744(0.015)& 0.738(0.010)\\
        &Cat vs Dog & 0.899(0.028)&  \textbf{0.936}(0.013) & 0.868(0.014)& 0.919(0.017)& 0.928(0.012)\\
        &Melanoma & 0.795(0.019) & \textbf{0.807}(0.015)& 0.789(0.018)& 0.796(0.017) & 0.787(0.024)\\
        &Cardiomegaly & 0.874(0.004) & \textbf{0.876}(0.002)& \textbf{0.876}(0.002)& 0.873(0.001) & \textbf{0.876}(0.002)\\
        CSQ&HIV &   \textbf{0.874}(0.029) & 0.797(0.015) & 0.762(0.020) & 0.795(0.023) & 0.872(0.024)\\
        &MUV &   0.905(0.060) & \textbf{0.921}(0.047) & 0.908(0.038) & 0.882(0.058) & 0.917(0.051)\\
        &tox21(t0) &   \textbf{0.873}(0.023) & 0.826(0.013) & 0.837(0.013) & 0.840(0.021) & 0.863(0.017)\\
        &tox21(t2) &   0.957(0.007) & 0.954(0.011) & \textbf{0.959}(0.008) & 0.950(0.008) & 0.951(0.008)\\
        &toxcast(t8) &   \textbf{0.842}(0.027) & 0.772(0.046) & 0.841(0.048) & 0.830(0.038) & \textbf{0.842}(0.039)\\
        &toxcast(t12) &   0.962(0.050) & 0.902(0.060) & \textbf{0.986}(0.026) & 0.982(0.025) &0.956(0.041)\\
        \bottomrule
    \end{tabular}}
\end{table}
\begin{table}[h]
    \centering
    \caption{Optimizers with validation AUROC}
    \label{tab:val-opt}
    \resizebox{0.7\textwidth}{!}{
    \begin{tabular}{lllll}
       Loss &Dataset & SGD-style& Momentum-style& Adam-style \\
        \hline
        &STL10 &   0.802(0.014) &  \textbf{0.843}(0.019) & 0.824(0.013)\\
        &CIFAR10 &   \textbf{0.950}(0.009) &  0.946(0.008) & 0.918(0.012)\\
       &CIFAR100 &  \textbf{0.748}(0.012) &  0.747(0.009) & 0.747(0.014)\\
      &Cat vs Dog &   0.937(0.012) &  \textbf{0.940}(0.009) & 0.927(0.008)\\
        &Melanoma &   \textbf{0.807}(0.020) &  0.798(0.022) & 0.799(0.011)\\
         &Cardiomegaly &   0.879(0.002) &  0.879(0.002) & 0.878(0.002)  \\
        PSQ &HIV &   \textbf{0.940}(0.029) & \textbf{0.940}(0.030) & 0.939(0.031) \\
        &MUV &   \textbf{0.967}(0.031) & 0.965(0.034) & 0.965(0.031)  \\
        &tox21(t0) &   \textbf{0.917}(0.023) & 0.915(0.022) & 0.914(0.022) \\
        &tox21(t2) &   \textbf{0.985}(0.010) & \textbf{0.985}(0.009) & \textbf{0.985}(0.009)\\
        &toxcast(t8) &   0.898(0.038) & \textbf{0.899}(0.038) & 0.896(0.038) \\
        &toxcast(t12) &   \textbf{0.994}(0.011) & \textbf{0.994}(0.011) & \textbf{0.994}(0.012)\\
        \hdashline
        &STL10 &     1e-1 & 1e-1    &  1e-3 \\
        &CIFAR10 &   1e-1 & 1e-2    &  1e-3\\
       &CIFAR100 &   1e-1 &  1e-2   &  1e-4\\
        &Cat vs Dog &   1e-1  &   1e-2  & 1e-3  \\
        &Melanoma &  1e-2  &   1e-3  &  1e-3 \\
        &Cardiomegaly &  1e-1  &   1e-2  &  1e-5   \\
        PSQ &HIV &   1e-4 &  1e-5 &  1e-5  \\
        &MUV &    1e-5 & 1e-5 & 1e-5  \\
        &tox21(t0) &   1e-5 & 1e-5  & 1e-5   \\
        &tox21(t2) &   1e-2 & 1e-3 & 1e-5\\
        &toxcast(t8) &   1e-3 & 1e-4  & 1e-5 \\
        &toxcast(t12) &  1e-2 &  1e-3 & 1e-5  \\
        
    \end{tabular}}
\end{table}
\begin{table}[h]
    \centering
    \vspace{-0.3in}
    \resizebox{0.8\textwidth}{!}{
    \begin{tabular}{llllll}
    \toprule
       Loss &Dataset & SGD-style& Momentum-style& Adam-style & PESG\\
        \hline
        &STL10 &   0.818(0.011) & \textbf{0.844}(0.024) & 0.819(0.015) & 0.803(0.026)\\
        &CIFAR10 &   \textbf{0.950}(0.007) &  0.942(0.007) & 0.905(0.003) & 0.932(0.005)\\
        &CIFAR100 &  \textbf{0.756}(0.020) &  0.742(0.023) & 0.736(0.004) & 0.741(0.007)\\
        &Cat vs Dog &    0.943(0.010) &  \textbf{0.944}(0.011) & 0.925(0.010) & 0.926(0.010)\\
        &Melanoma &   0.798(0.014) &  \textbf{0.803}(0.009) & 0.800(0.009) & 0.788(0.015)\\
      &Cardiomegaly &   0.878(0.001) &  0.878(0.001) & 0.872(0.002) & 0.875(0.002)\\
        CSQ &HIV &  0.940(0.030) & 0.940(0.030) & 0.938(0.032) & \textbf{0.945}(0.030) \\
         &MUV &   \textbf{0.966}(0.032) & 0.963(0.032) & 0.963(0.031) & \textbf{0.966}(0.031)\\
         &tox21(t0) &   \textbf{0.917}(0.024) & 0.915(0.024) & 0.907(0.023) & 0.916(0.022) \\
         &tox21(t2) &   \textbf{0.986}(0.009) & \textbf{0.986}(0.009) & 0.985(0.010) & \textbf{0.986}(0.010)\\
        &toxcast(t8) &   \textbf{0.895}(0.041) & \textbf{0.895}(0.041) & 0.888(0.044) & 0.892(0.038) \\
         &toxcast(t12) &   0.993(0.013) & 0.993(0.013) & 0.993(0.014) & 0.993(0.015)\\
        \hdashline
         &STL10 &   1e-1 & 1e-1    & 1e-3  &1e-1\\
        &CIFAR10 &    1e-1 &    1e-2 & 1e-3  &1e-1\\
        &CIFAR100 &  1e-1  &   1e-2  &  1e-4 &1e-1\\
         &Cat vs Dog &    1e-1&   1e-2  & 1e-3  &1e-1 \\
        &Melanoma & 1e-1  &   1e-3  & 1e-4 &1e-2 \\
         &Cardiomegaly & 1e-1  &   1e-2  & 1e-5 &1e-2 \\
         CSQ &HIV &    1e-3& 1e-4 & 1e-5   &1e-3\\
         &MUV &    1e-5& 1e-4 & 1e-5 &1e-5 \\
         &tox21(t0) &   1e-5  & 1e-5  & 1e-5   &1e-5 \\
         &tox21(t2) &    1e-3& 1e-4 & 1e-5 &1e-2 \\
         &toxcast(t8) &   1e-4 &  1e-5 &  1e-5 &1e-5 \\
        &toxcast(t12) &  1e-3 & 1e-4  & 1e-5  &1e-2\\
        \bottomrule
    \end{tabular}}
\end{table}

\end{document}